%% file: neurips_2026.tex
\documentclass{article}

\usepackage[main, final]{neurips_2026}

\usepackage[utf8]{inputenc} 
\usepackage{amsmath}
\usepackage[ruled,vlined,linesnumbered]{algorithm2e}
\usepackage[T1]{fontenc}    
\usepackage{hyperref}       
\usepackage{url}            
\usepackage{booktabs}       
\usepackage{amsfonts}       
\usepackage{nicefrac}       
\usepackage{microtype}      
\usepackage{xcolor}         
\usepackage{enumitem}
\usepackage{cleveref}
\usepackage{pgfplots,pgfplotstable}
\usepackage{svg}
\usepackage{amsmath}
\usepackage{pgfplots}
\usepackage{graphicx}   
\usepackage{wrapfig}
\usepackage{longtable}
\usepackage{ragged2e}
\usepackage{xcolor}
\usepackage{cancel}

\pgfplotsset{compat=1.18}
\usepackage{pgfplots, pgfplotstable, subcaption, tcolorbox, adjustbox}
\usepgfplotslibrary{colorbrewer}
\usepgfplotslibrary{fillbetween}
\usepgfplotslibrary{groupplots}
\usetikzlibrary{calc}
\usetikzlibrary{shadings}
\usetikzlibrary{shapes.geometric}
\usetikzlibrary{arrows.meta, positioning, shapes.multipart}

\newcommand{\Gt}{G_t}
\newcommand{\Gwin}{G_{t-w:t}}
\newcommand{\Sel}{\mathcal{S}}
\newcommand{\Pert}{\mathcal{O}}
\newcommand{\Tq}{T_q}
\newcommand{\Ta}{T_a}
\newcommand{\desc}{d}
\newcommand{\type}[1]{\mbox{\texttt{#1}}}
\newcommand{\rel}[1]{\mbox{\texttt{#1}}}
\newcommand{\prop}[1]{\mbox{\texttt{#1}}}

\newcommand{\ie}{\textit{i.e.}\@}
\newcommand{\eg}{\textit{e.g.}\@}

\input{colors.tex}

\title{Bridging Structure and Language: Graph-Based Visual Reasoning for Autonomous Road Understanding}

\author{
Lena Wild$^{1,2}$ 
\quad Katie Z Luo$^{3,4}$
\quad Marco Pavone$^{3,5}$  \vspace{4pt}\\
$^{1}$KTH Royal Institute of Technology ~~\quad $^2$TRATON ~~\quad $^3$Stanford University \\ \quad $^4$Waymo~~\quad $^5$NVIDIA
\\
{\tt lwild@kth.se, \{katieluo, pavone\}@stanford.edu}
}

\renewcommand{\paragraph}[1]{\vspace{.5em}\noindent\textbf{#1}}

\begin{document}

\maketitle

\input{sections/0_abstract}

\input{sections/1_introduction}
\input{sections/2_relatedworks}
\input{sections/3_methodology}
\input{sections/4_experiments}
\input{sections/5_conclusion}

\input{sections/6_acknowledgements}

\bibliographystyle{unsrt}
\bibliography{references}


\appendix
\include{sections/supplementary_material}




\end{document}

%% file: colors.tex
\usepackage{xcolor}

\definecolor{_darkyellow}{HTML}{E2A652}   
\definecolor{_orange}{HTML}{FF7F13}   
\definecolor{_pink}{HTML}{FF7FDF}   
\definecolor{_red}{HTML}{E36A5A}     
\definecolor{_green}{HTML}{9ED34C}  
\definecolor{_teal}{HTML}{54C7BE}  
\definecolor{_lightgray}{HTML}{E6E6E6}     
\definecolor{_gray}{HTML}{888888}  
\definecolor{_purple}{HTML}{BC88CF}  
\definecolor{_darkgreen}{RGB}{9,133,71}

\definecolor{m1}{HTML}{E2A652} 
\definecolor{m2}{HTML}{FF7F13}   
\definecolor{m3}{HTML}{FF7FDF}   
\definecolor{m4}{HTML}{E36A5A}     
\definecolor{m5}{HTML}{9ED34C}  
\definecolor{m6}{HTML}{54C7BE}  
\definecolor{m7}{HTML}{E6E6E6}     
\definecolor{m8}{HTML}{888888}  
\definecolor{m9}{RGB}{188,189,34}
\definecolor{m10}{RGB}{23,190,207}
\definecolor{m11}{RGB}{123, 114, 237}
\definecolor{m12}{RGB}{9,133,71} 
\definecolor{m13}{RGB}{100,200,100}
\definecolor{m14}{RGB}{9,133,71}  

\definecolor{withimg}{HTML}{54C7BE}   
\definecolor{noimg}{HTML}{FF7F13}     
\definecolor{gtcol}{HTML}{E36A5A}     

%% file: sections/0_abstract.tex
\begin{abstract}
Structured road understanding of lane geometry, topology, and traffic element relationships is foundational to safe autonomous driving. While vision-language models (VLMs) offer promising semantic flexibility, they lack the geometric and relational grounding required for precise road reasoning. Conversely, traditional modular systems, \eg, HD maps and topological road graphs, provide structural precision but remain semantically rigid. To bridge this gap, we introduce the Combined Road Substrate (CRS), a graph-grounded framework that makes geometric road structure and open-vocabulary semantics jointly executable in a single representation. CRS enables the automatic generation of compositionally complex and linguistically varied question-answer pairs via recursive graph queries, augmented with a ``grounding for free'' mechanism that ensures logical traceability to specific map elements, and procedurally extracted chain-of-thought supervision traces. We demonstrate that state-of-the-art VLMs---including large, closed-source models---struggle significantly with structured road reasoning, yet training a small 2- or 4-billion-parameter model with as few as 20 to 80 CRS-enriched scenes yields stable gains in compositional reasoning tasks of varying depth. Analysis of model behavior via verifiable reasoning traces reveals a systematic shift in failure modes: whereas baseline models fail at relational scene understanding, CRS-trained models reduce failures to attribute recognition, suggesting that for structured road reasoning tasks, the lack of structured supervision is a greater limitation than model scale. Project page: \url{https://kth-rpl.github.io/CRS/}.
\end{abstract}

%% file: sections/1_introduction.tex
\section{Introduction}
\label{sec:intro}
For self-driving cars to operate safely, they must be able to reason about the roads they are on---understanding lane structure and geometry, interconnectivity, and relationships between traffic elements such as signs and lights---to determine what behaviors are permitted. This geometrical and relational ``substrate’’ of road understanding forms the foundation for downstream decision-making, and has motivated extensive work in road map prediction \cite{li2022hdmapnet,liu2023vectormapnet,liao2022maptr}, topology reasoning \cite{wang2023openlane,li2023graph}, and methods for incorporating high- and standard-definition maps into downstream tasks \cite{yang2018hdnet}. In such pipelines, structured road understanding is explicitly represented and can be directly supervised. 

More recently, however, end-to-end driving approaches \cite{casas2021mp3, hu2023planning} have gained traction, learning policies that map observations directly to actions. While effective, these methods often do not rely on explicit intermediate road representations, obscuring the causal link between road structure and vehicle behavior. Similarly, foundation models such as vision-language models (VLMs) \cite{liu2023visual, bai2023qwen} exhibit strong generalist capabilities, but often rely on coarse visual and semantic correlations, without explicit mechanisms to ensure precise geometric and relational grounding required for complex road reasoning tasks \cite{deitke2025molmo, chen2026babyvision, stanfordhai2026technical}.

At the same time, VLMs offer a unique opportunity. Unlike modular approaches, which rely on fixed schemas and often compress road structure into predefined, vectorized representations (HD maps, road graphs), VLMs support open-vocabulary reasoning and capture long-tail, semantically rich scene details such as faded markings, temporary changes, or ambiguous road configurations \cite{radford2021learning,alayrac2022flamingo,addepalli2024leveraging}. We hence identify a fundamental representation gap: traditional modular systems are structurally precise but semantically rigid, whereas VLMs today are semantically flexible but structurally ungrounded. 
The challenge is therefore not only to recover the geometrical and topological “substrate” of road understanding in modular approaches, but to extend it beyond the constraints of traditional representations by combining geometric precision with semantic flexibility.

To address this, we introduce the \textbf{Combined Road Substrate (CRS)}, a graph-grounded supervision framework for visual reasoning in road scenes. 
By enforcing a fixed compositional structure while allowing open-vocabulary semantic instantiation---and crucially, by making both directly queryable in a single graph representation---the CRS ensures that semantic expressivity remains grounded in, and constrained by, the underlying road structure. Concretely, reasoning tasks are formulated as queries over the spatio-temporal graph, allowing extraction of precise, compositional, and verifiable supervision signals in natural language in the form of question-answer pairs with hard negative mining. In moving beyond simple QA extraction, the CRS framework enables two key technical capabilities. First, it introduces a ``grounding for free'' mechanism via \textit{Recursive Uniqueness}, which ensures that every generated reasoning task remains logically traceable to specific map elements with unambiguous spatial anchoring. Second, it allows for procedural extraction of the Chain-of-Thought (CoT), providing a verifiable supervision trace for explicit, graph-grounded reasoning.

We show that while even massive closed-source models \cite{google2026gemini31,openai2026gpt54,anthropic2025claudesonnet45} struggle to reason about complex road scenes, training on a remarkably small set (20–80) of CRS-enriched scenes enables VLMs to overcome inherent struggles with road understanding, achieving stable gains in compositional reasoning tasks of varying reasoning depth. 
Leveraging the verifiable CoT traces provided by CRS, we further analyze model behavior and localize reasoning failures to specific stages of the reasoning process. This reveals a systematic shift in failure modes: whereas standard models fail predominantly at relational reasoning (\ie, navigating the scene structure), failures in models trained with CRS are no longer caused primarily by lack of scene understanding, but reduced to failures in attribute recognition (\ie, visual perception). Taken together, these findings suggest that on road understanding questions provided by CRS, the absence of structured supervision limits current VLMs more than model scale.
In summary, our main contributions are as follows:
\begin{itemize}
    \item We introduce the \textbf{Combined Road Substrate (CRS)}, a graph-based representation that unifies geometric structure and open-vocabulary semantics in a jointly executable form, enabling grounded and compositional reasoning over road scenes. 
    
    \item To make the CRS executable, we formulate reasoning tasks as \textbf{recursive queries over the graph}, enabling automatic generation of compositionally complex question–answer pairs with hard negatives. Grounding comes for free via \textit{recursive uniqueness}, and verifiable CoT traces are derived directly from the graph in natural language.

    \item We demonstrate that state-of-the-art VLMs struggle with structured road understanding, and that training a smaller model on as few as 20–80 CRS-enriched scenes yields \textbf{substantial gains in compositional reasoning}, suggesting that CRS-style structured supervision can matter more than scale for compositional road reasoning tasks.
\end{itemize}

%% file: sections/2_relatedworks.tex
\section{Related Work}
\label{sec:related_work}
\paragraph{Road understanding in modular AV stacks.}
Autonomous driving scene understanding has long relied on high-definition (HD) maps providing centimeter-accurate lane geometry, road topology, and semantic annotations \cite{caesar2020nuscenes, Argoverse2, Sun_2020_CVPR}. Early approaches rasterized these representations into bird's-eye-view (BEV) feature maps \cite{zhou2022cross,Seif2016AutonomousDI,yang2018hdnet}, while subsequent work shifted towards predicting HD maps online as vectorized outputs \cite{li2022hdmapnet,liu2023vectormapnet, shin2023instagram}. Today, Transformer-based permutation-equivalent modeling is the dominant real-time methodology \cite{liao2022maptr, liao2025maptrv2}.
While online rasterized or vectorized HD maps represent the geometry of road structure, they usually do not represent crucial relationships between road elements needed for path planning, such as signs and traffic lights to lanes \cite{wang2023openlane}.
Different from HD map prediction, lane-topology prediction jointly predicts road geometry and the governance relationships between traffic controls and lanes \cite{li2023graph,li2023lanesegnet,wu2023topomlp,luo2024augmenting}, enabling richer road reasoning.
Complementing map geometry, scene graph representations encode both static road entities and dynamic actors (\ie, vehicles, pedestrians) as nodes with relational edges, enabling joint reasoning over infrastructure and behavior \cite{zhang2024graphad,schmidt2025graphpilot}.
However, these structured representations over road scenes still need to be consumed by downstream modules for actionable road understanding.

\paragraph{Vision-language reasoning in autonomous driving.}
Language has recently emerged as a flexible interface for querying and reasoning about visual scenes, enabling models to express spatial relationships, object attributes, and compositional structure in natural language \cite{antol2015vqa,goyal2017making,hudson2019gqa,chen2024spatialvlm,wang2025embodied}. 
Seminal work has transferred this paradigm to autonomous driving, probing spatial relationships, object states, and navigational intent through QA-based benchmarks \cite{qian2024nuscenes, marcu2024lingoqa}. Beyond single-step perception queries, subsequent work couples vision-language models with structured reasoning over driving scenes - modeling perception, prediction, and planning - as connected question-answer nodes \cite{sima2024drivelm}, or integrates large VLMs directly with motion planners \cite{deruyttere2019talk2car,malla2023drama,choudhary2024talk2bev,tian2024drivevlm}.
Most recently, work has shifted toward explicit chain-of-thought and causal reasoning benchmarks \cite{nie2024reason2drive,wang2025embodied}, including retrieval-augmented approaches that ground reasoning chains in visual crops \cite{corbiere2025retrieval}.
While these approaches successfully build a bridge between language and driving, their primary focus often lies in action-oriented tasks rather than explicitly grounding the rich language-guided reasoning in the underlying geometric and topological substrate. 

\paragraph{Representations for bridging structure and language.}
Beyond driving-specific approaches, a broader line of work has explored how to combine structured representations with language as a means to derive expressive supervision for grounded reasoning.
A central challenge is balancing semantic flexibility with structural grounding: learned captioning \cite{vinyals2015show} and vision-language models trained on image-text pairs \cite{radford2021learning,li2023blip,liu2023visual,marcu2024lingoqa} are expressive, but remain weakly grounded and hard to scale, while supervision derived from structured representations - such as template-based QA generation  \cite{hudson2019gqa,qian2024nuscenes} - ensures verifiability, but constrains linguistic naturalness and diversity.
For reasoning tasks specifically, recent work leverages procedural generation to produce datasets with controlled reasoning complexity and strong guarantees such as leakage resistance \cite{zhou2025gsm,gong2025phantomwiki}, demonstrating the benefits of structured, executable supervision. Extending this idea to visual domains, scene graphs provide a structured yet expressive representation of entities and their relationships, and have been used to ground visual understanding in VLMs \cite{johnson2015image,yang2019auto,shi2019explainable,damodaran2021understanding,he2024g,graphvis2024}. 
Together, these lines of work suggest that structured, controllable representations are key for probing compositional reasoning, but lack a flexible language interface that tightly couples semantics with the underlying structure. 

%% file: sections/3_methodology.tex
\section{The Combined Road Substrate}
\label{sec:methodology}
To couple the compositional structure required for reasoning (\eg, ``the lane that is left of the ego lane'') with the semantic flexibility needed to capture real-world complexity in road reasoning, we introduce the \textbf{Combined Road Substrate}, a graph-based scene representation that makes structure and semantics jointly executable.
Concretely, our framework consists of four key components:
(i)~\textbf{graph primitives} that define the minimal \textit{formal structure} of the graph representation,  
(ii)~\textbf{canonical operators} that act as an interface between the graph primitives and the open vocabulary language space that provides the content of the graph, 
(iii)~\textbf{well-posedness constraints} that ensure linguistic references to graph elements are unambiguous, hence reasoning over the graph is executable; and
(iv) a \textbf{query instantiation mechanism} that defines reasoning tasks as structured answer retrieval through search queries over the graph, while mining hard negatives and chain-of-thought reasoning traces within the same framework.

\newcommand{\uni}{\scalebox{0.7}{$^*$}}
\newcommand{\suni}{\scalebox{0.6}{$^*$}}
\newcommand{\ImgPin}[4]{%
  \node[
    circle,
    draw=#4,
    fill=white,
    solid,
    line width=0.25pt,
    inner sep=0.5pt,
    font=\fontsize{3.5}{4}\selectfont,
    text=#4,
    anchor=center
  ] at (#1,#2) {#3};
}

\newcommand{\Pin}[3]{%
  \node[
    circle,
    draw=#3,
    fill=white,
    solid,
    line width=0.25pt,
    inner sep=0.4pt,
    font=\fontsize{3.5}{4}\selectfont,
    text=#3,
    anchor=center
  ] at ([xshift=-1pt,yshift=-1pt]#1.north east) {#2};
}

\newcommand{\TextPin}[2]{%
  \tikz[baseline=(p.base)]{
    \node[
      circle,
      draw=#2,
      solid,
      fill=white,
      line width=0.25pt,
      inner sep=0.5pt,
      font=\fontsize{3.5}{4}\selectfont,
      text=#2
    ] (p) {#1};
  }%
}

\newcommand{\InlineEntity}[3]{%
  \tikz[baseline=(e.base)]{
    \node[
      rounded corners=2pt,
      fill=#2,
      text=black,
      font=\fontsize{5}{7}\selectfont,
      inner xsep=2pt,
      inner ysep=1pt
    ] (e) {\textbf{#1}};
  }%
}

\newcommand{\ImgBoxPin}[8]{%

  \draw[
    draw=#6,
    line width=0.4pt,
    rounded corners=1pt
  ] (#1,#2) rectangle (#3,#4);

  \node[
    circle,
    draw=#6,
    fill=white,
    solid,
    line width=0.25pt,
    inner sep=0.5pt,
    font=\fontsize{3.5}{4}\selectfont,
    text=#6,
    anchor=center
  ]
  at ([xshift=#7,yshift=#8]#3,#4)
  {#5};
}

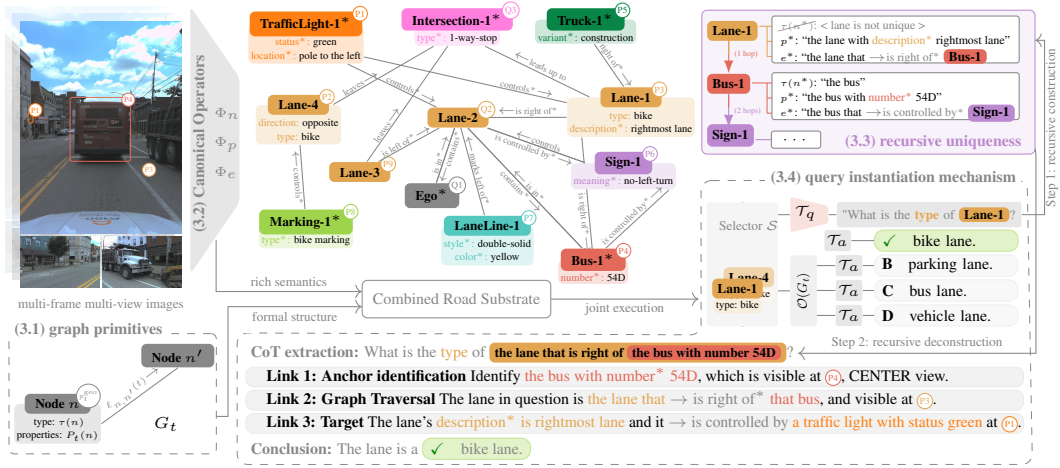
\begin{figure}
\centering
\usetikzlibrary{fit,backgrounds}

\begin{tikzpicture}[
  box/.style={
    draw,
    fill=teal!100!black,
    text=white,
    align=center,
    inner sep=6pt,
    minimum height=2.2cm,
    text width=#1
  },
  graphbox/.style={
    draw,
    fill=teal!100!black,
    align=center,
    inner sep=4pt,
    minimum height=2.7cm,
    text width=#1
  },
  arrow/.style={-{Latex[length=3mm]}, thick}
]

\foreach \i/\dx/\dy/\op in {
  1/0mm/0mm/0.1,
  2/1mm/-1mm/0.2,
  3/2mm/-2mm/1
} {
\node[
  draw=none,
  fill=none,
  align=center,
  inner sep=0pt,
  anchor=north west,
  opacity=\op   
] (A\i) at (\dx,\dy) {%
\scalebox{0.85}{%
\begin{minipage}[t]{0.18\linewidth}
\centering

\begin{tikzpicture}
\node[inner sep=0, anchor=south west] (img) at (0,0) {
  \includegraphics[width=\linewidth,height=3.3cm]{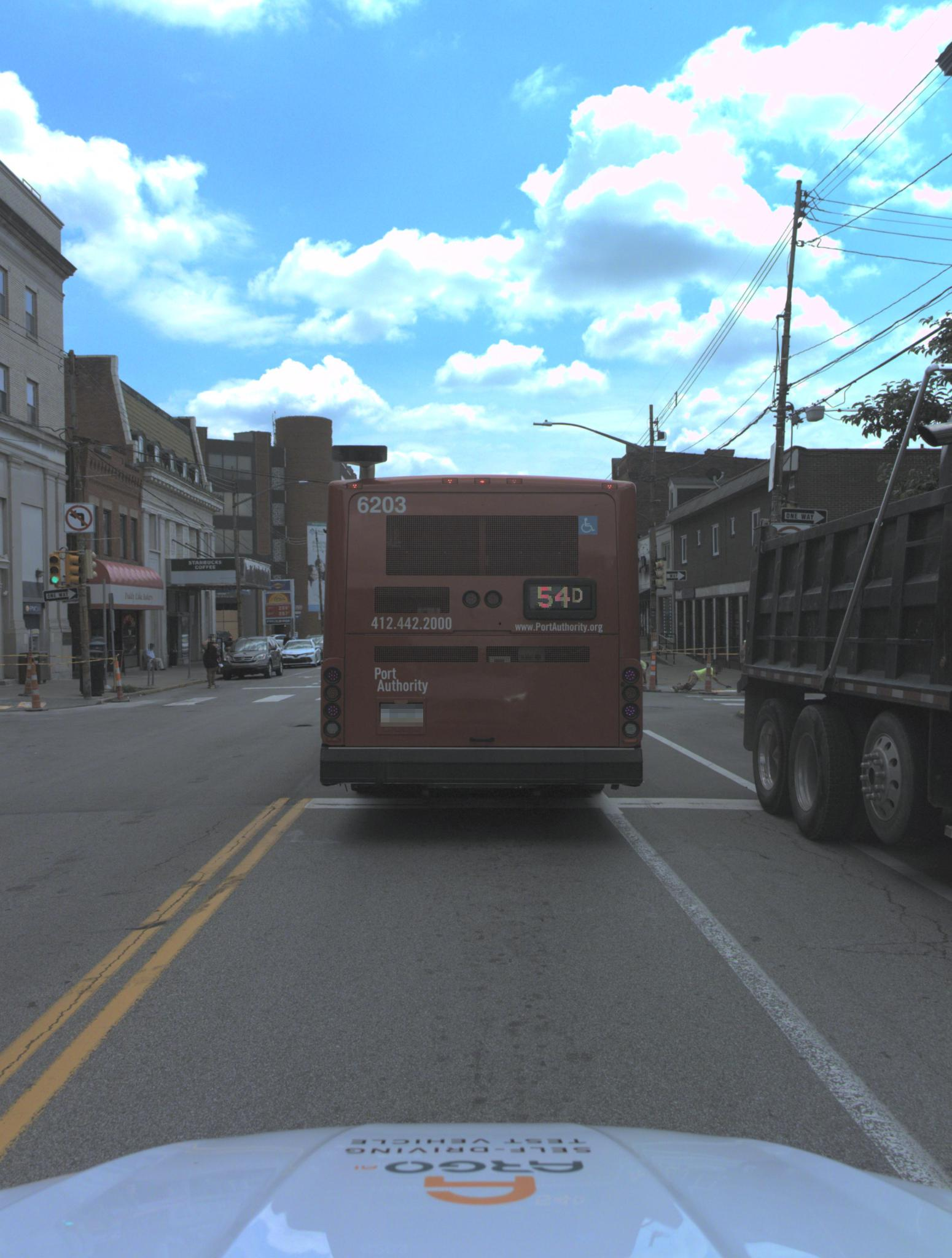}
};

\begin{scope}[x={(img.south east)}, y={(img.north west)}]
  \ImgPin{0.77}{0.30}{P3}{_darkyellow}
   \ImgBoxPin
  {0.05}{0.52}   
  {0.08}{0.57}   
  {P1}           
  {_orange}  
  {1pt}         
  {1pt}         
  \ImgBoxPin
  {0.3}{0.34}   
  {0.66}{0.64}   
  {P4}           
  {_red}  
  {-1pt}         
  {-1pt}         
\end{scope}
\end{tikzpicture}

\vspace{0.1mm}

\begin{minipage}[c]{0.49\linewidth}
\centering
\begin{tikzpicture}
\node[inner sep=0, anchor=south west] (img) at (0,0) {
  \includegraphics[width=\linewidth,height=0.8cm]{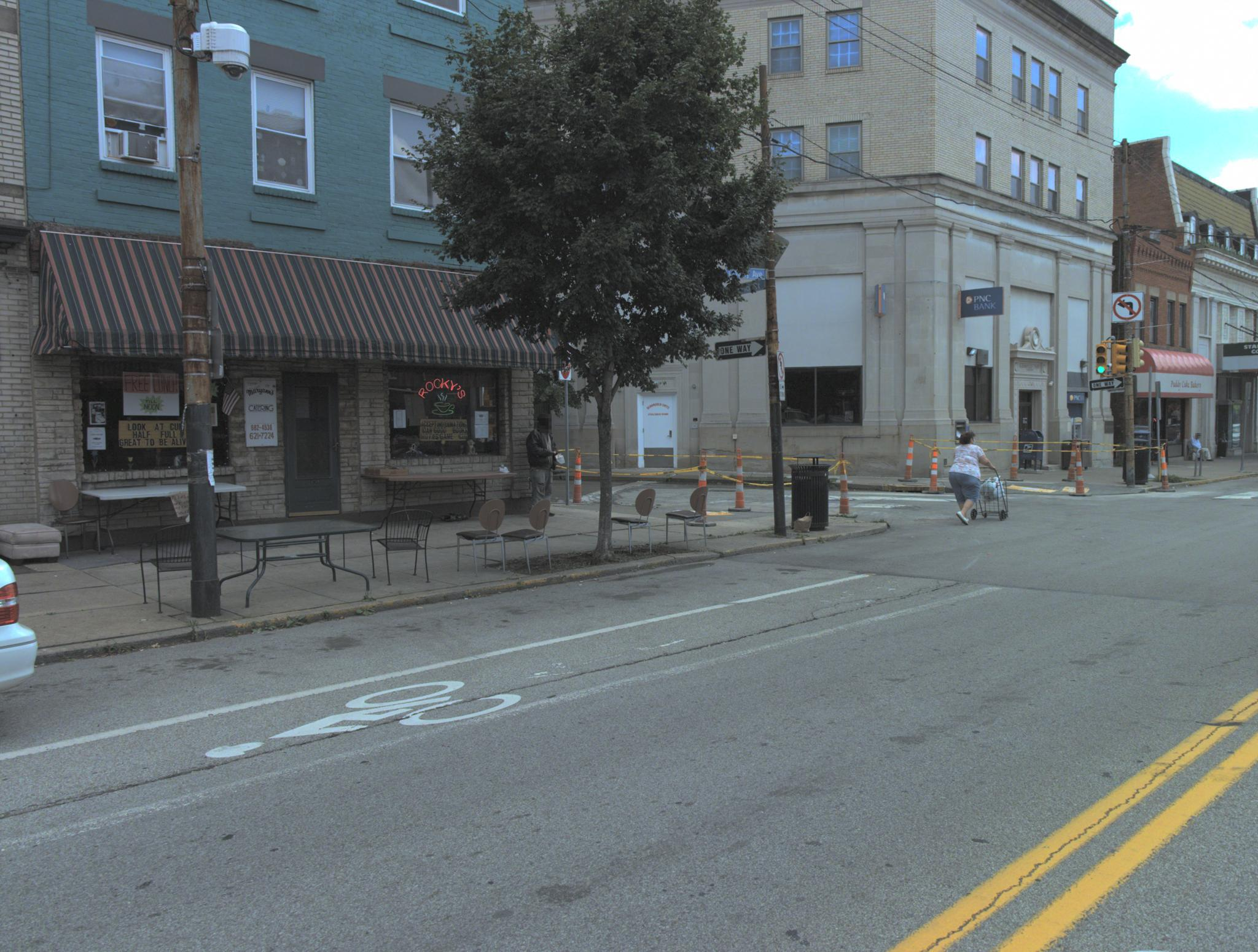}
};
\end{tikzpicture}
\end{minipage}\hfill
\begin{minipage}[c]{0.49\linewidth}
\centering
\begin{tikzpicture}
\node[inner sep=0, anchor=south west] (img) at (0,0) {
  \includegraphics[width=\linewidth,height=0.8cm]{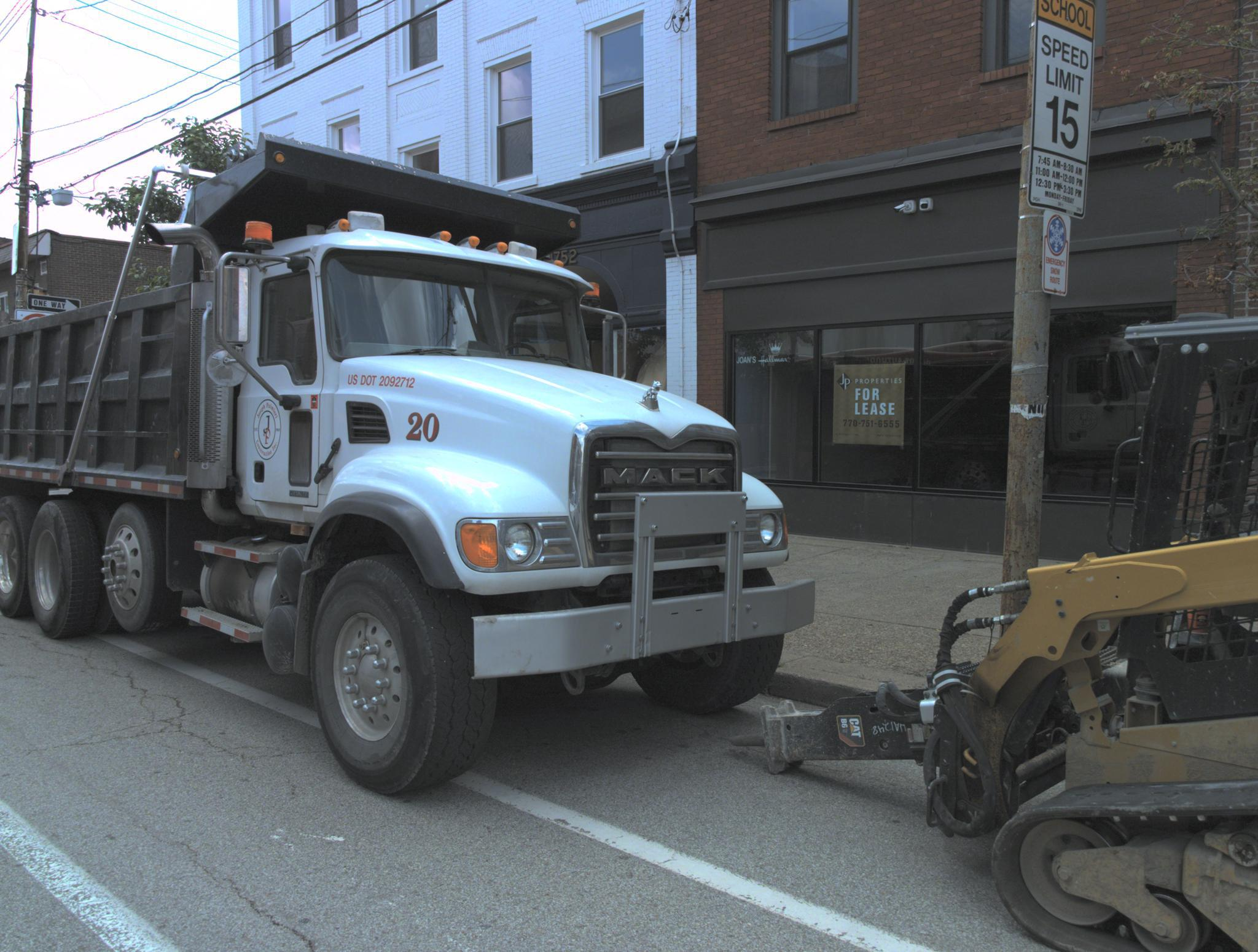}
};
\end{tikzpicture}
\end{minipage}

\end{minipage}%
}%
};
}
\node[
  font=\fontsize{5}{7}\selectfont,
  text=_gray,
  anchor=north,
  align=center
] (m) at ([yshift=-0.5pt]A3.south) {multi-frame multi-view images};

\node[
  draw=_gray,
  dashed,
  line width=0.25pt,
  fill=white,
  rounded corners=2pt,
  align=center,
  inner sep=2pt,
  anchor=north west
] at ([xshift=0pt,yshift=-5pt]m.south west) (C) {%
\begin{minipage}[c][1.45cm][c]{0.18\linewidth}  
\centering
\begin{tikzpicture}[
    >=Latex,
    edge/.style={-, thin, draw=_gray, solid},
    edgelabel/.style={font=\fontsize{3.5}{4}\selectfont, inner sep=0.3pt, _gray},
    titlebase/.style={
        rounded corners=2pt,
        align=center,
        font=\fontsize{5}{7}\selectfont,
        inner xsep=4pt,
        inner ysep=2pt,
        minimum width=0.6cm
    },
    bodybase/.style={
        rounded corners=2pt,
        align=center,
        font=\fontsize{4}{5}\selectfont,
        inner sep=1pt,
        minimum width=0.6cm
    },
    primitiveTitle/.style={titlebase, draw=none, fill=_gray},
    primitiveBody/.style={bodybase, draw=none, fill=_gray!20}
]

\coordinate (n1-pos) at (0,0);
\node[primitiveBody, anchor=north] (n1)
    at ([yshift=0.005cm]n1-pos) {%
    \rule{0pt}{7pt}
    type: \scalebox{0.7}{$\tau(n)$}\\
    properties: \scalebox{0.7}{$P_t(n)$}
};
\node[primitiveTitle, anchor=center] (n1-title)
    at (n1-pos) {\textbf{Node $n$}};
\Pin{n1-title}{\scalebox{0.5}{$p^{\mathrm{geo}}_t$}}{_gray}      

\node[primitiveTitle] (n2) at (1.1,0.8) {\textbf{Node $n'$}};
\node[] (n3) at (1.2,-0.1) {\tiny $G_t$};

\draw[edge, solid] (n1.east) -- node[above, sloped, edgelabel] {\scalebox{0.7}{$\ell_{n,n'}(t)$}$\rightarrow$} (n2.south);

\end{tikzpicture}
\end{minipage}
};

\node[
  fill=white,
  draw=none,
  font=\bfseries\tiny,
  text=_gray,
  inner xsep=2pt,
  inner ysep=0.5pt,
  anchor=north west
] at ([xshift=0pt,yshift=4.5pt]C.north west) {(3.1) graph primitives};

\node[
  draw=none,
  inner sep=0pt,
  anchor=north west
] (B) at ([xshift=3mm]A1.north east) {%
\begin{tikzpicture}

\def\W{0.65}   
\def\H{2.6}    
\pgfmathsetmacro{\dx}{\H / tan(70)} 

\fill[_lightgray, rounded corners=2pt]
  (0,0-\dx)
  -- (\W,0)
  -- (\W,\H-\dx)
  -- (0,\H)
  -- cycle;

\node[
  rotate=90,
  font=\bfseries\tiny,
  text=_gray,
  anchor=center
] at (0.2*\W,0.3*\H) {(3.2) Canonical Operators};

\node[
  draw=none,
  fill=none,
  rounded corners=1pt,
  font=\fontsize{5}{6}\selectfont,
  text=_gray,
  inner xsep=1.5pt,
  inner ysep=1pt,
  minimum width=0.2cm
] at (0.4*\W,0.50*\H) {$\Phi_n$};

\node[
  draw=none,
  fill=none,
  rounded corners=1pt,
  font=\fontsize{5}{6}\selectfont,
  text=_gray,
  inner xsep=1.5pt,
  inner ysep=1pt,
  minimum width=0.2cm
] at (0.4*\W,0.35*\H) {$\Phi_p$};

\node[
  draw=none,
  fill=none,
  rounded corners=1pt,
  font=\fontsize{5}{6}\selectfont,
  text=_gray,
  inner xsep=1.5pt,
  inner ysep=1pt,
  minimum width=0.2cm
] at (0.4*\W,0.20*\H) {$\Phi_e$};

\end{tikzpicture}};

\node[
    draw=none,
    dashed,
    line width=0.1pt,
    fill=_lightgray!0,
    rounded corners=2pt,
    anchor=north west,
    at=(A1.north east),
    xshift=2mm,
    align=left,
    inner xsep=0pt,
    inner ysep=0pt,
    text width=0.8\textwidth,
](X){%
};

\node[
  draw=none,
  fill=none,
  align=center,
  inner sep=4pt,
  anchor=north west,
  at=(B.north east),
  yshift=2mm,
  xshift=-1mm
] (D) {%
\begin{tikzpicture}[
    >=Latex,
    edge/.style={-, thin, draw=_gray},
    edgelabel/.style={font=\fontsize{3.5}{4}\selectfont, inner sep=0.3pt, _gray},
    titlebase/.style={
        rounded corners=2pt,
        align=center,
        font=\fontsize{5}{7}\selectfont,
        inner xsep=4pt,
inner ysep=2pt,
        minimum width=0.6cm
    },
    bodybase/.style={
        rounded corners=2pt,
        align=center,
        font=\fontsize{4}{5}\selectfont,
        inner sep=1pt,
        minimum width=0.6cm
    },
    laneTitle/.style={titlebase, draw=none, fill=_darkyellow},
    laneBody/.style={bodybase, draw=none, fill=_darkyellow!20},
    trafficTitle/.style={titlebase, draw=none, fill=_orange},
    trafficBody/.style={bodybase, draw=none, fill=_orange!20},
    markingTitle/.style={titlebase, draw=none, fill=_green},
    markingBody/.style={bodybase, draw=none, fill=_green!20},
    lanelineTitle/.style={titlebase, draw=none, fill=_teal},
    lanelineBody/.style={bodybase, draw=none, fill=_teal!20},
    intersectionTitle/.style={titlebase, draw=none, fill=_pink},
    intersectionBody/.style={bodybase, draw=none, fill=_pink!20},
    truckTitle/.style={titlebase, draw=none, fill=_darkgreen},
    truckBody/.style={bodybase, draw=none, fill=_darkgreen!20},
    signTitle/.style={titlebase, draw=none, fill=_purple},
    signBody/.style={bodybase, draw=none, fill=_purple!20},
    vehicleTitle/.style={titlebase, draw=none, fill=_red},
    vehicleBody/.style={bodybase, draw=none, fill=_red!20},
    egoTitle/.style={titlebase, draw=none, fill=_gray},
    egoBody/.style={bodybase, draw=none, fill=_gray!20}
]


\coordinate (tl1-pos) at (2,3.62);
\node[trafficBody, anchor=north] (tl1)
    at ([yshift=0.005cm]tl1-pos) {%
    \rule{0pt}{7pt}\textcolor{_orange}{status\uni:} green \\ \textcolor{_orange}{location\uni:} pole to the left};
\node[trafficTitle, anchor=center] (tl1-title)
    at (tl1-pos) {\textbf{TrafficLight-1$^*$}};
\Pin{tl1-title}{P1}{_orange}

\coordinate (lane4-pos) at (1.9,2.55);
\node[laneBody, anchor=north] (lane4)
    at ([yshift=0.005cm]lane4-pos) {%
    \rule{0pt}{7pt}\textcolor{_darkyellow}{direction:} opposite\\
    \textcolor{_darkyellow}{type:} bike};
\node[laneTitle, anchor=center] (lane4-title)
    at (lane4-pos) {\textbf{Lane-4}};
\Pin{lane4-title}{P2}{_darkyellow}

\coordinate (lane1-pos) at (6.3,2.65);
\node[laneBody, anchor=north] (lane1)
    at ([yshift=0.005cm]lane1-pos) {%
    \rule{0pt}{7pt}\textcolor{_darkyellow}{type:} bike\\
   \textcolor{_darkyellow}{description\uni:} rightmost lane};
\node[laneTitle, anchor=center] (lane1-title)
    at (lane1-pos) {\textbf{Lane-1}};
\Pin{lane1-title}{P3}{_darkyellow}

\coordinate (bus1-pos) at (5.8,0.5);
\node[vehicleBody, anchor=north] (bus1)
    at ([yshift=0.005cm]bus1-pos) {%
    \rule{0pt}{7pt}\textcolor{_red}{number\uni:} 54D};
\node[vehicleTitle, anchor=center] (bus1-title)
    at (bus1-pos) {\textbf{Bus-1$^*$}};
\Pin{bus1-title}{P4}{_red}

\coordinate (truck-pos) at (5.7,3.68);
\node[truckBody, anchor=north] (truck)
    at ([yshift=0.005cm]truck-pos) {%
    \rule{0pt}{7pt}\textcolor{_darkgreen}{variant\uni:} construction};
\node[truckTitle, anchor=center] (truck-title)
    at (truck-pos) {\textbf{Truck-1$^*$}};
\Pin{truck-title}{P5}{_darkgreen}

\coordinate (sign-pos) at (6.2,1.78);
\node[signBody, anchor=north] (sign)
    at ([yshift=0.005cm]sign-pos) {%
    \rule{0pt}{7pt}\textcolor{_purple}{meaning\uni:} no-left-turn};
\node[signTitle, anchor=center] (sign-title)
    at (sign-pos) {\textbf{Sign-1}};
\Pin{sign-title}{P6}{_purple}

\coordinate (line1-pos) at (4.4,0.95);
\node[lanelineBody, anchor=north] (line1)
    at ([yshift=0.005cm]line1-pos) {%
    \rule{0pt}{7pt}\textcolor{_teal}{style\uni:} double-solid\\
    \textcolor{_teal}{color\uni:} yellow};
\node[lanelineTitle, anchor=center] (line1-title)
    at (line1-pos) {\textbf{LaneLine-1}};
\Pin{line1-title}{P7}{_teal}

\coordinate (mark1-pos) at (2,1);
\node[markingBody, anchor=north] (mark1)
    at ([yshift=0.005cm]mark1-pos) {%
    \rule{0pt}{7pt}\textcolor{_green}{type\uni:}  bike marking};
\node[markingTitle, anchor=center] (mark1-title)
    at (mark1-pos) {\textbf{Marking-1$^*$}};
\Pin{mark1-title}{P8}{_green}

\node[laneTitle] (lane3) at (2.3,1.8) {\textbf{Lane-3}};
\Pin{lane3}{P9}{_darkyellow}

\node[laneTitle] (lane2) at (3.6,2.52) {\textbf{Lane-2}};
\Pin{lane2}{Q2}{_darkyellow}

\node[egoTitle]  (ego)   at (3.3,1.52) {\textbf{Ego$^*$}};
\Pin{ego}{Q1}{_gray}

\coordinate (intersection-pos) at (4,3.68);
\node[intersectionBody, anchor=north] (intersection1)
    at ([yshift=0.005cm]intersection-pos) {%
    \rule{0pt}{7pt}\textcolor{_pink}{type\uni:}  1-way-stop};
\node[intersectionTitle, anchor=center] (intersection)
    at (intersection-pos) {\textbf{Intersection-1$^*$}};
\Pin{intersection}{Q3}{_pink}

\draw[edge] (tl1) -- node[pos=0.7, above, sloped, edgelabel] {controls\suni$\rightarrow$} (lane2);
\draw[edge] (tl1) -- node[pos=0.8, above, sloped, edgelabel] {controls\suni$\rightarrow$} (lane1);
\begin{scope}[on background layer]
\draw[
  -, 
  thin, 
  draw=_gray
]
  ([xshift=2mm]lane1.south west) -- node[pos=0.7,
    below,
    sloped,
    font=\fontsize{3.5}{4}\selectfont,
    inner sep=0.3pt,
    text=_gray
  ] {is right of\suni$\rightarrow$}
  (bus1.north);
\end{scope}
\draw[edge] ([xshift=-1mm]sign.south east) -- node[above, sloped, edgelabel] {is controlled by\suni$\rightarrow$} (bus1-title);
\draw[edge] (lane3) -- node[above, sloped,edgelabel] {is left of\suni$\rightarrow$} (lane2);
\draw[edge] (lane1) -- node[above, sloped, edgelabel] {$\leftarrow$is right of\suni} (lane2);
\draw[edge] (sign) -- node[above, sloped, edgelabel] {$\leftarrow$controls} (lane2);
\draw[edge] (sign) -- node[below, sloped, edgelabel] {is controlled by\suni$\rightarrow$} (lane2);
\draw[edge] (bus1-title) -- node[above, sloped, edgelabel] {$\leftarrow$ is in\suni} (lane2);
\draw[edge] (bus1-title) -- node[below, sloped, edgelabel] {contains\suni $\rightarrow$} (lane2);
\draw[edge] (ego) -- node[above, sloped, edgelabel] {is in\suni$\rightarrow$} (lane2);
\draw[edge] (ego) -- node[below, sloped, edgelabel] {$\leftarrow$contains\suni} (lane2);
\draw[edge] (line1-title) -- node[above, sloped, edgelabel] {$\leftarrow$marks left of\suni} (lane2);
\draw[edge] (mark1-title) -- node[below, sloped, edgelabel] {$\leftarrow$controls\suni} (lane4);
\draw[edge] (lane4) -- node[pos=0.3, above, sloped, edgelabel] {leaves$\rightarrow$} (intersection1);
\draw[edge] (lane3) -- node[pos=0.3, above, sloped, edgelabel] {leaves$\rightarrow$} (intersection1);
\draw[edge] (lane1) -- node[above, sloped, edgelabel] {$\leftarrow$leads up to} (intersection1);
\draw[edge] (truck) -- node[above, sloped, edgelabel] {is right of\suni$\rightarrow$} (lane1-title);

\path (current bounding box.south) ++(0,-5mm) coordinate (dummy);
\useasboundingbox (current bounding box.north west) rectangle (dummy);
\end{tikzpicture}};

\node[
  draw=none,
  fill=_lightgray!0,
  rounded corners=2pt,
  align=center,
  font=\bfseries\tiny,
  text=_gray,
  inner sep=0pt,
  minimum width=1.8cm,
  minimum height=0.01,
  anchor=west,
  rotate=90
] (Selector) at ([xshift=0mm, yshift=0mm]D.south east) {};

\node[
  draw=none,
  fill=_lightgray!40,
  rounded corners=1pt,
  inner sep=2pt,
  anchor=west
] (SelectedNodes) at ([xshift=-0.02mm,yshift=0mm]Selector.south) {%
\begin{minipage}[t][1.5cm][c]{0.8cm}
\begin{tikzpicture}[
  titlebase/.style={
    rounded corners=2pt,
    align=center,
    font=\fontsize{5}{7}\selectfont,
    inner sep=2pt,
    minimum width=0.6cm
  },
  bodybase/.style={
    rounded corners=2pt,
    align=center,
    font=\fontsize{4}{5}\selectfont,
    inner sep=1pt,
    minimum width=0.6cm
  },
  laneTitle/.style={titlebase, draw=none, fill=_darkyellow},
  laneBody/.style={bodybase, draw=none, fill=_darkyellow!20}
]

\node[
  font=\fontsize{5}{7}\selectfont,
  text=_gray,
  anchor=south
] at (-0,0.55) {Selector $\mathcal{S}$};

\coordinate (l4) at (0,0);
\node[laneBody, anchor=north, fill=_darkyellow!10] (l4body)
  at ([yshift=-0.125cm]l4)
  {type: bike};
\node[laneTitle, anchor=center, fill=_darkyellow!50] at (l4)
  {\textbf{Lane-4}};

\coordinate (l1) at (-0.15,-0.15);
\node[laneBody, anchor=north] (l1body)
  at ([yshift=-0.125cm]l1)
  {type: bike};
\node[laneTitle, anchor=center] at (l1)
  {\textbf{Lane-1}};

\end{tikzpicture}
    
\end{minipage}
};

\node[
  draw=none,
  inner sep=0pt,
  anchor=south west
] (SelectorBoxes) at ([xshift=0mm,yshift=0mm]SelectedNodes.south east) {%
\begin{tikzpicture}[
  smallbox/.style={
    draw=none,
    fill=_lightgray,
    rounded corners=1pt,
    minimum width=0.35cm,
    minimum height=0.28cm,
    inner sep=0pt,
    font=\fontsize{4}{5}\selectfont,
    text=black,
    align=center
  }
]
\node[smallbox] (s1) at (0.4,0.99) {$\mathcal{T}_a$};
\node[smallbox] (s2) at (0.55,0.66) {$\mathcal{T}_a$};
\node[smallbox] (s3) at (0.55,0.33) {$\mathcal{T}_a$};
\node[smallbox] (s4) at (0.55,0.00) {$\mathcal{T}_a$};

\node[smallbox, fill=none] (spanbox) at (-0.2,0.00) {};

\draw[-, thin, draw=_gray] (spanbox.east) |- (s2.west) ;
\draw[-, thin, draw=_gray] (spanbox.east) |- (s3.west) ;
\draw[-, thin, draw=_gray] (spanbox.east) |- (s4.west) ;

\tikzset{
  answerpill/.style={
    draw=#1!60,
    fill=#1!30,
    rounded corners=1.2mm,
    line width=0.25pt,
    inner xsep=1mm,
    inner ysep=0.1mm,
    minimum width=1.7cm,
    text width=1.7cm, 
    minimum height=0.28cm,
    font=\tiny,
    align=left,
    anchor=west
  }
}

\node[answerpill=_lightgray] (AnsA) at ([xshift=1.5mm]s2.east)
  {\textbf{B}\quad parking lane.};

\node[answerpill=_green] (AnsB) at ([xshift=3mm]s1.east)
  {{\color{green!50!black}\checkmark}\quad bike lane.};

\node[answerpill=_lightgray] (AnsC) at ([xshift=1.5mm]s3.east)
  {\textbf{C}\quad bus lane.};

\node[answerpill=_lightgray] (AnsD) at ([xshift=1.5mm]s4.east)
  {\textbf{D}\quad vehicle lane.};

\draw[-, thin, draw=_gray] (s1.east) -- (AnsB.west);
\draw[-, thin, draw=_gray] (s2.east) -- (AnsA.west);
\draw[-, thin, draw=_gray] (s3.east) -- (AnsC.west);
\draw[-, thin, draw=_gray] (s4.east) -- (AnsD.west);
\end{tikzpicture}
};

\node[
  draw=none,
  fill=_lightgray!60,
  rounded corners=1pt,
  minimum width=0.35cm,
  minimum height=0.94cm,
  inner sep=0pt,
  anchor=south west
] (spanbox) at ([xshift=1.5mm,yshift=0mm]SelectedNodes.south east){%
  \rotatebox{90}{\scalebox{0.5}{$\mathcal{O}(G_t)$}}%
};

\node[
  draw=none,
  inner sep=0pt,
  anchor=south west
] (SelectorTrap) at ([xshift=1.5mm,yshift=-0.5mm]SelectorBoxes.north west) {%
\begin{tikzpicture}
\def\W{0.5}
\def\H{0.35}
\def\dx{0.12}

\fill[_red!20, rounded corners=1pt]
  (0,0-\dx)
  -- (\W,0)
  -- (\W,\H-\dx)
  -- (0,\H)
  -- cycle;

\node[
  rotate=0,
  align=center,
  font=\fontsize{5}{7}\selectfont,
  text=black
] at (0.1*\W,0.1*\H) {%
\scalebox{1}{$\mathcal{T}_q$}%
};
\end{tikzpicture}
};

\node[
  draw=none,
  fill=_gray!20,
  rounded corners=1pt,
  align=left,
  font=\fontsize{5}{7}\selectfont,
  text=_gray,
  inner xsep=2pt,
  inner ysep=2pt,
  text width=2.3cm,
  anchor=west
] (QuestionBox) at ([xshift=1mm,yshift=0mm]SelectorTrap.east) {"What is the \textcolor{_darkyellow}{\textbf{type}} of \InlineEntity{Lane-1}{_darkyellow}{P1}?%
};

\node[
  draw=_purple,
  solid,
  line width=0.1pt,
  fill=_purple!10,
  align=center,
  inner sep=2pt,
  rounded corners=2pt,
  anchor=south west
] (F) at ([yshift=5.5mm, xshift=-0.5mm]Selector.north east) {%
\begin{tikzpicture}[
    >=Latex,
    entity/.style={
        rounded corners=2pt,
        text=black,
        font=\fontsize{5}{7}\selectfont,
        inner xsep=2pt,
        inner ysep=2pt
    },
    spine/.style={_gray, line width=0.8pt},
    arrow/.style={-{Latex[length=2mm]}, _gray, line width=0.6pt},
    key/.style={text=gray!60, font=\bfseries\scriptsize},
    val/.style={text=black, font=\scriptsize, align=left}
]

\node[entity, fill=_darkyellow] (v1) at (-0,0.1) {\fontsize{5}{7}\selectfont{\textbf{Lane-1}}};

\node[entity, fill=_red] (BusFree) at (0,-0.6) {\fontsize{5}{7}\selectfont{\textbf{Bus-1}}};

\node[entity, fill=_purple] (IntersectionFree) at (0,-1.3)
  {\fontsize{5}{7}\selectfont{\textbf{Sign-1}}};

\def\xA{0.8}
\def\xB{0.8}
\draw[spine, thin, draw=_darkyellow, solid] (\xA,0.3) -- (\xA,-0.1);
\draw[spine, thin, draw=_darkyellow, solid] (\xA,0.3) -- (\xA-0.1,0.3);
\draw[spine, thin, draw=_purple, solid] (\xA+0.1,-0.65+0.2-0.75) -- (\xA-0.15,-0.65+0.2-0.75);

\def\yOne{0.3}
\def\yTwo{0.1}
\def\yThree{-0.1}
\def\yFour{-0.65}
\def\yFive{-1.5}

\draw[spine, thin, draw=_red, solid] (\xB,\yFour+0.2) -- (\xB,\yFour-0.2);

\draw[spine, thin, draw=_red, solid] (\xB,\yFour+0.2) -- (\xB-0.2,\yFour+0.2);

\foreach \y in {\yOne,\yTwo,\yThree} {
    \draw[thin, draw=_darkyellow, solid] (\xA,\y) -- ++(0.1,0);
}
\foreach \y in {\yOne,\yTwo,\yThree} {
   \draw[thin, draw=_red, solid] (\xB,\y+\yFour-0.1) -- ++(0.1,0);
}

  \node[
  draw=_gray,
  fill=_lightgray!0,
  rounded corners=1pt,
  inner xsep=4pt,
  inner ysep=2pt,
  font=\fontsize{4.5}{7}\selectfont,
  anchor=west
] (IntersectionDots) at (\xB+0.05,-1.2) {$\cdots$};

\node[
  draw=_gray,
  fill=_lightgray!0,
  rounded corners=1pt,
  inner xsep=1pt,
  inner ysep=1pt,
  fit={(\xA+0.1,\yOne+0.1) (\xA+3.3,\yThree-0.05)}
] {};

\node[
  draw=_gray,
  fill=_lightgray!0,
  rounded corners=1pt,
  inner xsep=1pt,
  inner ysep=1pt,
  fit={(\xB+0.1,\yOne+\yFour+0) (\xB+3.3,\yThree+\yFour-0.1-0.05)}
] {};

\node[val, anchor=west] at (\xA+0.1,\yOne) {\fontsize{4.5}{5}\selectfont{\scalebox{0.7}{\textcolor{_gray}{$\cancel{\tau(n^*)}$}}\textcolor{_gray}{: < lane is not unique > }}};

\node[val, anchor=west] at (\xA+0.1,\yTwo) {\fontsize{4.5}{7}\selectfont{\scalebox{0.7}{$p^*$}: ``the lane with \textcolor{_darkyellow}{description\uni} rightmost lane''}};

\node[val, anchor=west] at (\xA+0.1,\yThree) {\fontsize{4.5}{7}\selectfont{\scalebox{0.7}{$e^*$}: ``the lane that \textcolor{_gray}{$\rightarrow$is right of\uni} \InlineEntity{Bus-1}{_red}{P1}  }};

\node[val, anchor=west] at (\xB+0.1,\yOne+\yFour-0.1) {\fontsize{4.5}{7}\selectfont \scalebox{0.7}{$\tau(n^*)$}: ``the bus''};

\node[val, anchor=west] at (\xB+0.1,\yTwo+\yFour-0.1) {\fontsize{4.5}{7}\selectfont \scalebox{0.7}{$p^*$}: ``the bus with \textcolor{_red}{number\uni} 54D''};

\node[val, anchor=west] at (\xB+0.1,\yThree+\yFour-0.1) {\fontsize{4.5}{7}\selectfont \scalebox{0.7}{$e^*$}: ``the bus that \textcolor{_gray}{$\rightarrow$is controlled by\uni} \InlineEntity{Sign-1}{_purple}{P1} };

\coordinate (mid) at ($(BusFree.north)+(0,0.5)$);
\coordinate (mid2) at ($(BusFree.south)+(0,-0.44)$);

\draw[->, thin, draw=_red]
(mid)
  -- node[
       pos=0.5,
       right,
       font=\fontsize{3}{4}\selectfont,
       text=_red
     ] {(1 hop)}
(BusFree.north);
\draw[line width=0.4pt, _darkyellow] (mid) -- ($(mid)!0.5!(BusFree.north)$);

\draw[->, thin, draw=_purple] (BusFree.south) -- node[
       pos=0.5,
       right,
       font=\fontsize{3}{4}\selectfont,
       text=_purple
     ] {(2 hops)} (mid2);
\draw[line width=0.4pt, _red] (BusFree.south) -- ($(BusFree.south)!0.4!(mid2)$);

\end{tikzpicture}
};

\node[
  fill=none,
  draw=none,
  font=\bfseries\tiny,
  text=_purple,
  inner xsep=2pt,
  inner ysep=0.5pt,
  anchor=north east
] at ([xshift=-3pt,yshift=8pt]F.south east) {(3.3) recursive uniqueness};

\node[
  draw=none,
  dashed,
  line width=0.25pt,
  fill=none,
  rounded corners=2pt,
  anchor=north west,
  at=(D.south west),
  inner sep=3pt,
  yshift=-0mm
] (G) {%
\begin{minipage}{10.3cm}
\tcbset{
  stepbox/.style={
    colback=gray!10,
    colframe=gray!20,
    boxrule=0.25pt,
    arc=1mm,
    left=0.6mm,
    right=0.6mm,
    top=0.1mm,
    bottom=0.1mm,
    boxsep=0pt,
    before skip=1pt,
    after skip=1pt
  }
}

\newcommand{\StepBox}[1]{%
  \begin{tcolorbox}[stepbox]
  {\tiny #1}
  \end{tcolorbox}
}

\newcommand{\PlainStepBox}[1]{%
  \begin{tcolorbox}[
    stepbox,
    colback=white,
    colframe=gray!0
  ]
  {\tiny \textcolor{_gray}{#1}}
  \end{tcolorbox}
}

\tikzset{
  answerpill/.style={
    draw=#1!60,
    fill=#1!30,
    solid,
    rounded corners=1.2mm,
    line width=0.25pt,
    inner xsep=1mm,
    inner ysep=0.1mm,
    minimum height=0.28cm,
    font=\tiny,
    align=left
  }
}

\newcommand{\InlineAnswerPill}[2]{%
\tikz[baseline=(a.base)]{
\node[
    answerpill=#2,
    anchor=base,
    inner xsep=1mm,
    inner ysep=0.1mm,
    minimum height=0.28cm,
    font=\tiny,
    align=left
] (a) {#1};
}
}

\PlainStepBox{%
\textbf{CoT extraction:} What is the \textcolor{_darkyellow}{type} of \InlineEntity{the lane that is right of
\InlineEntity{the bus with number 54D}{_red}{P4}}{_darkyellow}{P1}?
}
    
\StepBox{\textbf{\quad Link 1: Anchor identification} Identify \textcolor{_red}{the bus with number$^*$ 54D}, which is visible at \TextPin{P4}{_red}, CENTER view.}

\StepBox{\textbf{\quad Link 2: Graph Traversal} The lane in question is \textcolor{_darkyellow}{the lane that} \textcolor{_gray}{$\rightarrow$ is right of$^*$} \textcolor{_red}{that bus}, and visible at \TextPin{P3}{_darkyellow}.}

\StepBox{\textbf{\quad Link 3: Target} The lane's \textcolor{_darkyellow}{description$^*$ is rightmost lane} and it \textcolor{_gray}{$\rightarrow$ is controlled by} \textcolor{_orange}{a traffic light with status green} at \TextPin{P1}{_orange}. }

\PlainStepBox{\textbf{Conclusion:} The lane is a 
\InlineAnswerPill{{\color{green!50!black}\checkmark}\quad bike lane.}{_green}}

\vspace{-1mm}

\end{minipage}
};
\node[
  fill=white,
  draw=none,
  font=\bfseries\tiny,
  text=_gray,
  inner xsep=2pt,
  inner ysep=0.5pt,
  anchor=south west
] at ([xshift=2pt,yshift=0pt]G.north west) {};

\coordinate (p) at (4.2,-3.5);
\coordinate (q) at (3.1,-3.5);
\coordinate (a) at (2.8,-3.1);

  \draw[-,thin,draw=_gray, rounded corners=0pt] (SelectorTrap.east) -- (QuestionBox.west);
\begin{scope}[on background layer]
\draw[-, thin, draw=_gray]
  (SelectedNodes.north) |- (SelectorTrap.west);

\draw[-, thin, draw=_gray]
  (SelectedNodes.south) |- (spanbox.west);
\end{scope}


\coordinate (LtopLeft)  at ([xshift=-2pt,yshift=8pt]SelectedNodes.north west);
\coordinate (LtopRight) at ([xshift=2pt,yshift=8pt]QuestionBox.north east);
\coordinate (LbotLeft)  at ([xshift=-0pt,yshift=-0pt]G.south west);
\coordinate (LbotRight) at ([xshift=0pt,yshift=-0pt]G.south east);
\coordinate (LbotRight2) at ([xshift=0pt,yshift=-0pt]G.north west);
\coordinate (LtopLeft2)  at ([xshift=-2pt,yshift=-3pt]SelectedNodes.south west);

\begin{scope}[on background layer]
\draw[
  _gray,
  fill=none,
  dashed,
  line width=0.25pt,
  rounded corners=4pt
]
  (LtopLeft)
  -- (LtopRight)
  -- (LbotRight)
  -- (LbotLeft)
  -- (LbotRight2)
  -- (LtopLeft2)
  -- cycle;
\end{scope}

\node[
  fill=white,
  draw=none,
  font=\bfseries\tiny,
  text=_gray,
  inner xsep=2pt,
  inner ysep=0.5pt,
  anchor=north east
] at ([xshift=-2pt,yshift=4pt]LtopRight) {(3.4) query instantiation mechanism};

\pgfdeclarelayer{foreground}
\pgfsetlayers{background,main,foreground}

\node[
  draw=_gray,
  fill=white,
  rounded corners=2pt,
  line width=0.25pt,
  font=\tiny,
  text=_gray,
  inner xsep=4pt,
  inner ysep=5pt,
  anchor=north
] (RoadSubstrate) at ([xshift=-2mm,yshift=7mm]D.south) {Combined Road Substrate};

\draw[->, thin, draw=_gray]
  (RoadSubstrate.east) -- ++(1.95cm,0)
  node[midway, below,_gray, yshift=0.7mm] {\fontsize{5}{7}\selectfont joint execution};
\draw[
  ->,
  thin,
  draw=_gray
] ([yshift=-3mm]C.east)
  -- ([yshift=-3mm,xshift=1.5mm]C.east)
  |- node[pos=0.75, below, _gray,  font=\fontsize{5}{7}\selectfont, yshift=0.5mm] {formal structure}
     ([yshift=-0.1cm] RoadSubstrate.west);
\draw[
  ->,
  thin,
  draw=_gray,
  rounded corners=0pt
] (a)  |- node[pos=0.75, above, _gray, font=\fontsize{5}{7}\selectfont, yshift=-0.5mm] {rich semantics} ([yshift=0.1cm] RoadSubstrate.west);

\coordinate (qtof) at ([xshift=3mm]QuestionBox.east);
\coordinate (ptof) at ([yshift=0mm, xshift=0.1mm]F.south east);
\coordinate (rtof) at (10.5,-4.6);

\draw[->, thin, draw=_gray, rounded corners=0pt]
  (QuestionBox.east)
  -- (qtof)
  |- node[
       pos=0.25,
       rotate=90,
       _gray,
       font=\fontsize{5}{7}\selectfont,
       yshift=-1mm
     ] {Step 1: recursive construction}
     ([yshift=-4mm, xshift=0mm]F.north east);
  
\draw[->, thin, draw=_gray, rounded corners=0pt]
  ([yshift=0.75mm, xshift=0mm]F.south east)
  -- (ptof)
  |- node[pos=0.75, above, _gray, font=\fontsize{5}{7}\selectfont, yshift=-0.7mm]
     {Step 2: recursive deconstruction}(rtof);

  \end{tikzpicture}

\caption{Language descriptions of road scenes are matched to graph primitives through canonical operators to build a semantically rich graph. This Combined Road Substrate is executable and produces questions, answers, decoys and CoT from the same underlying scheme. }
\label{fig:method_figure}
\vspace{-0.3cm}
\end{figure}

\subsection{Graph Primitives}
We represent each scene as a directed graph over time steps $t \in [0, ..., T]$,
\(
G = \{G_0,  \dots, G_T\} \quad \text{with} \quad G_t = (N_t, E_t),
\)
where \(N_t\) is the set of nodes and \(E_t\) the set of directed edges at time $t$. Each node \(n \in N_t\) has a  type $\tau(n)$ and a set of properties
\(
P_t(n) = \{p_t^1, p_t^2, \dots \},
\)
where a property is a tuple \(
p_t^i = (k^i, y^i(t)),
\)
with \(k^i \in \mathcal{K}\) its property key and \(y^i(t) \in \mathcal{Y}\) its corresponding value, which may be static for invariant properties (\(y^i(t) = y_0\)).
Geometric grounding is represented as a special property \(p_t^{\text{geo}} \in P_t(n)\) with values in the image-space \( \mathbb{R}^2 \) or real world coordinates \( \mathbb{R}^3 \).
Finally, each edge \(e_t \in E_t\) is defined as
\(
e_t = (n_t, n'_t, \ell_{n,n'}(t)),
\)
where \(n_t, n'_t \in N_t\) are the source and target nodes, and \(\ell_{n,n'}(t)\) is a (possibly time-dependent) edge label. This formulation provides a minimal structural backbone upon which language-grounded semantics are later imposed.
\subsection{Canonical operators}
A central design goal of CRS is to avoid defining a priori what real-world objects (lanes, vehicles, crossings, etc.) are mapped to the graph. Instead, types, property keys and values, and edge labels should be freely expressible within the open-language space. However, without a well-specified language-structure interface that defines how open-vocabulary statements translate into their graph counterparts, the representation risks becoming inconsistent, ambiguous, and ultimately non-executable.
To resolve this, we impose structure not on \emph{what} can be expressed, but on \emph{how} it must be expressed through three \textit{canonical operators} \(\Phi_{n}, \Phi_{p}, \Phi_{e}\): An open language graph element is valid only if it produces a grammatical and semantically well-typed English sentence under one of the three canonical operators. For node types $\tau(n)$, we define
\begin{equation}
\Phi_{\text{n}}(\tau(n)) =
\texttt{``There exists a <}\tau(n)\texttt{>''}.
\end{equation}
Thus, a node type is valid if it completes an existential statement denoting an identifiable entity or group of objects, \eg, \texttt{``There exists a traffic jam''}. For properties, each \(p_t^i=(k^i,y^i(t)) \in P_t(n)\) must satisfy
\begin{equation}
\Phi_{\text{p}}(n,k^i,y^i(t)) =
\texttt{``The <}k^i\texttt{> of <}\tau(n)\texttt{> is `<}y^i(t)\texttt{>' ''}.
\end{equation}
This rules out ill-typed encodings such as \texttt{is\_closed\_for\_construction = true}, since \texttt{``The is\_closed\_for\_construction of the lane is `true' ''} is not a well-formed attribute statement. Instead, the same information should be represented as, for example, \texttt{status = closed\_for\_construction}. For relations, each edge \(e=(n,n',\ell_{n,n'}(t))\) must satisfy
\begin{equation}
\Phi_{\text{e}}(n,n',\ell_{n,n'}(t)) =
\texttt{``The <}\tau(n)\texttt{> <}\ell_{n,n'}(t)\texttt{> the <}\tau(n')\texttt{>''}.
\end{equation}
Thus, relation labels must form meaningful directed relational statements, which excludes forms such as \texttt{`left of'} or \texttt{`controlling'}, but allows \texttt{`is left of'} or \texttt{`controls'}.
By enforcing a fixed structural form while leaving the semantic space unconstrained, the canonical schema decouples compositional structure from linguistic expressivity. As a result, the graph remains consistent and queryable without restricting the richness or extensibility of its semantic content.

\subsection{Well-definedness constraints}\label{sec:welldefinedness}
To make reasoning over \(G_t\) executable, linguistic references to nodes must be unambiguous and the relevant parts of the graph sufficiently specified. This is non-trivial under our open-vocabulary design: multiple nodes may share the same type or properties (e.g., \texttt{``the traffic light with status `red' ''} may refer to multiple nodes), making language-based references ambiguous. Furthermore, since we do not constrain the space of potential node types, the graph operates under an \emph{open-world assumption}, where absence does not imply non-existence but rather unknown state, and queries requiring exhaustive knowledge (\eg, \texttt{``How many traffic lights are there?''}) could potentially be ill-posed. We address this through two conditions: \emph{uniqueness} and \emph{completeness}.

\paragraph{(a) Uniqueness.}
A node in the graph is well-defined only if it admits a \emph{unique} descriptor for the corresponding object in the scene. A descriptor \(d\) is a language expression derived from \(G_t\) that refers to the real-world object $n$ represented via node type-, property-, or edge-based descriptions (\eg, \texttt{``the traffic light''}, \texttt{``the traffic light with status `red' ''}, \texttt{``the traffic light that controls the ego lane''}). To ensure identifiability, we introduce \emph{uniqueness anchors} (denoted by \(^\star\)), which signal that the marked node, property or edge can be used to produce a descriptor that refers to \emph{exactly one} object in the image. We define uniqueness at
\begin{itemize}
    \item \textbf{node-level,} where the node's type \(\tau(n^\star)\) alone is sufficient to identify the object in the scene (\eg, the `only' sign), 
    \item \textbf{property-level}, where one of its properties \(p^\star\) uniquely identifies the node (\eg, the `only' sign with meaning ``no left turn''), and \item \textbf{edge-level,} \(e^\star\), where a node is uniquely identified through its relation to another node (\eg, the `only' sign that controls the lane.)
\end{itemize}
Descriptors are constructed recursively by iterating over unique anchors, until uniqueness is achieved; if necessary, geometric grounding \(p_t^{\text{geo}}\) is used for disambiguation. Objects that cannot be uniquely identified are excluded from well-defined queries. Further details are provided in \Cref{app:recursive-unique-descriptors}.

\paragraph{(b) Completeness.}
To assure that queries about completeness or cardinality of nodes remain compatible with our open-world assumption, we define a completeness indicator at \emph{node-type level}. For a type \(\tau(n)\), let \(c_t(\tau) \in \{0,1\}
\) denote whether all instances of type \(\tau\) that are visible in the input image at time \(t\) are also present in \(N_t\).
A query \(q\) that requires exhaustive enumeration over nodes of type \(\tau\) is then considered valid only if \(c_t(\tau) = 1\). 
Taken together, the first three components of the CRS framework yield a representation that is structurally well-defined yet remains semantically expressive, while ensuring that objects in the scene can be uniquely and unambiguously associated with individual nodes in the graph. 

\subsection{The query instantiation mechanism}
At the core of CRS is the query instantiation mechanism, which turns the graph representation into an executable framework, hence formulates reasoning tasks as structured computations over the graph. Each query extracts a question, its correct answer, and a set of plausible decoys, all expressed in natural language. Additionally, CoT traces are automatically derived in the same framework, while reasoning depth is controlled as a hyperparameter of the method.  

\paragraph{Query structure.}\label{sec:queries}
Each query is defined by four components: 
(i) a selection operator $\mathcal{S}$ that retrieves a set of target nodes and associated attributes or relations from $G_t$, 
(ii) a question template $\mathcal{T}_q$ that maps the selected elements to a natural language question, 
(iii) an answer template $\mathcal{T}_a$ that renders the correct answer, and
(iv) a perturbation operator $\mathcal{O}(G)$ that produces controlled variations of the graph or its attributes.
To generate hard negatives, applying $\mathcal{T}_a$ to $\mathcal{O}(G)$ yields semantically plausible but incorrect answer candidates.

While a virtually infinite set of queries can be defined to inquire different aspects of the graph, we identify five fundamental categories: queries targeting node properties, inter-node relations, counting and comparison, of nodes, and existence/positional queries. For our study, we instantiate 19 different queries targeting a broad range of graph properties (lane type, line color, counting of lanes and crossings, relationships between lanes, etc.). While all query types used in this study are listed in \Cref{sec:qa-types_supplementary} of the supplementary material, we discuss an example of property-related queries below --- together with edge-related queries the most fundamental operation on the graph. Furthermore, they are relatively simple and hence an important means for model analysis (see \Cref{sec:experiments}). 

\usetikzlibrary{arrows.meta, positioning, shapes.multipart}

Let us assume we have a road scene as illustrated in \Cref{fig:method_figure} with the CRS.
Since property-related queries inquire the value of node properties for nodes of a certain type, our selector can be written as 
\begin{equation}
\mathcal{S}(N_t, \hat{\tau}, \hat{k})=\{n\in N_t, (k^i,y^i(t)) \in P_t(n) \ | \ \tau(n)=\hat{\tau} \ \cap \ k^i =\hat{k}\},
\end{equation}
while the question template is $\mathcal{T}_q=$``What is the <$\hat{k}$> of the $d^*(n)$''? The answer template simply becomes the property value, $\mathcal{T}_a=$``$y(t)$''. If we select $\tau(n)=\texttt{lane}$ and $\hat{k}=\texttt{type}$, $\mathcal{S}$ will extract exactly two matching nodes from the graph (\texttt{Lane-1}, \texttt{Lane-4}), while $\mathcal{T}_a$ yields ``bike lane'' for both. For decoys, $\mathcal{O}$ perturbs the graph by substituting other potential types (``bus lane'', etc.). 
The semantic diversity (and variation) of our framework is then activated through the uniqueness mechanism inside $\mathcal{T}_q$. When resolving for $d^*(n)$ (\texttt{Lane-1} in \Cref{fig:method_figure}), we get a variety of unique options, ranging from \texttt{``What is the type of the lane that contains the bus with number 54D?''} to \texttt{``What is the type of the right curb lane?''}, for the same query and node.  

\paragraph{Reasoning depth and chain-of-thought.}
\label{sec:reasoning}
Because queries refer to objects in the scene through the unique descriptor present in $\mathcal{T}_q$, $\mathcal{T}_a$, or both, the compositional complexity of a query is directly controlled by the maximum descriptor depth, \ie, a hyperparameter that defines how many neighboring nodes can be traversed to derive unique descriptors. Concretely, zero reasoning hops are necessary to resolve \texttt{``the right curb lane''}, while \texttt{``the lane that contains the bus with number 54D''} requires one. The consequences of this explicit dependency are threefold: Not only are semantically diverse multi-hop questions created automatically and with controlled reasoning depth, but a chain-of-thought reasoning trace can also be generated directly from the graph by traversing the descriptor in reverse order and grounding the nodes present in the descriptor chain. While - as for the other parts of the query - individual queries can combine these links in different ways, we identify three primary links in the reasoning chain:
\begin{enumerate}
    \item \textbf{Anchor identification:} We resolve the outermost element of $d^*(n)$ to a unique node $n_a \in N_t$, guaranteed by the uniqueness constraint. This provides an initial grounding in image space via $p_t^{\text{geo}}(n_a)$. For the example in \Cref{fig:method_figure}, this would consist of identifying and grounding \texttt{``the bus with number 54D''}. 
     \item \textbf{Graph traversal:} We iteratively resolve the nested structure of $d^*(n)$ by traversing edges in $G_t$, producing a sequence of nodes $(n_a, \dots, n)$ that terminates at the target node (\ie, \texttt{``the lane that contains the bus with number 54D''}).
     \item \textbf{Target extraction:}  To help ground the target node, we first extract a brief description by combining - unique and non-unique - properties and edges of $n$  to also visually ground the target. Finally, we retrieve the attribute or relation from $n$ needed for the answer.
\end{enumerate}     
     
Each intermediate node in the traversal is grounded via its geometric property, ensuring that the entire reasoning trace is spatially anchored. This formulation yields three key properties. First, the CoT trace is \emph{deterministic}, as it is uniquely defined by the query and the graph. Second, it is \emph{fully grounded}, since each step corresponds to explicit nodes and relations in $G_t$. Third, it is \emph{compositional} and can be potentially repeated multiple times for queries such as counting or comparison, where the procedure is applied once per enumerated element. Importantly, this CoT arises for free alongside question and answer generation, requiring no additional annotation effort. Implementation details about the three links and the specific CoT instantiations per query are provided in \Cref{app:cot-construction}.

\subsection{Data sources and graph construction}
To test the effectiveness of our framework, we instantiate CRS by integrating standard geometric and semantic representations from modular driving stacks, including HD maps (lane geometry and layout), topological annotations (lane connectivity and control relations), and object annotations (dynamic agents and scene elements), and subsequently enrich them with natural language. 
Concretely, we use the front-facing camera views from Argoverse2 \cite{Argoverse2} as visual input, and combine the provided HD maps with the topological structure from OpenLane-V2 \cite{wang2023openlane} to populate our graph. These sources are mapped into CRS using the canonical operators: map elements become nodes ($\Phi_n$), attributes are encoded as properties ($\Phi_p$), and relationships (\eg, adjacency, control) constitute edges ($\Phi_{e}$). 

Although these representations are available and compatible with the CRS, naive aggregation of all information would result in a graph that is overly dense with respect to number of nodes, poorly aligned with the selectiveness of human visual reasoning, and lacking in semantic richness. Rather than directly aggregating all data, we construct graphs through a \emph{selective enrichment} process with human oversight. Starting from the initial graph, expert human annotators 
(i) \emph{filter} for salience by removing irrelevant or non-visible elements,
(ii) \emph{add} entities that are important for scene understanding, including objects beyond existing annotations,
(iii) \emph{enrich} nodes with additional properties in natural language, and
(iv) \emph{introduce} new relations that capture context-specific interactions. This process also establishes uniqueness and completeness required for executable reasoning.

Using this pipeline, we construct 80 CRS graphs, from which we generate 23k training samples via 19 queries on a temporal window of 4 frames at 2~Hz. We also create a held-out validation set of 1000 expert-verified samples from 120 distinct scenes following \cite{lilja2024localization}, subsampling at most one sample per query and scene and computing confidence intervals with scene-level cluster bootstrap, to account for statistical dependence between questions from the same scene. Further details on graph construction, as well as human annotation and quality assurance effort are provided in \Cref{sec:graph_construction_supplementary} and \Cref{sec:stats}. The graph creation tool will be released together with the code and data. 

%% file: sections/4_experiments.tex
\section{Experiments}
\label{sec:experiments}

\pgfplotstableread[col sep=comma]{performance_results.csv}\perfdata

\newcommand{\PlotOneModel}[3]{%
\addplot+[
  ybar,
  bar width=2.5pt,
  draw=black,
  line width=0.1pt,
  fill=#3,
  mark=none,
  solid,
  bar shift=0pt,
  forget plot,
  unbounded coords=discard,
  x filter/.code={
    \pgfplotstablegetelem{\coordindex}{bucket}\of{\perfdata}%
    \edef\thisbucket{\pgfplotsretval}%
    \edef\targetbucket{#1}%
    \pgfplotstablegetelem{\coordindex}{model}\of{\perfdata}%
    \edef\thismodel{\pgfplotsretval}%
    \edef\targetmodel{#2}%
    \ifx\thisbucket\targetbucket
      \ifx\thismodel\targetmodel
      \else
        \def\pgfmathresult{nan}%
      \fi
    \else
      \def\pgfmathresult{nan}%
    \fi
  }
]
table[
  x=model,
  y=accuracy
]{\perfdata};
}

\newcommand{\PlotBucket}[1]{%
\PlotOneModel{#1}{claude-sonnet-4.5_corrected}{m8}
\PlotOneModel{#1}{gemini-3.1-pro_corrected}{m6}
\PlotOneModel{#1}{google_gemini-3.1-flash-lite-preview_corrected}{m9}
\PlotOneModel{#1}{google_gemma-4-31b-it_corrected}{m7}
\PlotOneModel{#1}{moonshotai_kimi-k2.6_corrected}{m14}
\PlotOneModel{#1}{openai_gpt-4o-mini_corrected}{m13}
\PlotOneModel{#1}{openai_gpt-5.4_corrected}{m5}
\PlotOneModel{#1}{qwen-2b-sft-80graphs}{m3}
\PlotOneModel{#1}{qwen-2b-sft-rcot}{m4}
\PlotOneModel{#1}{qwen2b}{m12}
\PlotOneModel{#1}{qwen4b}{m11}
\PlotOneModel{#1}{qwen_4b_sft-80graphs}{m1}
\PlotOneModel{#1}{qwen_4b_sft_cot}{m2}
\PlotOneModel{#1}{qwen_qwen3-vl-235b-a22b-instruct_corrected}{m10}
}
\newcommand{\BucketPlot}[3]{%
\ifnum#3=1
  \def\YAxisLine{left}%
  \def\YTicks{0,1}%
\else
  \def\YAxisLine{none}%
  \def\YTicks{}%
\fi
\begin{tikzpicture}
\begin{axis}[
  width=0.8\linewidth,
  height=1.5cm,
  scale only axis,
  enlargelimits=false,
enlarge x limits=0.1,
  clip=true,
  axis line style={-},
   ymin=0,
  ymax=1,
  xtick=\empty,
  ytick={0,1},
  axis x line*=bottom,
  axis y line*=\YAxisLine,
  symbolic x coords={
  qwen_4b_sft-80graphs,
  qwen_4b_sft_cot,
  qwen-2b-sft-80graphs,
  qwen-2b-sft-rcot,
  spacer,
  openai_gpt-5.4_corrected,
  gemini-3.1-pro_corrected,
  google_gemma-4-31b-it_corrected,
  claude-sonnet-4.5_corrected,
  google_gemini-3.1-flash-lite-preview_corrected,
  qwen_qwen3-vl-235b-a22b-instruct_corrected,
  qwen4b,
  qwen2b,
  openai_gpt-4o-mini_corrected,
  moonshotai_kimi-k2.6_corrected
},
  inner sep=0pt,
  outer sep=0pt,
  tick style={draw=none},
  title={#1},
  title style={
  anchor=north,
  yshift=-5pt,      
  font=\scriptsize,
},
  tick label style={font=\tiny},
  label style={font=\tiny},
]
\PlotBucket{#2}
\end{axis}
\end{tikzpicture}%
}

\newcommand{\BucketText}[3]{%
\begin{minipage}{\linewidth}
\begin{tcolorbox}[
  colback=white!0,
  colframe=white!35,
  arc=0mm,
  left=0mm,
  right=0mm,
  top=0mm,
  bottom=1mm,
  coltext=_gray,
  fontupper=\tiny\linespread{0.75}\selectfont
]
\textbf{#1}
#3
\end{tcolorbox}
\end{minipage}%
}

\newcommand{\BucketColumn}[6]{%
\begin{minipage}[t]{0.2\linewidth}
\centering
\BucketPlot{#1}{#2}{#6}
\BucketText{#3}{#4}{#5}
\end{minipage}%
}

\begin{figure}[t]
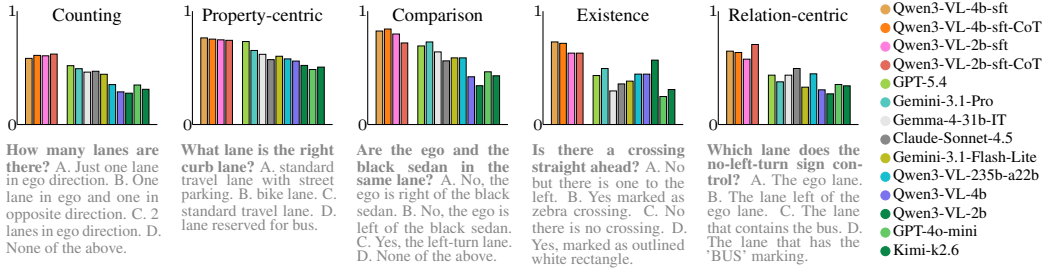

\centering
\noindent
\hspace{-0.5cm}
\begin{minipage}[t]{0.83\textwidth}
\centering

\BucketColumn
  {Counting}
  {counting}
  {How many lanes are there?}
  {A. Just one lane.}
  {A. Just one lane in ego direction. B. One lane in ego and one in opposite direction. C. 2 lanes in ego direction. D. None of the above.}
  {1}%
\BucketColumn
  {Property-centric}
  {properties}
  {What lane is the right curb lane?}
  {C. standard travel lane.}
  {A. standard travel lane with street parking. B. bike lane. C. standard travel lane. D. lane reserved for bus.}
  {0}%
\BucketColumn
  {Comparison}
  {comparison}
  {Are the ego and the black sedan in the same lane?}
  {A. Yes, in the right curb lane.}
  {A. No, the ego is right of the black sedan. B. No, the ego is left of the black sedan. C. Yes, the left-turn lane. D. None of the above.}
  {0}%
\BucketColumn
  {Existence}
  {localization+existance}
  {Is there a crossing straight ahead?}
  {A. CENTER view at \texttt{<box>(192,543,258,547)</box>}.}
  {A. No but there is one to the left. B. Yes marked as zebra crossing. C. No there is no crossing. D. Yes, marked as outlined white rectangle.}
  {0}%
\BucketColumn
  {Relation-centric}
  {relational}
  {Which lane does the no-left-turn sign control?}
  {Replace with correct answer.}
  {A. The ego lane. B. The lane left of the ego lane. C. The lane that contains the bus. D. The lane that has the 'BUS' marking.}
  {0}%
\vspace{-0.2cm}
\end{minipage}%
\begin{minipage}[t]{0.15\textwidth}
\vspace{-55pt}
\begin{tcolorbox}[
  colframe=gray!0,
  colback=white,
  boxrule=0.5pt,
  arc=2pt,
  left=0mm,
  right=0mm,
  top=1mm,
  bottom=1mm
]
\begingroup
\tiny
\newcommand{\LegendItem}[2]{%
  \tikz[baseline=-0.5ex]{
    \draw[#2, fill=#2] (0,0) circle (1.7pt);
  }\,#1%
}
\setlength\tabcolsep{3pt}
\begin{tabular}{@{}*{1}{l@{\hspace{0pt}}}@{}}
  \LegendItem{Qwen3-VL-4b-sft}{m1} \\
  \LegendItem{Qwen3-VL-4b-sft-CoT}{m2} \\
  \LegendItem{Qwen3-VL-2b-sft}{m3} \\
  \LegendItem{Qwen3-VL-2b-sft-CoT}{m4} \\
  \LegendItem{GPT-5.4}{m5} \\
  \LegendItem{Gemini-3.1-Pro}{m6} \\
  \LegendItem{Gemma-4-31b-IT}{m7} \\
  \LegendItem{Claude-Sonnet-4.5}{m8} \\
  \LegendItem{Gemini-3.1-Flash-Lite}{m9} \\
  \LegendItem{Qwen3-VL-235b-a22b}{m10} \\
  \LegendItem{Qwen3-VL-4b}{m11} \\
  \LegendItem{Qwen3-VL-2b}{m12} \\
  \LegendItem{GPT-4o-mini}{m13} \\
  \LegendItem{Kimi-k2.6}{m14} 
\end{tabular}
\endgroup
\end{tcolorbox}
\vspace{-0.5cm}
\end{minipage}

\caption{Model performance by bucket with example questions.} \label{fig:type_spider}
\end{figure}

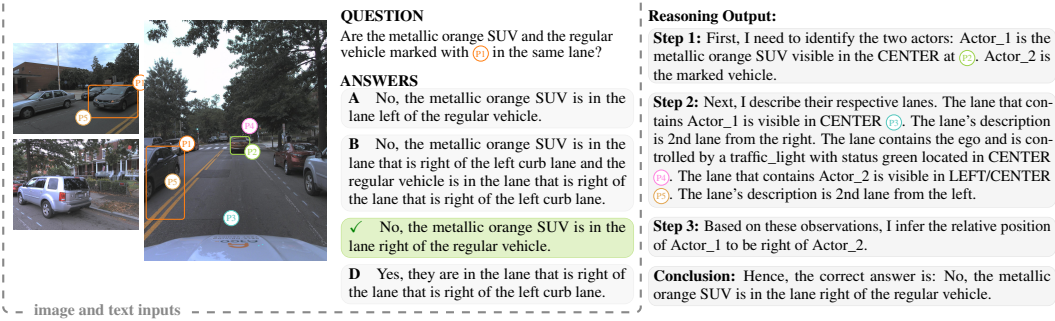
\begin{figure}

\centering

\begin{tikzpicture}

\node[inner sep=0,scale=0.99, transform shape] (wholefig) {

\newcommand{\AnswerPill}[3]{%
  \tcbox[
    answerpill,
    colback=#3!30,
    colframe=#3!60
  ]{%
    \parbox{3.7cm}{\tiny\textbf{#1}\quad #2} 
  }%
  \vspace{0.5mm}
}
\tikzset{
  point/.style={
    circle,
    fill=#1,
    draw=white,
    line width=0.35pt,
    inner sep=1.4pt,
    font=\tiny\bfseries,
    text=white
  }
}

\begin{minipage}[t]{0.3\linewidth}
\vspace{0pt}
\centering
\begin{minipage}[t]{0.4\linewidth}
\centering
\vspace{-2.9cm}
\begin{minipage}{\linewidth}
\centering
\begin{tikzpicture}
\node[inner sep=0, anchor=south west] (img) at (0,0) {
  \includegraphics[width=\linewidth,height=1.2cm]{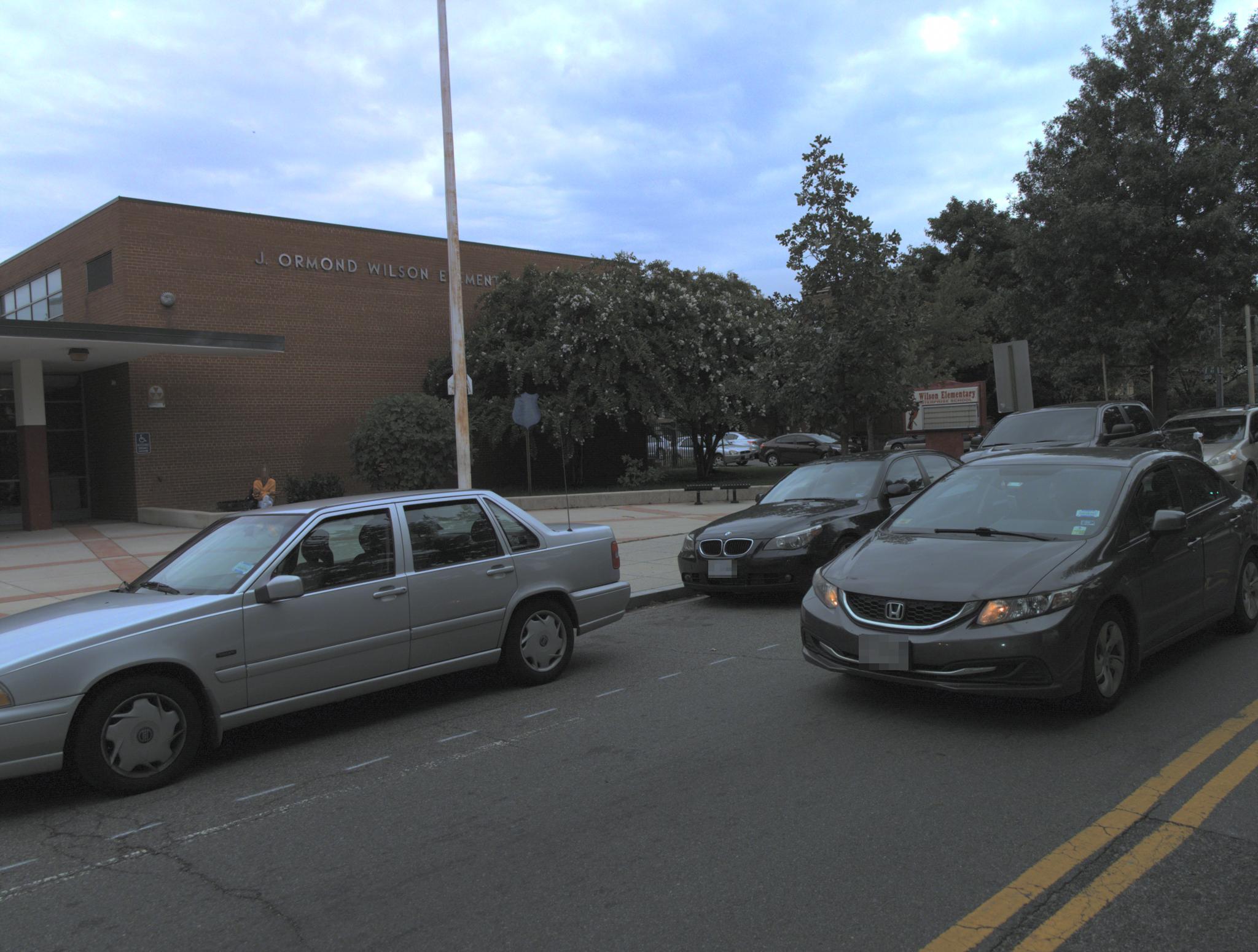}
};
\begin{scope}[x={(img.south east)}, y={(img.north west)}]
\ImgBoxPin
  {0.599}{1-0.824}
  {0.99}{1-0.467}
  {P1}
  {_orange}
  {1pt}
  {1pt}
\ImgPin{0.559}{1-0.823}{P5}{_darkyellow}
\end{scope}
\end{tikzpicture}
\end{minipage}

\vspace{0.5mm}

\begin{minipage}{\linewidth}
\centering
\begin{tikzpicture}
\node[inner sep=0, anchor=north west] (img) at (0,0) {
  \includegraphics[width=\linewidth,height=1.2cm]{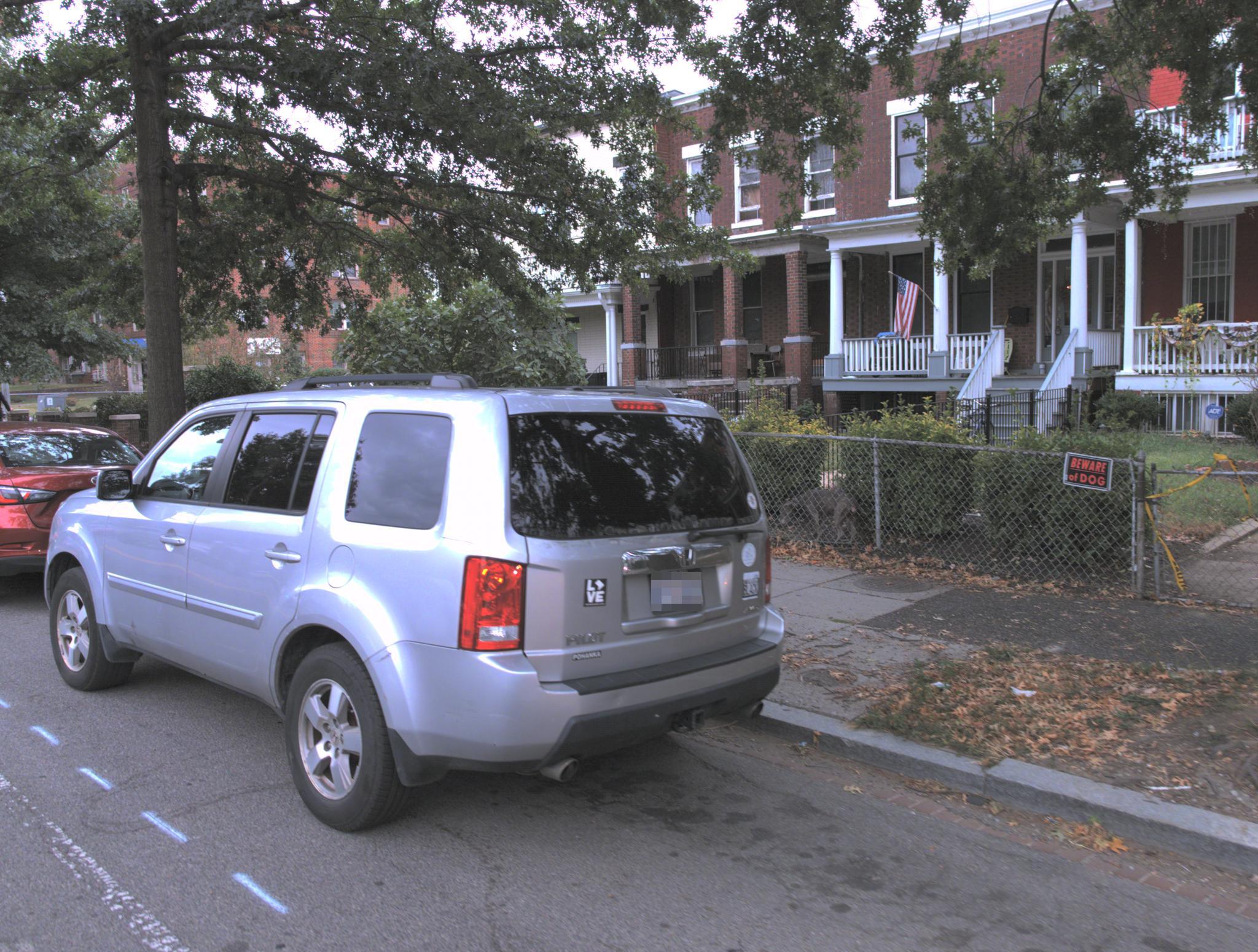}
};
\begin{scope}[x={(img.south east)}, y={(img.north west)}]
\end{scope}
\end{tikzpicture}
\end{minipage}

\end{minipage}
\begin{minipage}[t]{0.58\linewidth}
\centering
\begin{tikzpicture}
\node[inner sep=0, anchor=south west] (img) at (0,0) {
  \includegraphics[width=\linewidth,height=3.2cm]{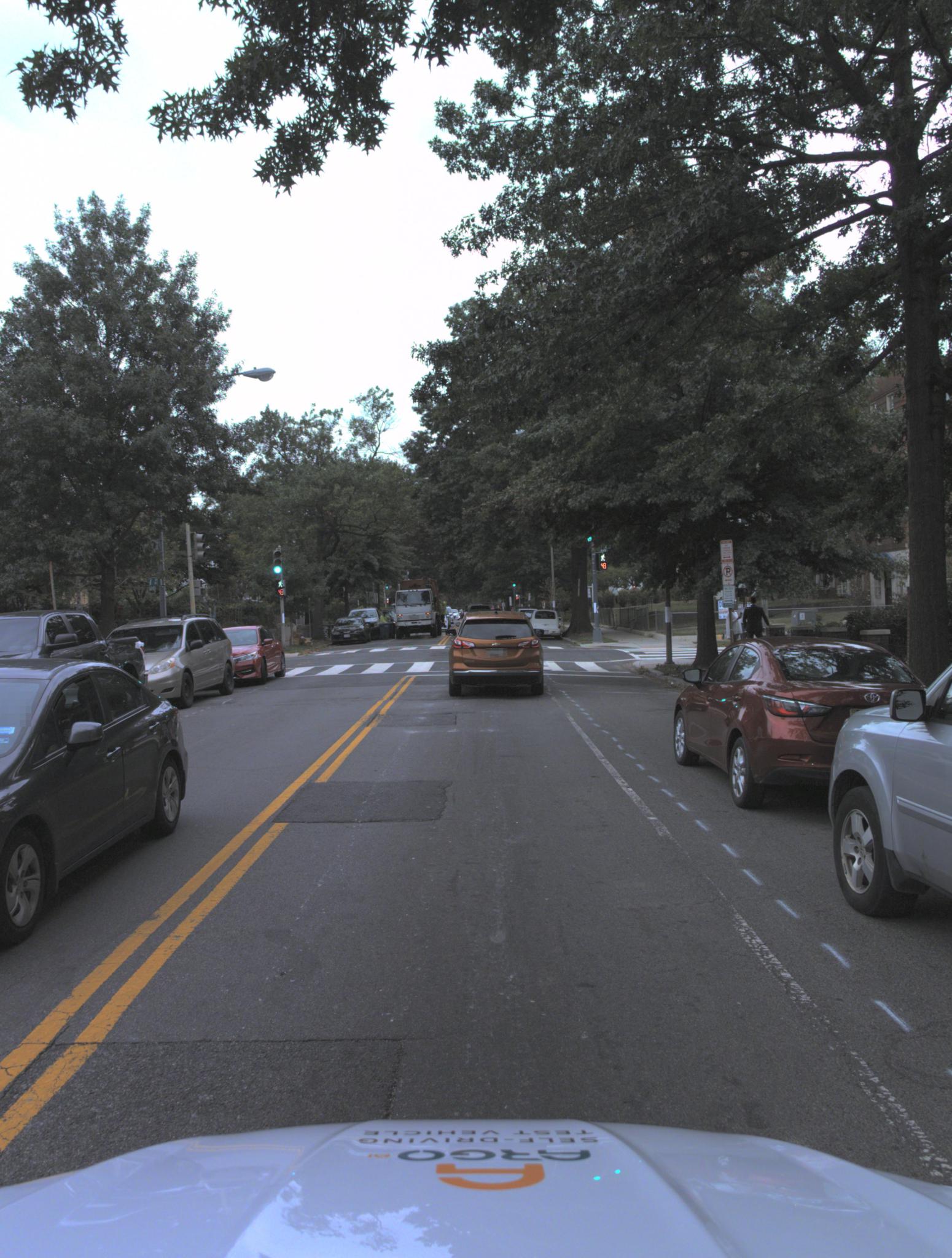}
};
\begin{scope}[x={(img.south east)}, y={(img.north west)}]
\ImgBoxPin
  {0.01}{1-0.827}
  {0.219}{1-0.523}
  {P1}
  {_orange}
  {1pt}
  {1pt}

\ImgBoxPin
  {0.470}{1-0.483}
  {0.576}{1-0.557}
  {P2}
  {_green}
  {1pt}
  {1pt}

\ImgPin{0.475}{1-0.822}{P3}{_teal}

\ImgBoxPin
  {0.552}{1-0.434}
  {0.561}{1-0.450}
  {P4}
  {_pink}
  {1pt}
  {1pt}

\ImgPin{0.154}{1-0.670}{P5}{_darkyellow}
\end{scope}
\end{tikzpicture}
\end{minipage}%

\end{minipage}%
\hspace{0.1cm}
\begin{minipage}[t]{0.26\linewidth}

{\tiny\bfseries QUESTION}\par\par

{\tiny
\parbox{\linewidth}{%
  \baselineskip=6pt
 Are the metallic orange SUV and the regular vehicle marked with \TextPin{P1}{_orange} in the same lane?
}}\par

\tcbset{
  answerpill/.style={
    on line,
    boxrule=0.25pt,
    arc=1.2mm,
    left=1mm,
    right=1mm,
    top=0.4mm,
    bottom=0.4mm,
    boxsep=0pt,
    width=4cm  
  }
}

\vspace{0.1cm}
{\tiny\bfseries ANSWERS}\par
\AnswerPill{A}{No, the metallic orange SUV is in the lane left of the regular vehicle.}{_lightgray}\par
\AnswerPill{B}{No, the metallic orange SUV is in the lane that is right of the left curb lane and the regular vehicle is in the lane that is right of the lane that is right of the left curb lane.}{_lightgray}\par
\AnswerPill{\color{green!50!black}\checkmark}{No, the metallic orange SUV is in the lane right of the regular vehicle.}{_green}\par
\AnswerPill{D}{Yes, they are in the lane that is right of the lane that is right of the left curb lane.}{_lightgray}\par
\end{minipage}
\hspace{0.3cm}
\begin{minipage}[t]{0.39\linewidth}
\tcbset{
  stepbox/.style={
    colback=gray!8,
    colframe=gray!20,
    boxrule=0.25pt,
    arc=1mm,
    left=0.6mm, right=0.6mm,
    top=0.4mm, bottom=0.4mm,
    boxsep=0pt
  }
}

\newcommand{\StepBox}[1]{%
  \begin{tcolorbox}[stepbox]
  {\tiny #1}
  \end{tcolorbox}
}
{\tiny

\textbf{Reasoning Output:}
\vspace{-0.1cm}
\StepBox{\textbf{Step 1:} First, I need to identify the two actors: Actor\_1 is the metallic orange SUV visible in the CENTER at \TextPin{P2}{_green}. Actor\_2 is the marked vehicle.}

\StepBox{\textbf{Step 2:} Next, I describe their respective lanes. The lane that contains Actor\_1 is visible in CENTER \TextPin{P3}{_teal}. The lane's description is 2nd lane from the right. The lane contains the ego and is controlled by a traffic\_light with status green located in CENTER \TextPin{P4}{_pink}. The lane that contains Actor\_2 is visible in LEFT/CENTER \TextPin{P5}{_darkyellow}. The lane's description is 2nd lane from the left.}

\StepBox{\textbf{Step 3:} Based on these observations, I infer the relative position of Actor\_1 to be right of Actor\_2.}

\StepBox{\textbf{Conclusion:} Hence, the correct answer is: No, the metallic orange SUV is in the lane right of the regular vehicle.}
}

\end{minipage}
};


\draw[
    dashed,
    _gray,
    line width=0.8pt,
    rounded corners=2pt
]([xshift=0.5mm,yshift=1.5mm]wholefig.north west) rectangle ([xshift=-5.5cm,yshift=-0.5mm]wholefig.south east);
\node[
    anchor=south west,
    fill=white,
    text=_gray,
    font=\tiny\bfseries,
    inner xsep=1.5mm,
    inner ysep=0.6mm,
    rounded corners=1pt
]at ([xshift=0.3cm,yshift=-2mm]wholefig.south west)
{image and text inputs};
\end{tikzpicture}

\vspace{-0.1cm}
\caption{Example output of the Qwen3-VL-4b model with CoT (additional examples in \Cref{sec:additional_results_qualitative})\vspace{-0.2cm}}

\label{fig:reasoning}

\end{figure}

We evaluate the effectiveness of CRS-based supervision by fine-tuning vision-language models on graph-derived training data and analyzing their behavior across multiple axes. In particular, we focus on (i) overall performance gains, (ii) data efficiency and scaling behavior, and (iii) robustness under increasing reasoning complexity. We fine-tune models from the Qwen family using full-parameter training on 4×H100 GPUs for 5 epochs, with a learning rate of $4 \times 10^{-5}$ and cosine decay. The exact training and evaluation prompt is reported in \Cref{sec:evaluation_protocol}. 

\paragraph{Performance comparison with and without CRS supervision.}
We first evaluate the impact of CRS supervision by comparing models with and without graph-grounded supervision. To contextualize these gains, we additionally evaluate a range of state-of-the-art open- and closed-weight VLMs, which serve as a reference point for current capabilities on structured road reasoning. To disentangle perception from reasoning, we partition the evaluation data into \textit{perception-like} and \textit{reasoning-heavy} samples based on the hardness of decoy answers. In perception-like tasks, the primary challenge lies in identifying the correct object or attribute from the image, after which answer selection is largely straightforward (\eg, ``What is the status of the traffic light that controls the crossing to the right of the intersection?''---``Walk / Stop / Countdown''). In contrast, reasoning-heavy tasks require compositional grounding: models must resolve nested descriptors, identify relevant entities through relational structure, and perform multi-step reasoning  (\eg, ``Which lane is the ego in?''---``The lane that is left of the emergency vehicle's lane. / The lane that contains the arrow. / The right curb lane. / None of the above''). A breakdown of the query templates in each category and details on the question distribution are provided in \Cref{sec:qa-types_supplementary} and \Cref{sec:question-type-analysis}.

As shown in Table~\ref{tab:model_accuracy_detailed}, performance across all models degrades substantially on reasoning-heavy samples, indicating that structured reasoning remains a key challenge. After fine-tuning with CRS-derived data, we observe substantial improvements in overall accuracy and robustness to reasoning complexity. Notably, even small models exhibit strong gains, particularly on reasoning-heavy tasks (see \Cref{fig:type_spider} and \Cref{fig:reasoning}). We further train a baseline model without visual input (cf. \textit{$-$ images} in \Cref{tab:model_accuracy_detailed}) on our CRS-derived data, and find that while the model captures some road structure---because roads inherently contain structure---there remains a significant performance gap compared to vision-augmented models and human performance, highlighting the necessity of visual grounding. A detailed investigation of which aspects of the road the no-image baseline is able to learn, as well as an ablation of the role of the decoys are provided in \Cref{sec:additional_results}.

\input{tables/benchmarks}
\begin{figure}[t]
\centering

\newlength{\fixedfigheight}
\setlength{\fixedfigheight}{0.3cm}

\begin{minipage}[t]{0.48\linewidth}
\centering
\begin{minipage}[t][\fixedfigheight][t]{\linewidth}
\centering
\begin{tikzpicture}
\pgfplotsset{set layers}

\begin{axis}[
    width=0.75\linewidth,
    height=2.7cm,
    scale only axis,
    xlabel={CRS Graphs},
    ylabel={Accuracy},
    xmin=0, xmax=80,
    ymin=0.38, ymax=0.76,
    xtick={0,20,40,60,80},
    tick label style={font=\footnotesize},
    label style={font=\footnotesize},
    every axis plot/.append style={thick, mark size=2pt},
    axis y line*=left,
]
\addplot[name path=upper, draw=none] coordinates {
    (0,0.4753) (20,0.6115) (40,0.6578) (60,0.7140) (80,0.7387)
};
\addplot[name path=lower, draw=none] coordinates {
    (0,0.4122) (20,0.5493) (40,0.5976) (60,0.6588) (80,0.6854)
};

\addplot[name path=upper2b, draw=none] coordinates {
    (0,0.4516) (20,0.6036) (40,0.6410) (60,0.6933) (80,0.7219)
};
\addplot[name path=lower2b, draw=none] coordinates {
    (0,0.3905) (20,0.5404) (40,0.5819) (60,0.6331) (80,0.6617)
};

\addplot[fill=_teal, fill opacity=0.15, draw=none]
fill between[of=upper and lower];

\addplot[fill=_red, fill opacity=0.15, draw=none]
fill between[of=upper2b and lower2b];

\addplot[_teal, thick, mark=*] coordinates {
    (0,0.4428) (20,0.5819) (40,0.6292) (60,0.6864) (80,0.7110)
};

\addplot[_red, thick, mark=*] coordinates {
    (0,0.420) (20,0.5720) (40,0.6114) (60,0.6637) (80,0.6923)
};

\node[
    anchor=south east,
    font=\tiny,
    align=left
] at (axis description cs:0.98,0.02) {
    \tikz{\draw[_red, fill=_red] (0,0) circle (2pt);}~Qwen-2B-SFT\\
    \tikz{\draw[_teal, fill=_teal] (0,0) circle (2pt);}~Qwen-4B-SFT\\
\tikz[baseline={(0,0)}]{\draw[dashed] (0,0.01) -- (2.7,0.01);}~\#Samples
};
\end{axis}

\begin{axis}[
    width=0.75\linewidth,
    height=2.7cm,
    scale only axis,
    xmin=0, xmax=80,
    ymin=0, ymax=24,
    axis y line*=right,
    axis x line=none,
    ylabel={Number of Samples},
    ytick={0,5,10,15,20,25},
    yticklabels={0,5k,10k,15k,20k,25k},
    tick label style={font=\footnotesize},
    label style={font=\footnotesize},
    every axis plot/.append style={thick, mark size=2pt},
]
\addplot[
    gray,
    dashed,
    mark size=1.5pt
] coordinates {
    (0,0) (20,5.740) (40,10.512) (60,16.544) (80,22.833)
};
\end{axis}

\end{tikzpicture}
\end{minipage}
\caption{Accuracy over number of CRS-scenes, with sample counts on the secondary axis.}
\label{fig:scaling}
\end{minipage}
\hfill
\begin{minipage}[t]{0.48\linewidth}
\centering
\begin{minipage}[t][\fixedfigheight][t]{\linewidth}
\centering
\begin{tikzpicture}
\pgfplotsset{set layers}

\begin{axis}[
    width=0.75\linewidth,
    height=2.7cm,
    scale only axis,
    ymin=0.23, ymax=0.88,
    xmin=0, xmax=3,
    xtick={0,1,2,3},
    xticklabels={1,2,3,{$>3$}},
    xlabel={Reasoning depth},
    ylabel={Accuracy},
    clip=false,
    tick label style={font=\footnotesize},
    label style={font=\footnotesize},
    every axis plot/.append style={thick, mark size=2pt},
]

\addplot[name path=qwenCotUpper, draw=none] coordinates {
    (0,0.856271) (1,0.825613) (2,0.865811) (3,0.7860882)
};
\addplot[name path=qwenCotLower, draw=none] coordinates {
    (0,0.752064) (1,0.685581) (2,0.711649) (3,0.6261177)
};
\addplot[fill=_red, fill opacity=0.15, draw=none]
fill between[of=qwenCotUpper and qwenCotLower];

\addplot[name path=qwenUpper, draw=none] coordinates {
    (0,0.860206) (1,0.808710) (2,0.884865) (3,0.778117)
};
\addplot[name path=qwenLower, draw=none] coordinates {
    (0,0.760968) (1,0.653226) (2,0.740712) (3,0.618705)
};
\addplot[fill=_teal, fill opacity=0.15, draw=none]
fill between[of=qwenUpper and qwenLower];

\addplot[name path=gptUpper, draw=none] coordinates {
    (0,0.809525) (1,0.806452) (2,0.702703) (3,0.588235)
};
\addplot[name path=gptLower, draw=none] coordinates {
    (0,0.706350) (1,0.664516) (2,0.522523) (3,0.426471)
};
\addplot[fill=_green, fill opacity=0.15, draw=none]
fill between[of=gptUpper and gptLower];

\addplot[name path=baseUpper, draw=none] coordinates {
    (0,0.591270) (1,0.612903) (2,0.549549) (3,0.455882)
};
\addplot[name path=baseLower, draw=none] coordinates {
    (0,0.468254) (1,0.458065) (2,0.369368) (3,0.294118)
};
\addplot[fill=_orange, fill opacity=0.15, draw=none]
fill between[of=baseUpper and baseLower];

\addplot+[solid, mark=*, color=_red, mark options={fill=_red}]
coordinates {
    (0,0.8030) (1,0.7582) (2,0.7884) (3,0.7057)
};

\addplot+[solid, mark=*, color=_teal, mark options={fill=_teal}]
coordinates {
    (0,0.8174) (1,0.7342) (2,0.8188) (3,0.7027)
};

\addplot+[solid, mark=*, color=_green, mark options={fill=_green}]
coordinates {
    (0,0.7478) (1,0.7482) (2,0.625) (3,0.496)
};

\addplot+[solid, mark=*, color=_orange, mark options={fill=_orange}]
coordinates {
    (0,0.5261) (1,0.5244) (2,0.4711) (3,0.3588)
};

\node[
    anchor=south west,
    font=\tiny,
    align=left
] at (axis description cs:0.02,0.02) {
    \tikz{\draw[_red, fill=_red] (0,0) circle (2pt);}~Qwen-4B-SFT-CoT\\
    \tikz{\draw[_teal, fill=_teal] (0,0) circle (2pt);}~Qwen-4B-SFT\\
    \tikz{\draw[_orange, fill=_orange] (0,0) circle (2pt);}~Qwen-4B\\
    \tikz{\draw[_green, fill=_green] (0,0) circle (2pt);}~GPT-5.4
};
\end{axis}
\end{tikzpicture}
\end{minipage}
\caption{Accuracy by reasoning depth with 95\% CI. Additional results in \Cref{sec:additional_results}.}
\label{fig:accuracy_by_reasoning_depth}
\end{minipage}\vspace{-0.2cm}\end{figure}

\paragraph{Data efficiency and scaling.}
Next, we analyze how performance scales with the number of CRS graphs used for training by varying the number of scenes and measuring downstream accuracy. As shown in \Cref{fig:scaling}, performance improves consistently with additional data, while exhibiting diminishing returns at larger scales. Notably, training on as few as 40 scenes already surpasses the performance of strong closed-source baselines. This highlights the data efficiency of CRS supervision: even in the low-data regime, the compositional structure of graph-derived queries provides dense and informative training signals, while creating a large training set from just a few scenes.

\paragraph{Reasoning depth analysis.}
A central question is whether CRS supervision improves robustness to increasing reasoning complexity. Leveraging the programmatic structure of CRS, we quantify reasoning depth directly from the graph by measuring the number of compositional steps required to resolve a query. While the maximum descriptor depth for individual nodes is limited to two, queries may involve multiple uniquely referenced objects across questions, answers, and distractors, resulting in a broad spectrum of effective reasoning depths. To evaluate the impact of CRS supervision, we partition the evaluation set by reasoning depth, \ie, number of hops on the graph required to solve the task. As shown in \Cref{fig:accuracy_by_reasoning_depth}, models trained with CRS supervision maintain stable performance as reasoning depth increases, in contrast to both the base model and strong closed-source baselines, which exhibit a pronounced degradation (see \Cref{sec:additional_results} for additional results). This indicates that CRS supervision enables models to reliably decompose nested descriptors and execute multi-step relational reasoning over the scene. 

\paragraph{Chain-of-thought analysis.}
Building on this, the CRS framework provides a unique opportunity to trace performance differences back to specific reasoning failures. While CoT supervision does not necessarily improve aggregate accuracy (cf. \Cref{tab:model_accuracy_detailed}), it enables fine-grained analysis of where and how models fail.
For this analysis, we sample 250 questions from the validation set following the property-query schema (see \Cref{sec:queries}). These questions target relatively simple properties---such as lane direction, marking type, or color---but require a well-defined sequence of reasoning steps. Importantly, each question can only be solved by completing the three links in the reasoning chain (cf. \Cref{sec:reasoning}): (i) Localizing the anchor object mentioned in the question, \ie, the last step in a multi-hop descriptor, (ii) resolving the relational description to identify the target node and (iii) retrieving the relevant property. Finally, the last step consists of (iv) choosing the correct option. 

We inspect, for both correct and incorrect predictions, which of these steps are executed correctly. For correct answers, this provides a measure of reasoning consistency: cases where the answer is correct despite incorrect intermediate reasoning indicate spurious correlations. For incorrect answers, the first failed step reveals the origin of the error. 
For correctly answered questions, our model exhibits higher consistency between reasoning and final predictions (\Cref{fig:left_hist}). More importantly, the analysis of incorrect predictions reveals a qualitative difference: in our model, the first failure most often occurs at the property extraction stage, after successfully resolving the object through multi-hop reasoning (\Cref{fig:first_failure_parts}). In contrast, the closed-source model predominantly fails at earlier stages of anchor identification and descriptor decomposition, indicating difficulties in navigating the scene structure in the first place. This suggests that models fine-tuned with CRS-data have internalized the compositional structure of the road scene.

\begin{figure}[t]
\centering

\begin{minipage}[t]{0.48\linewidth}
\centering

\begin{tikzpicture}
\begin{axis}[
    width=5cm,
    height=4cm,
    xbar,
    xmin=0, xmax=24,
    ytick=data,
    symbolic y coords={
    wrong answer selected,
    property not detected,
    deconstruction wrong,
    anchor misidentified
},
yticklabels={
    {\renewcommand{\arraystretch}{0.8}%
     \shortstack[r]{\scriptsize{wrong} \\ \scriptsize{answer selected}}},
    {\renewcommand{\arraystretch}{0.8}%
     \shortstack[r]{\scriptsize{property} \\ \scriptsize{not detected}}},
    {\renewcommand{\arraystretch}{0.8}%
     \shortstack[r]{\scriptsize{deconstruction} \\ \scriptsize{wrong}}},
    {\renewcommand{\arraystretch}{0.8}%
     \shortstack[r]{\scriptsize{anchor} \\ \scriptsize{misidentified}}}
},
    xlabel={Errors in reasoning for correct samples (\%)},
    tick label style={font=\scriptsize},
    label style={font=\scriptsize},
    legend style={
        at={(0.5,1.05)},
        anchor=south,
        legend columns=1,
        font=\tiny,
        draw=none
    },
    bar width=3pt,
    enlarge y limits=0.2,
]

\addplot+[xbar, fill=_green, draw=_green, yshift=0pt]
coordinates {
    (0.00,wrong answer selected)
    (0.71,property not detected)
    (22.3,deconstruction wrong)
    (14.3,anchor misidentified)
};

\addplot+[xbar, fill=_orange, draw=_orange, yshift=0pt]
coordinates {
    (0.00,wrong answer selected)
    (1.2,property not detected)
    (16.40,deconstruction wrong)
    (10.00,anchor misidentified)
};

\addplot+[xbar, fill=_red, draw=_red, yshift=0pt]
coordinates {
    (0.4,wrong answer selected)
    (0.8,property not detected)
    (7.6,deconstruction wrong)
    (6.0,anchor misidentified)
};

\node[
    anchor=south east,
    font=\tiny,
    align=left
] at (axis description cs:0.98,0.02) {
    \tikz[baseline=-0.7ex]{\draw[_red, fill=_red] (0pt,0pt) circle[radius=2pt];}~Qwen-4B-SFT\\
    \tikz[baseline=-0.7ex]{\draw[_orange, fill=_orange] (0pt,0pt) circle[radius=2pt];}~GPT-5.4\\
    \tikz[baseline=-0.7ex]{\draw[_green, fill=_green] (0pt,0pt) circle[radius=2pt];}~Qwen-4B
};
\end{axis}
\end{tikzpicture}

\caption{Distribution of all wrong links in the reasoning chain for samples where the model's final answer is correct.}
\label{fig:left_hist}
\end{minipage}
\hfill
\begin{minipage}[t]{0.48\linewidth}
\centering
\begin{minipage}[t][\fixedfigheight][t]{\linewidth}
\centering
\begin{tikzpicture}

\pgfplotsset{
    /pgfplots/legend image code/.code={
        \draw[
            mark=*,
            mark size=0.5pt,
            only marks,
            draw=none,
            #1
        ] plot coordinates {(0cm,0cm)};
    }
}

\begin{axis}[
    width=5cm,
    height=1.7cm,
    scale only axis,
    xbar stacked,
    xmin=0, xmax=100,
    ytick=data,
    symbolic y coords={
        Qwen-4B-sft,
        GPT-5.4,
        Qwen-4B
    },
    yticklabels={
        Qwen-4B-SFT,
        GPT-5.4,
        Qwen-4B
    },
    xlabel={First failure in reasoning chain for incorrect samples (\%)},
    xtick={0,25,50,75,100},
    tick label style={font=\scriptsize},
    label style={font=\scriptsize},
    legend style={
        at={(0.5,1.0)},
        anchor=south,
        legend columns=2,
        font=\tiny,
        draw=none
    },
    legend cell align=left,
    legend image post style={draw=none},
    bar width=10pt,
    enlarge y limits=0.35,
]

\addplot+[xbar, fill=_red, draw=_red]
coordinates {
    (13.16,Qwen-4B-sft)
    (31.15,GPT-5.4)
    (38.33,Qwen-4B)
};

\addplot+[xbar, fill=_teal, draw=_teal]
coordinates {
    (18.42,Qwen-4B-sft)
    (37.70,GPT-5.4)
    (38.33,Qwen-4B)
};

\addplot+[xbar, fill=_orange, draw=_orange]
coordinates {
    (65.79,Qwen-4B-sft)
    (31.15,GPT-5.4)
    (20.00,Qwen-4B)
};

\addplot+[xbar, fill=_green, draw=_green]
coordinates {
    (2.63,Qwen-4B-sft)
    (0.00,GPT-5.4)
    (3.33,Qwen-4B)
};

\legend{
    anchor misidentified,
    deconstruction wrong,
    property not detected,
    wrong answer selected
}
\end{axis}
\end{tikzpicture}
\end{minipage}
\caption{Distribution of first wrong link in the reasoning chain for samples where the model's final answer is incorrect.}

\label{fig:first_failure_parts}
\end{minipage}
\vspace{-0.3cm}
\end{figure}

\paragraph{Generalization.}
Finally, we consider 3 axes of generalization for our models: generalization (i) across scenes, (ii) across linguistic patterns, and (iii) across different question types. (i) In constructing the validation split used for the experiments, we already test for generalization across different scenes, since our expert-verified samples stem from scenes fully disjoint both geographically and temporally from the training scenes following \cite{lilja2024localization}, with at most one sample per query type per scene to avoid inter-question redundancy. With respect to language and formulation overfitting (ii), we paraphrase 300 validation questions with an LLM (Gemini-2.5-Flash \cite{gemini25}, same graph, same underlying generated questions and answers, reworded surface form) and re-evaluate our Qwen-4B-SFT and Qwen-4B-SFT-CoT models. The results are provided in \Cref{tab:paraphrasing_results} with 95\% CI and McNemar p values, with the modest decrease not being statistically significant, suggesting performance is not tied to CRS's specific template wording. Lastly, to probe (iii) generalization to unseen question types, we retrain the Qwen-4B-SFT model on four of the five query buckets in turn (leaving only isolated question types out would result in very small per-type evaluation samples with low statistical significance): we exclude each bucket entirely from the 23k training samples and evaluate zero-shot on the held-out bucket against the untrained baseline, using McNemar's test on the paired per-item outcomes (\Cref{tab:heldout_results}). We observe promising zero-shot generalization across \texttt{Comparison} and \texttt{Counting} buckets, while \texttt{Relational} and \texttt{Existence} alone are directionally positive (+9.4~pp, +3.7~pp) but not individually significant given the small sample sizes of n=81–85 (p=0.23, p=0.70). The last row is a separate retraining with the three smallest buckets removed simultaneously, evaluated on their union, not an aggregation of the rows above. This is the strictest of our settings: the remaining training set contains no reasoning-heavy query type at all, yet still improves significantly over baseline. The only bucket that shows no generalization is \texttt{Properties}, which we attribute to smaller headroom from a substantially higher baseline. Note that every retrained model sees strictly less training data than the full model, so these gains are conservative.

\begin{table*}[t]
\centering
\caption{Model performance on original formulation and after paraphrasing 300 randomly chosen samples with Gemini-2.5-Flash \cite{gemini25}. Accuracy reported with 95\% CI with McNemar's test.}
\label{tab:paraphrasing_results}
\begin{small}
\begin{tabular}{lccccc}
\toprule
Model & Original Accuracy & Accuracy after paraphrasing & $\Delta$ & p (McNemar) \\
\midrule
QwenVL-4B-SFT & 0.717 {\footnotesize (0.663, 0.767)} & 0.680 {\footnotesize (0.627, 0.733)} & $-$3.67~pp & 0.13 \\
QwenVL-4B-SFT+CoT & 0.720 {\footnotesize (0.670, 0.770)} & 0.683 {\footnotesize (0.630, 0.737)} & $-$3.67~pp & 0.12 \\
\bottomrule
\end{tabular}%
\end{small}
\end{table*}

\begin{table*}[t]
\centering
\caption{Zero-shot performance when training on all but that held-out bucket, compared to the Qwen3-VL-4b-Instruct \cite{Qwen3-VL} baseline. Accuracy reported with 95\% CI with McNemar's test.}
\label{tab:heldout_results}
\begin{small}
\begin{tabular}{lccccc}
\toprule
Held-out bucket & n & Held-out Accuracy & Baseline Accuracy & $\Delta$ & p (McNemar) \\
\midrule
Comparison & 114 & 0.649 {\footnotesize (0.553, 0.737)} & 0.421 {\footnotesize (0.333, 0.509)} & $+$22.81~pp & 0.0001 \\
Counting & 264 & 0.383 {\footnotesize (0.326, 0.439)} & 0.288 {\footnotesize (0.231, 0.345)} & $+$9.47~pp & 0.0187 \\
Relational & 85 & 0.400 {\footnotesize (0.294, 0.506)} & 0.306 {\footnotesize (0.212, 0.400)} & $+$9.41~pp & 0.2295 \\
Existence & 81 & 0.482 {\footnotesize (0.370, 0.593)} & 0.444 {\footnotesize (0.333, 0.556)} & $+$3.70~pp & 0.7011 \\
Properties & 470 & 0.568 {\footnotesize (0.526, 0.613)} & 0.560 {\footnotesize (0.517, 0.602)} & $+$0.85~pp & 0.8312 \\
Rel.+Comp.+Exist. & 280 & 0.532 {\footnotesize (0.475, 0.593)} & 0.393 {\footnotesize (0.332, 0.454)} & $+$13.93~pp & 0.0005 \\
\bottomrule
\end{tabular}%
\end{small}
\vspace{-0.3cm}
\end{table*}

%% file: tables/benchmarks.tex
\begin{table*}[t]
\centering
\caption{Accuracy is reported as decimals with 95\% scene-clustered CI. Accuracy for additional models is provided in \Cref{sec:additional_results}. $^*$We randomly select 300 samples, which are given to 4 human driving experts. }
\label{tab:model_accuracy_detailed}
\begin{small}
\begin{tabular}{l|c|ccc}
\toprule
Model  & Total & Perception-like & Reasoning-heavy \\
\midrule
$\text{Human expert}^*$  & 0.955 {\footnotesize (0.943,0.967)} & - & - \\

\midrule 

\text{GPT-4o-mini} \cite{openai_2024_gpt4omini} & 0.417 {\footnotesize (0.383, 0.450)} & 0.445 {\footnotesize (0.408, 0.480)} & 0.305 {\footnotesize (0.237, 0.374)} \\
\text{Gemini-3.1-Flash-Lite} \cite{google2026gemini31} & 0.519 {\footnotesize (0.485, 0.550)} & 0.554 {\footnotesize (0.520, 0.590)} & 0.375 {\footnotesize (0.314, 0.438)} \\
\text{Claude-Sonnet-4.5} \cite{anthropic2025claudesonnet45} & 0.521 {\footnotesize (0.484, 0.555)} & 0.548 {\footnotesize (0.513, 0.581)} & 0.410 {\footnotesize (0.342, 0.473)} \\
\text{Gemini-3.1-Pro} \cite{google2026gemini31} & 0.584 {\footnotesize (0.549, 0.618)} & 0.609 {\footnotesize (0.571, 0.649)} & 0.480 {\footnotesize (0.415, 0.541)} \\
\text{GPT-5.4} \cite{openai2026gpt54} & 0.621 {\footnotesize (0.593, 0.654)} & 0.667 {\footnotesize (0.636, 0.704)} & 0.435 {\footnotesize (0.374, 0.495)} \\

\midrule
\text{Molmo2-8B} \cite{deitke2025molmo} & 0.405 {\footnotesize (0.378, 0.435)} & 0.410 {\footnotesize (0.378, 0.441)} & 0.385 {\footnotesize (0.327, 0.443)} \\
\text{Kimi-k2.6} \cite{moonshotai2026kimik26} & 0.417 {\footnotesize (0.385, 0.451)} & 0.439 {\footnotesize (0.404, 0.473)} & 0.330 {\footnotesize (0.262, 0.400)} \\
\text{Qwen3-VL-30b-a3b-Thinking} \cite{Qwen3-VL} & 0.460 {\footnotesize (0.422, 0.496)} & 0.484 {\footnotesize (0.445, 0.525)} & 0.360 {\footnotesize (0.296, 0.428)} \\
\text{Qwen3-VL-235b-a22b-Instruct} \cite{Qwen3-VL} & 0.499 {\footnotesize (0.467, 0.531)} & 0.512 {\footnotesize (0.476, 0.546)} & 0.445 {\footnotesize (0.380, 0.505)} \\
\text{Gemma-4-31b-IT} \cite{google_deepmind_2026_gemma4} & 0.539 {\footnotesize (0.502, 0.577)} & 0.576 {\footnotesize (0.536, 0.620)} & 0.390 {\footnotesize (0.325, 0.455)} \\

\midrule
\text{Qwen3-VL-2b-Instruct} \cite{Qwen3-VL} & 0.420 {\footnotesize (0.392, 0.450)} & 0.429 {\footnotesize (0.395, 0.461)} & 0.385 {\footnotesize (0.317, 0.459)} \\
\text{~~$+$ \textit{supervised fine-tune}} &  0.692 {\footnotesize (0.658, 0.727)} & 0.715 {\footnotesize (0.677, 0.753)} & 0.600 {\footnotesize (0.533, 0.670)} \\
\text{~~$+$ \textit{supervised fine-tune} $+$ \textit{CoT}} & 0.696 {\footnotesize (0.664, 0.727)} & 0.715 {\footnotesize (0.680, 0.750)} & 0.620 {\footnotesize (0.556, 0.683)} \\
\text{Qwen3-VL-4b-Instruct} \cite{Qwen3-VL}  & 0.443 {\footnotesize (0.410, 0.476)} & 0.466 {\footnotesize (0.429, 0.501)} & 0.350 {\footnotesize (0.283, 0.427)} \\
\text{~~$+$ \textit{supervised fine-tune} ($-$ images)} & 0.587 {\footnotesize (0.549, 0.625)}& 0.586 {\footnotesize (0.545, 0.630)} & 0.590 {\footnotesize (0.518, 0.656)} \\
\text{~~$+$ \textit{supervised fine-tune}}  & \textit{0.711} {\footnotesize (0.672, 0.746)} & \textit{0.718} {\footnotesize (0.679, 0.759)} & \textbf{0.680} {\footnotesize (0.605, 0.751)} \\
\text{~~$+$ \textit{supervised fine-tune} $+$ \textit{CoT}}  &\textbf{ 0.713} {\footnotesize (0.681, 0.746)} & \textbf{0.722} {\footnotesize (0.683, 0.759)} & \textit{0.675} {\footnotesize (0.616, 0.734)} \\
\bottomrule
\end{tabular}
\end{small}
\end{table*}

%% file: sections/5_conclusion.tex
\section{Discussion and conclusion}
\label{sec:conclusion}
This work introduces the Combined Road Substrate (CRS), a graph-grounded framework that unifies geometric structure and open-vocabulary semantics in a jointly executable representation. By formulating reasoning as queries over this substrate, CRS enables the automatic generation of grounded, compositional supervision signals, yielding substantial gains in structured road reasoning even in low-data regimes. Our analysis further demonstrates that structured supervision fundamentally shifts model behavior, redirecting failures from relational reasoning to localized perceptual errors.

\paragraph{Limitations and future work.} 
While CRS improves structural reasoning, performance remains bounded by the quality of underlying visual perception, indicating that further gains depend on advances in perception models. 
CRS construction relies on selective human curation by design, which adds annotation overhead, and assumes HD-map-like topology; automating abstraction and enrichment is future work. Our data stem from a single source dataset with limited geographic diversity, and training and evaluation share one query grammar; while held-out question types and paraphrases show no collapse, models may still overfit to the CRS schema. Reasoning traces may look interpretable without being faithful and must not be taken as safety evidence. Lastly, while this work focuses on the substrate of road understanding, we do not explicitly tackle downstream driving performance. However, we believe that CRS provides a promising grounded intermediate representation for future vision-language-action (VLA) driving systems, while executable and compositional structure may offer a natural interface between semantic reasoning and policy learning — an important direction left for future work.

%% file: sections/6_acknowledgements.tex
\paragraph{Acknowledgements and Disclosures.}
The research work was partially funded by the Swedish Foundation for Strategic Research (SSF) under the project DeltaMap (ID22-0045) and by a gift from Volkswagen Group of America. This work was partially supported by the Wallenberg AI, Autonomous Systems and Software Program
(WASP) funded by the Knut and Alice Wallenberg Foundation. This material is based upon work supported by the Defense Advanced Research Projects Agency (DARPA) under Agreement Number 00011869.
Any opinions, findings, and conclusions or recommendations expressed in this material are those of the author(s) and do not necessarily reflect the views of the Defense Advanced Research Projects Agency (DARPA).
We thank Patric Jensfelt, Rafael Valencia and Vahdat Abdelzad for their feedback on this work.

%% file: sections/supplementary_material.tex
\section{Supplementary Material}
\label{sec:supplementary}
\subsection{Query instantiation}\label{sec:qa-types_supplementary}
Let \(\Gt=(N_t,E_t)\) denote the CRS scene graph at the queried
frame and let \(\Gwin\) denote the temporal window of size $w=4$ used by temporal queries. \(\tau(n)\) is the node type, \(p_n=(k^i, y^i(t))\) is property \(k^i\) of node \(n\) with value \(y^i(t)\), and \((n,n',\ell_{n,n'}(t))\in E_t\) denotes a directed relation. \(\desc(n)\) is the recursive
unique descriptor selected by the recursive descriptor construction. \Cref{tab:cot-templates} below details the 19 queries used in this work.

\scriptsize
\captionsetup{font=normalsize}
\setlength{\tabcolsep}{3pt}
\renewcommand{\arraystretch}{1.18}

\begin{longtable}{p{0.115\linewidth} p{0.235\linewidth} p{0.18\linewidth} p{0.19\linewidth} p{0.235\linewidth}}
\caption{High-level formalization of all query types instantiated in the CRS.}\label{tab:cot-templates}\\
\toprule
\textbf{Type} &
\textbf{Selection operator \(\Sel(\Gt)\)} &
\textbf{Question template \(\Tq\)} &
\textbf{Answer template \(\Ta\)} &
\textbf{Perturbation procedure \(\Pert(\Gt)\)}\\
\midrule
\endfirsthead

\toprule
\textbf{Type} &
\textbf{Selection operator \(\Sel(\Gt)\)} &
\textbf{Question template \(\Tq\)} &
\textbf{Answer template \(\Ta\)} &
\textbf{Graph perturbation / decoy procedure \(\Pert(\Gt)\)}\\
\midrule
\endhead

\bottomrule
\endfoot

\texttt{lane direction} &
Select each visible lane \(l\) with \(\tau(l)=\type{lane}\), excluding lanes whose descriptor would already reveal the direction or a near-intersection turn-side relation. Return \((l,p_l[\prop{direction}])\). &
``What is the direction of \(\desc^*(l)\)?'' &
Render \(p_l[\prop{direction}]\) as either ``same direction as ego'' or ``opposite direction''. &
Attribute perturbation: replace the true binary direction with the other direction value. Distractors are the alternate direction labels.\\
\midrule

\type{line marking} &
Select lane-line/lane pairs \((m,l)\) with \(\tau(m)=\type{lane\_line}\), \(\tau(l)=\type{lane}\), and \((m,l,\rel{is left marking of})\) or \((m,l,\rel{is right marking of})\in E_t\). &
``What is the left/right lane line marking of \(\desc^*(l)\)?'' &
Render the full marking style \(p_m[\prop{mark\_type}]\), \eg the verbose form of a dashed/solid and color combination. &
Attribute perturbation: substitute another complete lane-marking style from the marking vocabulary; optionally include ``None of the above''.\\
\midrule

\type{line type} &
Use the same selected lane-line/lane pair \((m,l)\) as \type{line\_marking}. &
``What is the left/right lane line type of \(\desc^*(l)\)?'' &
Render the type component of \(p_m[\prop{mark\_type}]\), \eg solid, dashed, double solid, double dashed, or no visible line. &
Attribute perturbation: hold the lane and side fixed but replace the type component with another plausible line type; optionally include ``None of the above''.\\
\midrule

\type{line color} &
Use the same selected lane-line/lane pair \((m,l)\) as \type{line\_marking}. &
``What is the left/right lane line color of \(\desc^*(l)\)?'' &
Render the color component of \(p_m[\prop{mark\_type}]\), \eg, white or yellow, not visible if it is implicit. &
Attribute perturbation: hold the lane and side fixed but replace the color component with another color from the lane-line vocabulary.\\
\midrule

\texttt{existence of crossings} &
For a single nearby intersection, iterate over canonical crossing locations \(s\in\{\)left of , right of, straight ahead on, at far side of\(\}\) the intersection. Select all described crossing nodes \(c\) and check whether any visible \(c\) satisfies \(p_c[\prop{description}]=s\). &
``Is there a crossing \(s\) of/on the intersection?'' &
If present, answer either with the crossing marking style or ``Yes, there is a crossing \(s\).'' If absent, answer ``No, there is no crossing \(s\)'', sometimes naming another present side. &
Existence and spatial perturbation: deny an existing crossing, assert existence at a wrong side, substitute the wrong crossing style, or conflate absence at \(s\) with presence elsewhere; optionally include ``None of the above''.\\
\midrule

\type{lane type} &
Select each visible lane \(l\) with \(\tau(l)=\type{lane}\). If an intersection is present, restrict to lanes whose lane-intersection relation is relevant for the ego-intersection distance. Return \((l,p_l[\prop{type}])\). &
``What lane is \(\desc^*(l)\)?'' &
Render the verbose lane class for \(p_l[\prop{type}]\), \eg, standard vehicle lane, turn lane, bike-shared lane, etc.; sometimes answer ``None of the above'' when the true class is withheld from the options. &
Attribute perturbation: replace the true lane class with other classes from the lane-type vocabulary while keeping the selected lane fixed; optional ``None of the above''.\\
\midrule

\texttt{crossing type} &
Select each visible crossing \(c\) near a single intersection with a known \(p_c[\prop{marking\_style}]\). &
``What is the style of \(\desc^*(c)\)?''; ``How is \(\desc^*(c)\) marked?''; or ``What is the marking of \(\desc^*(c)\)?'' &
Render the pedestrian-crossing style vocabulary value for \(p_c[\prop{marking\_style}]\). &
Attribute perturbation: replace the true marking style with other pedestrian-crossing style values; optionally include ``None of the above''.\\
\midrule

\texttt{traffic light status} &
Select each visible traffic light \(z\) with \(\tau(z)=\type{traffic light}\). For vehicle signals, collapse duplicate traffic lights that control the same lane set; Return \((z,p_z[\prop{status}])\). &
``What is the status of \(\desc^*(z)\)?'' &
Render the current status in the last frame as a traffic-light color or as the corresponding pedestrian command. &
Attribute perturbation: replace the true status with other legal signal states or pedestrian commands, using the control relation to choose the relevant vocabulary; optionally include ``None of the above''.\\
\midrule

\texttt{traffic light change} &
From \(\Gwin\), select traffic lights \(z\) with \(\tau(z)=\type{traffic light}\) where \prop{status} is known for the full window $w$, \((p_z^0[\prop{status}],\ldots,p_z^t[\prop{status}])\). &
``Did the status of \(\desc^*(z)\) change over the last \(w\) frame(s)?'' &
If the first and last states differ, render ``Yes, it was \(a\) first but switched to \(b\).'' Otherwise render ``No, it stayed \(a\).'' Pedestrian signals use pedestrian-command labels. &
Temporal perturbation: create false transition patterns by changing the initial state, final state, or change/no-change claim; also create false constant-status answers and optional ``None of the above''.\\
\midrule

\texttt{counting at intersection per direction} &
For a single intersection, select lane-intersection relation pairs \((r,\bar r)\) such as \prop{approaches to the left of} / \prop{leaves to the left of}. Count lanes \(l\) with \((l,I,r)\) or \((l, I,\bar r)\in E_t\), subject to the ego-intersection distance. &
``How many lanes are there \(\mathrm{phrase}(r,\bar r)\)?'' &
Render the total count and directional breakdown, \eg lanes travelling away from versus towards the camera for the selected intersection side or road segment. &
Count perturbation: keep the queried relation pair fixed but perturb the two relation-specific counts by nearby integers; optionally replace one option with ``None of the above''.\\
\midrule

\texttt{counting per direction} &
Select all visible lanes when the scene is off-intersection, or lanes related by \prop{leads up to} / \prop{leaves} when an intersection is far enough away. Count them by \(p_l[\prop{direction}]\in\{\type{same},\type{opposite}\}\). &
``How many lanes are there?'' &
Render the total number of lanes and the breakdown into ego-direction and opposite-direction lanes. &
Count perturbation: perturb the same/opposite direction count pair by nearby integers while keeping the road extent fixed; optionally include ``None of the above''.\\
\midrule

\texttt{counting generic} &
For each direction \(d\in\{\type{same},\type{opposite}\}\), select visible lanes \(l\) satisfying \(p_l[\prop{direction}]=d\), restricted to the off-intersection road extent when needed. &
``How many lanes are there in ego direction?'' or ``How many lanes are there in direction opposite to the ego?'' &
Render a scalar count sentence, \eg ``There are \(k\) lanes in ego direction'' or ``There are no lanes travelling towards the camera''. &
Scalar count perturbation: replace the true count \(k\) with nearby nonnegative integers, with one-way constraints for ego-direction counts; optionally include ``None of the above''.\\
\midrule

\type{pointing} &
Select each visible localizable node \(n\) which has a defined position in $\mathbb{R}^2$. For lanes and crossings, require another candidate of the same type; for signs, lights, markings, cones, barrels, racks, bollards, and dividers, use candidates from the same broad visual class. &
``Where is \(\desc^*(v)\)?'' &
Render the point or bounding box extracted for \(n\)'s visual localization. &
Identity/location perturbation: replace the target marker with markers for visually similar or same-class nodes at different identities or locations.\\
\midrule

\texttt{counting crossing} &
Select all visible crossing nodes \(c\) with \(\tau(c)=\type{crossing}\). If each crossing has a canonical location description, count by location (\eg, at the far side of the intersection, left of the intersection); otherwise count the crossings generically.  &
``How many pedestrian crossings are there?'' &
Render the total and, when available, the location layout, \eg crossings to the left/right/straight ahead/far side. &
Count/layout perturbation: perturb the total number of crossings or replace the true set of crossing locations with another canonical combination; optionally include ``None of the above''.\\
\midrule

\texttt{vehicle position} &
Select each actor \(a\) (ego or annotated vehicle actor, such as \type{bus}, \type{truck}, \type{emergency vehicle}) with a unique lane \(l\) such that \((a,l,\rel{is in})\in E_t\) or \((l,a,\rel{contains})\in E_t\). &
``Which lane is \(\desc^*(a)\) in?'' &
Render the descriptor \(\desc^*(l)\) of the lane containing the actor. &
Relation perturbation: replace the true containing lane with adjacent left/right lanes, lanes containing other actors, or other visible lanes. The true containment edge is hidden during descriptor construction so the answer is not leaked.\\
\midrule

\texttt{sign controls lane} &
Select each sign \(s\) with a nonempty controlled lane set \(L_s=\{l:(s,l,\rel{controls})\in E_t\}\). &
``Which lane(s) does \(\desc^*(s)\) control?'' &
Render the joined descriptors of all lanes in \(L_s\). &
Controlled-set perturbation: substitute graph-near lane sets built from adjacent lanes, lanes containing vehicles, and other visible lanes; vary set cardinality as well as identity. Control true edges are hidden during descriptor construction.\\
\midrule

\texttt{traffic light controls lane} &
Select each traffic light \(z\) with a nonempty controlled lane set \(L_z=\{l:(z,l,\rel{controls})\in E_t\}\). &
``How many lanes does \(\desc^*(z)\) control?'' &
Render ``\(|L_z|,\) \(\desc^*(l_1)\) and \(\desc^*(l_2)\ldots\)'', \ie the number and descriptors of controlled lanes. &
Controlled-set perturbation: substitute adjacent or otherwise plausible lane sets, including sets with the wrong cardinality, while preserving road-local plausibility. Control edges are hidden during descriptor construction.\\
\midrule

\texttt{pairwise vehicle location} &
Select unordered pairs of visible actors \((a_1,a_2)\), including ego, where each actor has exactly one containing lane \(l_1,l_2\). Determine whether \(l_1=l_2\) and, if not, whether \(l_1\) is left/right of \(l_2\). &
``Are \(\desc^*(a_1)\) and \(\desc^*(a_2)\) in the same lane?'' &
If \(l_1=l_2\), render ``Yes, they are in \(\desc^*(l_1)\).'' Otherwise render a ``No'' answer with the two lane descriptors or the left/right relation. &
Comparison perturbation: flip the same-lane claim, reverse the left/right relation, or substitute one/both actor lanes with adjacent or common decoy lanes. \\
\midrule

\texttt{pairwise lane comparison by direction} &
Select unordered pairs of visible lanes \((l_1,l_2)\), excluding descriptors or intersection-side relations that would make the comparison trivial. Return \(p_{l_1}[\prop{direction}]\) and \(p_{l_2}[\prop{direction}]\). &
``Do \(\desc^*(l_1)\) and \(\desc^*(l_2)\) share the same direction?'' &
Render one of four templates: yes, both in ego direction; yes, both opposite ego; no, first ego and second opposite; no, first opposite and second ego. &
Comparison perturbation: substitute the incorrect same/different direction templates, including reversed first/second assignments; optionally include ``None of the above''.\\

\end{longtable}
\normalsize
\subsection{Construction of chain-of-thought traces}
\label{app:cot-construction}

This appendix details how the Chain-of-Thought (CoT) traces are constructed from the CRS graph in the implementation. The generator does not annotate reasoning traces separately. Instead, each trace is derived from the same query instantiation that produces the question, answer, and decoy options.

Let a query be written as
\[
q = (S,T_q,T_a,O),
\]
where \(S\) selects graph elements, \(T_q\) renders the question, \(T_a\) renders the answer, and \(O\) defines perturbations used for hard negatives. For a selected target node \(n\), the generator first samples a unique descriptor \(d^\star(n)\). Concretely, each descriptor stores its rendered description text in natural language, the target node, reasoning depth, its uniqueness anchor with respect to property or relation, and a list of dependencies
\[
\Delta(d^\star) =
\{(u_i,\rho_i,v_i,h_i)\}_{i=1}^m ,
\]
where \(u_i\) is an intermediate node that must be resolved, \(\rho_i\) is the relation used to connect it to the downstream node \(v_i\), and \(h_i\) is the hop depth. Descriptors can be obtained from four high-level sources: one of the three uniqueness anchors (unique node type, unique property, recursively unique edge), or a geometric point marker. If no unique descriptor is available, the query is rejected unless the template explicitly allows non-unique enumeration (for example, in counting queries).

The resulting CoT is the concatenation of three graph-derived links:
\[
C(q) =
C_{\mathrm{anchor}}(q)
\circ C_{\mathrm{traverse}}(q)
\circ C_{\mathrm{extract}}(q).
\]
The exact surface form varies by template, but the underlying computation is shared.

\paragraph{Link 1: Anchor identification.}
The first link grounds the initial object(s) from which the rest of the reasoning can proceed. For descriptor-based queries, the implementation sorts the dependencies in \(\Delta(d^\star)\) by hop depth and begins with the innermost uniquely identifiable node. If the descriptor has no dependencies, the target node itself is the anchor. For a node \(u\), grounding is rendered from its geometric property
\(p^{\mathrm{geo}}_t(u)\), which becomes a camera-view mention with a grounding point or box.

\newcommand{\steplabel}[1]{\makebox[4em][l]{\texttt{Step #1:}}}

Operationally, this link emits statements of the form: 
\begin{equation}
\begin{aligned}
\steplabel{1} & \texttt{Identify the <} d^*(u) \texttt{>, which is visible in the} \\
                  & \texttt{LEFT/CENTER/RIGHT view at <point>}(x,y)\texttt{</point>/} \\
                  & \texttt{<box>}(x_1,y_1, x_2, y_2)\texttt{</box>.}
\end{aligned}
\end{equation}
This identifies the object named by the unique descriptor, then states where it is visible. 

Point-based descriptors first identify the marker and then the object marked by it. 
\begin{equation}
\begin{aligned}
\steplabel{1}  & \texttt{Identify the location of <point\_1> in the last } \\
& \texttt{frame. The marker is visible in the} \\
                  & \texttt{LEFT/CENTER/RIGHT view at <point>}(x,y)\texttt{</point>}  \\
                  & \texttt{and marks a <}d(u)\texttt{>.},
\end{aligned}
\end{equation}
with $d(u)$ being a description of the marked object which is not necessarily unique. In counting and existence templates, the selection operator \(S\) provides a set of anchors rather than a single descriptor; each counted lane or crossing is introduced as an enumerated object and grounded in the image.

\paragraph{Link 2: Graph traversal.}
The second link resolves the nested relational part of the descriptor. If a descriptor depends on a chain of nodes
\[
u_0 \stackrel{\rho_1}{\longrightarrow} u_1
\stackrel{\rho_2}{\longrightarrow} \cdots
\stackrel{\rho_n}{\longrightarrow} n,
\]
the trace walks this chain from the already-grounded anchor \(u_0\) to the target \(n\). Each step instantiates the canonical relation statement in language, until reaching the target node, for example 
\begin{equation}
\begin{aligned}
\steplabel{2}  & \texttt{The } \tau(n) \texttt{ in question is the } \tau(n) \texttt{ that <} \rho_n \texttt{>}\\
& \texttt{that <} \tau(u_{n-1}) \texttt{>,}  \texttt{and is visible in the LEFT/CENTER/} \\ & \texttt{RIGHT view at <point>}(x,y)\texttt{</point>/}\\
                  & \texttt{<box>}(x_1,y_1, x_2, y_2)\texttt{</box>.}
\end{aligned}
\end{equation}

For templates with more than one relevant object, this traversal is repeated. Comparison templates resolve both compared objects, relation templates resolve both the controller and the controlled lane set, and option-verification templates additionally resolve candidate answer descriptors before comparing their node identities to the target node identities. Counting templates use a related pattern: each selected item is introduced, optionally related to previously introduced lanes through adjacency, and then added to the running enumeration.

\paragraph{Link 3: Target extraction.}
The final link extracts the answer-bearing quantity from the resolved target node, a linked answer node, or a target set. Before revealing the answer, most templates add a short auxiliary description of the target by stating both its unique and non-unique edges and properties, to anchor it linguistically in the scene. 

More concretely, for the resolved target node, the generator forms a local evidence set by selecting up to \(i\) properties of the node and up to \(j\) outgoing relations to neighboring nodes. For each selected relation, the neighboring node is itself summarized by up to \(l\) of its properties. The candidate properties and relations are first randomized and then ordered by a predefined priority hierarchy for the corresponding node type, so the rendered evidence is varied while still preferring semantically useful facts. These selected facts are finally converted into natural-language sentences using the canonical property and relation forms. For $i=j=l=1$, this yields:
\begin{equation}
\begin{aligned}
\steplabel{3}  & \texttt{The <}\tau(n)\texttt{>'s <}k^i\texttt{> is <}y^i(t)\texttt{>. The <}\tau(n)\texttt{> <}\ell_{n,n_j}(t)\texttt{>}\\ & \texttt{<}\tau(n_j)\texttt{>}  \texttt{ with <}k^{l}\texttt{> <}y^{l}(t)\texttt{>.}
\end{aligned}
\end{equation}

The extracted target quantity depends on the query family. For property queries it is a property value, such as lane type, lane direction, line color, crossing marking style, or traffic-light status; for lane-line queries, the answer-bearing node is reached through the lane-marking relation. For relation queries it is a set of target lanes connected by a relation such as \texttt{contains} or \texttt{controls}. For counting queries it is the cardinality, sometimes decomposed by direction or crossing location. For comparison queries it is an equality or relative-position judgment. For temporal queries it is a sequence of property values across frames. For pointing queries it is the option whose marker corresponds to the resolved target node. The extracted value is rendered through \(T_a\), compared to the answer options, and followed by a conclusion selecting the correct answer.

Table~\ref{tab:cot-template-links} summarizes how different queries used in this study (cf. \Cref{tab:cot-templates}) instantiate the three links.
\scriptsize
\captionsetup{font=normalsize}
\setlength{\tabcolsep}{3pt}
\renewcommand{\arraystretch}{1.18}

\begin{longtable}{p{0.16\linewidth} p{0.255\linewidth} p{0.255\linewidth} p{0.255\linewidth}}
\caption{CoT construction logic for query families.}
\label{tab:cot-template-links}\\

\toprule
\textbf{Query family} &
\textbf{Anchor identification $C_{\mathrm{anchor}}(q)$} &
\textbf{Graph traversal} $C_{\mathrm{traverse}}(q)$&
\textbf{Target extraction $C_{\mathrm{extract}}(q)$} \\
\midrule
\endfirsthead

\toprule
\textbf{Query family} &
\textbf{Anchor identification $C_{\mathrm{anchor}}(q)$} &
\textbf{Graph traversal} $C_{\mathrm{traverse}}(q)$&
\textbf{Target extraction $C_{\mathrm{extract}}(q)$} \\
\midrule
\endhead

\bottomrule
\endfoot

\textbf{Property-related}
(\texttt{lane direction},
\texttt{lane type},
\texttt{line marking},
\texttt{line type},
\texttt{line color},
\texttt{crossing type},
\texttt{traffic light status})
&
Resolve the descriptor of the queried lane, crossing, lane line, or traffic light and ground it in the camera views.
&
Follow any nested descriptor dependencies, such as lane adjacency, containment, marking, or control relations, until the queried target node is identified.
&
Describe auxiliary non-answer properties or edges, then read the requested property from the target node or linked answer node. Traffic-light status additionally checks the latest frame and may cross-verify earlier frames.
\\
\midrule

\textbf{Temporal property changes}
(\texttt{traffic light change})
&
Resolve and ground the traffic light directly or the lane/crossing controlled by it.
&
If several lights control the same target, enumerate their visible positions; otherwise follow the descriptor to the unique signal.
&
Read the status timeline over the frame window and answer whether the value stayed fixed or changed from the first to the last frame.
\\
\midrule

\textbf{Relations}
(\texttt{vehicle position},
\texttt{sign controls lane},
\texttt{traffic light controls lane})
&
Identify the actor, sign, or traffic light that appears in the question.
&
Traverse \texttt{is in}/\texttt{contains} or \texttt{controls}/\texttt{is controlled by} edges to obtain the target lane or lane set. Candidate options are resolved by their own descriptors.
&
Compare option node identities with the target lane identities and select the option that describes the correct lane or lane set.
\\
\midrule

\textbf{Counting }
(\texttt{counting at intersection per direction},
\texttt{counting per direction},
\texttt{counting generic},
\texttt{counting crossing})
&
Set the counting scope, usually the most recent frame and optionally the relevant intersection region, then introduce each selected lane or crossing as a grounded item.
&
For each item, resolve a short descriptor and, when useful, relate it to the ego lane, an intersection, or previously enumerated neighboring lanes.
&
Aggregate the enumerated items into totals, optionally split by traffic direction or crossing location, and eliminate/count-match the answer options.
\\
\midrule

\textbf{Comparison}
(\texttt{pairwise vehicle location},
\texttt{pairwise lane comparison by direction})
&
Identify and ground both compared actors or lanes.
&
Resolve the lane containing each actor, or the descriptor of each lane, then compare the resulting graph nodes or their direction properties.
&
Return whether the objects share a lane or direction, or state the relative lane position when they differ.
\\
\midrule

\textbf{Existence and pointing}
(\texttt{existence of crossings},
\texttt{pointing})
&
For crossing existence, anchor the requested side of the intersection. For pointing, anchor the object descriptor in the question.
&
Search the selected crossing set for the requested location, or compare each marker option to the target node.
&
Answer existence plus marking style when available, or select the marker option whose node is the target.
\\

\end{longtable}

\normalsize

\subsection{Recursive construction of unique node descriptors}
\label{app:recursive-unique-descriptors}

This appendix describes the descriptor recursion used to turn the uniqueness
signals in the CRS graph into natural-language object references. For a target
node \(n\), the procedure returns a set \(D(n)\) of candidate descriptors \(d(n)\).
Each candidate stores its rendered text, the target node, whether the reference is unique, the number of relational hops, and the intermediate nodes that must be resolved when the descriptor is later verbalized as a reasoning trace.

Let \(p_t(n)\) denote the property dictionary of \(n\), let \(\tau(n)\) denote its node type, and let \(U(n)\) be the list of property keys marked as unique for \(n\). The implementation also precomputes an edge anchor map from all edges $E_t$ and the list of edge labels marked as unique, \(U(\ell_{n,n'}(t))\) as
\[
A(n) =
\{(m,\ell_{n,m}(t)) : (n,m)\in N_t,\; l_{n,m}(t)\text{ is marked unique}\},
\]
where \(m\) is the neighboring anchor node and \(\ell_{n,m}(t)\) is the edge label. In our implementation, all unique edge labels are directionally applied to outgoing edges from the target node.

The construction is a candidate-generating cascade rather than an early-exit priority list. Node-level uniqueness, unique properties, recursive unique-edge descriptors, and point-marker fallbacks are all appended when available. Only
when no such descriptor can be built does our algorithm return a non-unique label. Recursive edge descriptors are bounded by a hop budget \(H\) (equivalent to the generated maximum reasoning depth) and a visited set to prevent cycles. 

\paragraph{Property rendering.}
Unique properties are treated differently depending on their key. If the key is
\texttt{description}, the property value is assumed to already be a compact
visual noun phrase, such as ``blue sedan'' or ``crossing to the right''. The
descriptor is therefore rendered directly as
\[
\text{``the } p_t(n)[\text{\texttt{description}}] \text{''}.
\]
For all other unique properties, the node type remains explicit and the
property is rendered as a constraint:
\[
\text{``the } \tau(n) \text{ whose } k \text{ is } p_t(n)[k]\text{''}.
\]
This distinction avoids awkward phrases such as ``the vehicle whose description
is blue sedan'' while preserving explicit property grounding for keys like
lane direction, lane type, line color, or traffic-light status.

\begin{algorithm}[t]
\LinesNotNumbered
\caption{Recursive construction of node descriptors.}
\label{alg:recursive-descriptor-construction}
\DontPrintSemicolon

\SetKwFunction{BuildDescriptor}{BuildDescriptor}
\SetKwProg{Fn}{Function}{:}{}

\KwIn{frame graph \(G_t\), target node \(n\), unique-edge anchor map \(A\), hop budget \(H\), visited set \(V\)}
\KwOut{candidate descriptor set \(D(n)\)}

\Fn{\BuildDescriptor{\(G_t, n, A, H, V\)}}{

\If{\(n\in V\)}{
    \Return{\(\emptyset\)}
}
\(V \leftarrow V\cup\{n\}\)\;
\(D \leftarrow \emptyset\)\;
\(P \leftarrow p_t(n)\), \(\nu \leftarrow \tau(n)\)\;

\If{\(P[\text{\texttt{is\_unique}}]\)}{
    add \(\langle\text{``the }\nu\text{''}, n, \mathrm{unique}, 0, \emptyset\rangle\) to \(D\)\;
}

\ForEach{\(k\in P[\text{\texttt{unique\_properties}}]\)}{
    \If{\(k=\text{\texttt{description}}\)}{
        \(s \leftarrow \text{``the }P[k]\text{''}\)\;
    }\Else{
        \(s \leftarrow \text{``the }\nu\text{ whose }k\text{ is }P[k]\text{''}\)\;
    }
    add \(\langle s,n,\mathrm{unique},0,\emptyset,\text{property}=k\rangle\) to \(D\)\;
}

\If{\(H>0\)}{
    \ForEach{\((m,\ell_{n,m}(t))\in A(n)\)}{
        \(C \leftarrow\) \BuildDescriptor{\(G_t,m,A,H-1,V\)}\;
        \ForEach{\(c\in C\)}{
            \(s \leftarrow \text{``the }\nu\text{ that }\rho\text{ }c.\mathrm{text}\text{''}\)\;
            \(\Delta \leftarrow c.\mathrm{deps}\cup\{(m,\ell_{n,m}(t),n,c.\mathrm{hops})\}\)\;
            add \(\langle s,n,c.\mathrm{unique},c.\mathrm{hops}+1,\Delta,\text{relation}=(\ell_{n,m}(t),c.n)\rangle\) to \(D\)\;
        }
    }
}

\If{\(P[\text{\texttt{position}}]\) exists}{
    add \(\langle\text{``the }\nu\text{ at }\langle P[\text{\texttt{position}}]\rangle\text{''}, n, \mathrm{unique},0,\emptyset\rangle\) to \(D\)\;
}

\If{\(D=\emptyset\)}{
    \Return{\(\{\langle n,n,\mathrm{nonunique},0,\emptyset\rangle\}\)}
}

\Return{\(D\)}
}
\end{algorithm}

\begin{figure}[t]
\centering
\usetikzlibrary{arrows.meta}
\begin{tikzpicture}[
    entity/.style={
        rounded corners=6pt,
        text=white,
        font=\bfseries\footnotesize,
        inner xsep=12pt,
        inner ysep=5pt
    },
    spine/.style={
        _gray,
        line width=0.8pt
    },
    arrow/.style={
        -{Latex[length=2mm]},
        _gray,
        line width=0.6pt
    },
    key/.style={
        text=gray!60,
        font=\bfseries\scriptsize
    },
    val/.style={
        text=black,
        font=\scriptsize,
        align=left
    }
]

\node[anchor=west, font=\bfseries\small] at (0,0) {Q: Is};

\node[entity, fill=_teal] (v1) at (2,0) {Vehicle-1};
\node[font=\bfseries\small] at (4.5,0) {in the same lane as};
\node[entity, fill=_green] (v3) at (6.7,0) {Bus-1};
\node[font=\bfseries\small] at (7.7,0) {\small?};

\def\xA{1.4}
\def\xB{6.4}
\draw[spine] (\xA,-0.3) -- (\xA,-3.1);
\draw[spine] (\xB,-0.3) -- (\xB,-3.1);

\def\yOne{-0.7}
\def\yTwo{-1.3}
\def\yThree{-1.9}
\def\yFour{-2.5}
\def\yFive{-3.1}

\foreach \y in {\yOne,\yTwo,\yFour,\yFive} {
    \draw[arrow] (\xA,\y) -- ++(0.38,0);
}
\node[key, anchor=west] at (1.8,\yOne) {(is UNIQUE NODE)};
\node[val, anchor=west] at (1.8,\yOne-0.25) {the vehicle};

\node[key, anchor=west] at (1.8,\yTwo) {(has UNIQUE PROP `description')};
\node[val, anchor=west] at (1.8,\yTwo-0.25) {the black jeep};

\node[key, anchor=west] at (1.8,\yThree) {(has UNIQUE PROP)};
\node[val, anchor=west] at (1.8,\yThree-0.25) {the vehicle with status 'parked'};

\node[key, anchor=west] at (1.8,\yFour) {(has UNIQUE EDGE)};
\node[val, anchor=west] at (1.8,\yFour-0.25) {the vehicle that cuts into the ego lane};

\node[key, anchor=west] at (1.8,\yFive) {(has BOUNDING BOX)};
\node[val, anchor=west] at (1.8,\yFive-0.25) {the vehicle at  \texttt{<box>...</box>}};

\foreach \y in {\yOne,\yTwo,\yFour,\yFive} {
    \draw[arrow] (\xB,\y) -- ++(0.28,0);
}
\node[key, anchor=west] at (6.8,\yOne) {(is UNIQUE NODE)};
\node[val, anchor=west] at (6.8,\yOne-0.25) {the bus};

\node[key, anchor=west] at (6.8,\yTwo) {(has UNIQUE PROP `description')};
\node[val, anchor=west] at (6.8,\yTwo-0.25) {the city-bus};

\node[key, anchor=west] at (6.8,\yThree) {(has UNIQUE PROP)};
\node[val, anchor=west] at (6.8,\yThree-0.25) {the bus with destination 'Downtown'};

\node[key, anchor=west] at (6.8,\yFour) {(has UNIQUE EDGE)};
\node[val, anchor=west] at (6.8,\yFour-0.25) {the bus that approaches the bus stop};

\node[key, anchor=west] at (6.8,\yFive) {(has BOUNDING BOX)};
\node[val, anchor=west] at (6.8,\yFive-0.25) {the bus at \texttt{<box>...</box>}};

\node[anchor=west, font=\bfseries\small] at (0,-4) {A: No, the first is in};

\node[entity, fill=_red] (a1) at (3.8,-4) {Lane-1};
\node[font=\bfseries\small] at (6.2,-4) {and the second is in};
\node[entity, fill=_red] (a3) at (8.5,-4) {Lane-2};
\node[font=\bfseries\small] at (9.5,-4) {\small.};

\def\xC{3.3}
\def\xD{8.0}
\draw[spine] (\xC,-4.3) -- (\xC,-7.1);
\draw[spine] (\xD,-4.3) -- (\xD,-7.1);

\def\yOnee{-4.7}
\def\yTwoe{-5.3}
\def\yThreee{-5.9}
\def\yFoure{-6.5}
\def\yFivee{-7.1}
\foreach \y in {\yOnee,\yTwoe,\yFoure,\yFivee} {
    \draw[arrow] (\xC,\y) -- ++(0.38,0);
}
\node[key, anchor=west] at (3.7,\yOnee) {(is UNIQUE NODE)};
\node[val, anchor=west] at (3.7,\yOnee-0.25) {the lane};

\node[key, anchor=west] at (3.7,\yTwoe) {(has UNIQUE PROP `description')};
\node[val, anchor=west] at (3.7,\yTwoe-0.25) {the left curb lane};

\node[key, anchor=west] at (3.7,\yThreee) {(has UNIQUE PROP)};
\node[val, anchor=west] at (3.7,\yThreee-0.25) {the lane with type 'express lane'};

\node[key, anchor=west] at (3.7,\yFoure) {(has UNIQUE EDGE)};
\node[val, anchor=west] at (3.7,\yFoure-0.25) {the lane that contains the ego};

\node[key, anchor=west] at (3.7,\yFivee) {(has BOUNDING BOX)};
\node[val, anchor=west] at (3.7,\yFivee-0.25) {the lane at  \texttt{<point>...</point>}};

\foreach \y in {\yOnee,\yTwoe,\yFoure,\yFivee} {
    \draw[arrow] (\xD,\y) -- ++(0.28,0);
}
\node[key, anchor=west] at (8.4,\yOnee) {(is UNIQUE NODE)};
\node[val, anchor=west] at (8.4,\yOnee-0.25) {the lane};

\node[key, anchor=west] at (8.4,\yTwoe) {(has UNIQUE PROP `description')};
\node[val, anchor=west] at (8.4,\yTwoe-0.25) {the left-turn lane};

\node[key, anchor=west] at (8.4,\yThreee) {(has UNIQUE PROP)};
\node[val, anchor=west] at (8.4,\yThreee-0.25) {the lane with direction 'opposite to ego'};

\node[key, anchor=west] at (8.4,\yFoure) {(has UNIQUE EDGE)};
\node[val, anchor=west] at (8.4,\yFoure-0.25) {the lane that leads up to the intersection};

\node[key, anchor=west] at (8.4,\yFivee) {(has BOUNDING BOX)};
\node[val, anchor=west] at (8.4,\yFivee-0.25) {the lane at \texttt{<point>...</point>}};

\end{tikzpicture}
\caption{Substituting all potential unique descriptors in relatively simple question templates leads to complicated multi-hop reasoning tasks and broad linguistic variety.}\label{fig:recursive_uniqueness}
\end{figure}
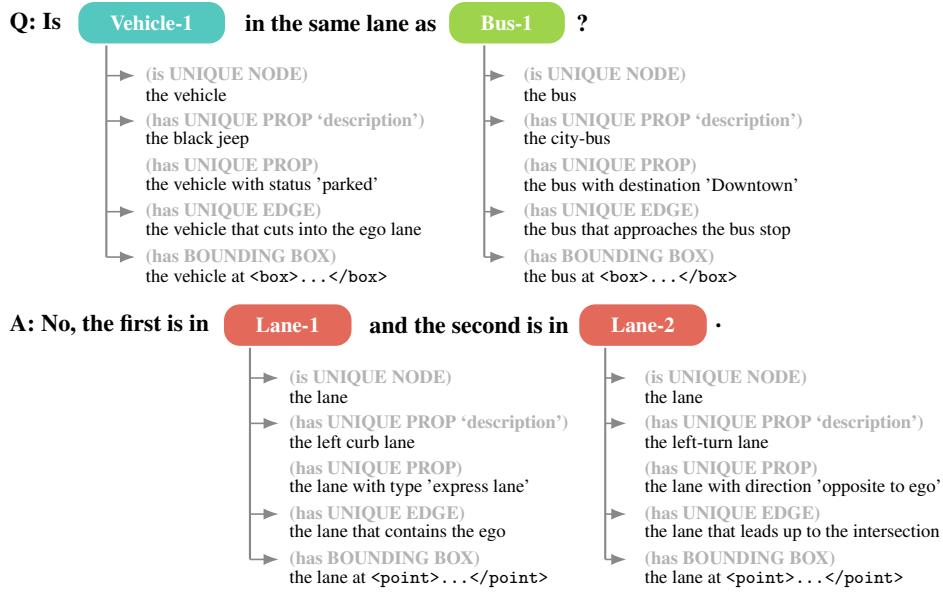

The recursive case transfers uniqueness from an already unique neighboring
descriptor to the target node through a unique relation. For example, if a lane
can be uniquely described as ``the ego lane'', then a neighboring lane connected
by a unique \texttt{left of} relation can be described as ``the lane that is
left of the ego lane''. The dependency list records the intermediate node and
relation so that the later CoT construction can first ground the anchor and
then walk the relation chain back to the queried target. The algorithm is detailed in \Cref{alg:recursive-descriptor-construction}, while an example of substituting unique descriptors into a query is provided in \Cref{fig:recursive_uniqueness}.

\subsection{Semi-automated graph construction with human enrichment}
\label{sec:graph_construction_supplementary}

This appendix describes the data-ingestion pipeline used to initialize the
graph interface.  The goal of this preprocessing is not to produce a final scene graph automatically.  Instead, it converts several heterogeneous autonomous-driving annotations into a common, frame-indexed representation that can be displayed in the annotation tool and then corrected, selected, and enriched by human annotators.

\paragraph{Input sources.}
For each Argoverse2 Sensor \cite{Argoverse2} log, we use the front-left, front-center, and front-right camera streams, ego poses, camera calibration, tracked 3D object annotations, and the Argoverse HD map.  We additionally use OpenLane-V2 \cite{wang2023openlane} annotations for traffic lights, road signs, their image-space
boxes, and their topological association to lane segments.  The derived annotations contain scene metadata such as 
image identifiers, per-camera calibration, and ego poses, as well as a broad range of scene elements (\texttt{lane\_segments}, \texttt{lane\_lines}, \texttt{intersections},
\texttt{splits}, \texttt{merges}, \texttt{pedestrian\_crossings},
\texttt{traffic\_elements}, and \texttt{objects}).  Dynamic quantities are keyed by an integer frame index.

\paragraph{Map and traffic-control ingestion.}
We first aggregate OpenLane-V2 frames that refer to the same Argoverse source log.  OpenLane-V2 traffic elements are merged over time by their element ID: traffic lights keep a per-frame state in \texttt{status}, while road signs
keep a symbolic \texttt{meaning}.  

We then load the corresponding Argoverse HD
map.  Lane topology is made bidirectionally consistent, geometrically contiguous lane segments are merged when their markings and topology permit it, and lane-marking boundaries are also merged into explicit lane-line objects.
For every lane, we store its \(\type{lane type} \in [\type{vehicle}, \type{bus}, \type{bike}]\). For every lane line we store its 3D polyline, \type{marking type}, as well as geometric connectivity to neighboring line segments and the lanes for which it is the left or right boundary, for semi-automatic edge instantiation.

Intersection regions are formed by clustering HD-map lane polygons marked as intersection segments.  Each cluster stores a 3D boundary, incoming and outgoing lanes, overlapping pedestrian crossings, and turn relations between incoming and outgoing lanes (for semi-automatic edge creation).  Lane splits and lane merges outside of intersections are detected from successor/predecessor branching in the lane topology.  Finally, map elements are filtered by visibility: at each ego pose we construct a forward-facing local window and keep only lanes, lane lines, crossings, intersections, splits, and merges that intersect that window.

\paragraph{Object ingestion.}
Tracked Argoverse objects are synchronized to each camera frame by selecting the closest 3D detection timestamp.  Because camera frames and lidar/object timestamps are not identical, lidar points and object poses are warped through
the ego-pose trajectory into the camera-frame timestamp.  We project each 3D box into the front cameras using the camera calibration, and keep boxes within the image and within the configured distance range of 50m.  The output object entry stores the object
type, as well as per-camera and per-frame 2D boxes. Additionally, we algorithmically determine the position of objects in the scene by testing if they are contained in one of the derived lane or intersection polygons.

\begin{figure}[h!]
    \centering
    \includegraphics[width=1\linewidth]{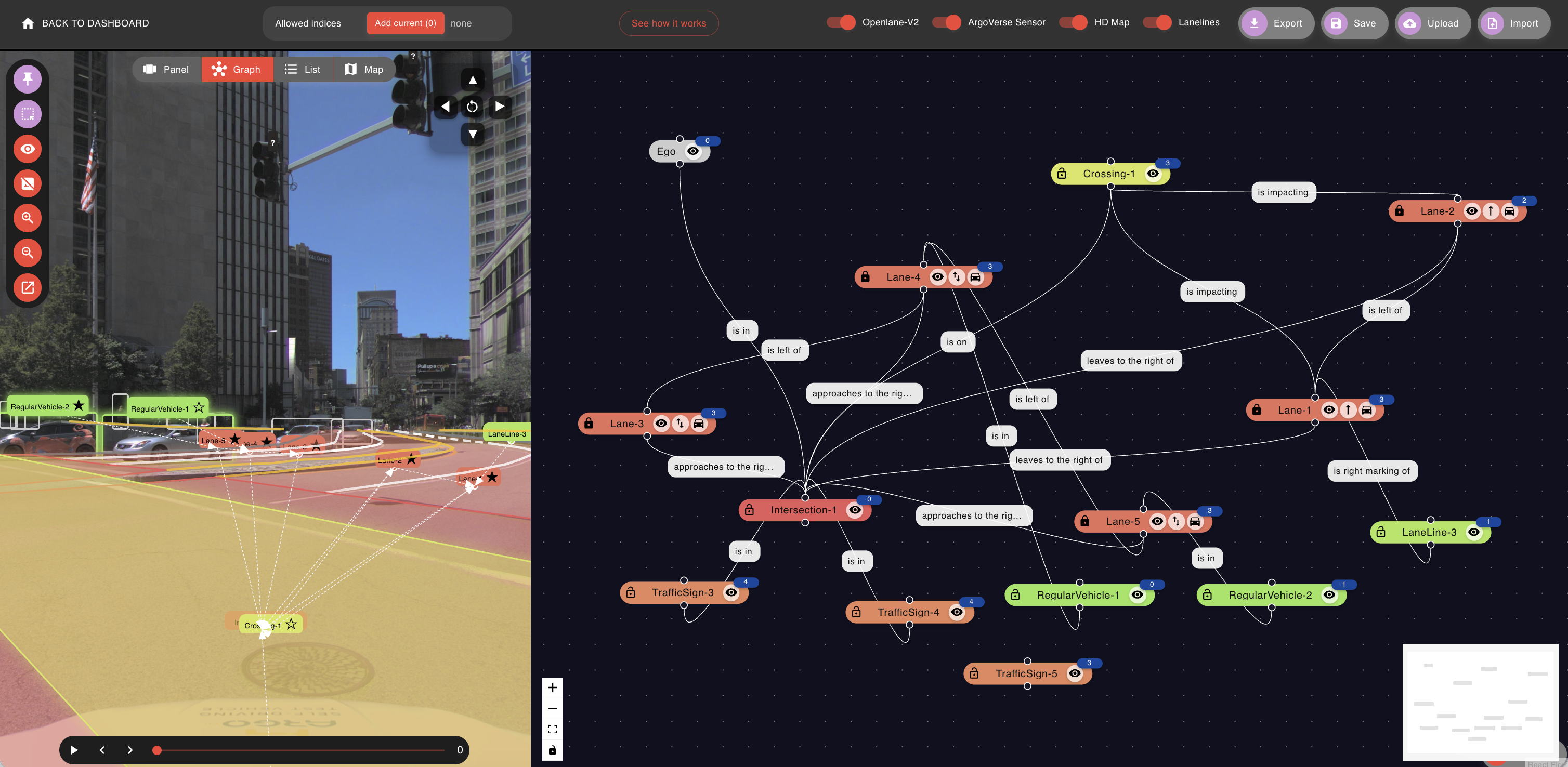}
    \caption{The graph creation tool semi-automatically creates a spatio-temporal scene graph, while human annotators filter and enrich the graph.}
    \label{fig:tool}
\end{figure}

\paragraph{Human enrichment in the graph tool.}
The graph creation tool in \Cref{fig:tool} displays the preprocessed scaffold as overlays on the three front camera images and as a top-down map view.  Human annotators then convert the scaffold into the final graph by selectively transferring relevant objects.  Clicking a projected source element creates a
typed graph node  with a unique node ID by transferring the pre-processed properties.  Annotators can also add manual point or bounding-box nodes when the source data misses an element, and they can attach or delete camera-specific position markers for a node in a specific frame.  Each node has a type, properties, helper
metadata such as position in the image space, and optional uniqueness anchors.  Properties are
template-driven but editable; 'locked' properties are treated as static, while 'unlocked' properties are stored per frame and can be propagated forward or backward in time.

Edges are created both automatically and manually.  Automatic edges use the source-derived helper fields: objects are dynamically placed in lanes or intersections (``is in''), lanes connected to traffic lights and those signs that are part of OpenLane-V2 (``controls''), lane neighbors are connected through pre-processed HD map data (``is left / right of''), as well as lane lines and their respective lanes (``is left / right marking of''), crossings to their intersection (``is on'').  In the graph editor, annotators can then add arbitrary edges, choose whether the predicate is static or frame-varying, and edit the label in natural language.  They can also remove edges for the current frame or globally, and propagate temporal edge labels across neighboring frames.  The exported
annotation therefore contains the refined source-derived geometric scaffold plus the human-curated semantic graph: selected nodes, corrected temporal attributes, natural-language predicates, and the uniqueness anchors used to identify objects in downstream question generation. An overview of the components of this process is provided in \Cref{tab:source-ingestion-human-enrichment}.

\

\scriptsize
\captionsetup{font=normalsize}
\setlength{\tabcolsep}{3pt}
\renewcommand{\arraystretch}{1.18}

\begin{longtable}{p{0.15\linewidth}p{0.19\linewidth}p{0.30\linewidth}p{0.28\linewidth}}
\caption{Source-derived fields used to initialize the graph annotation tool and the corresponding human enrichment operations. The automatic fields provide geometry, topology, visibility, and coarse semantics; human annotation converts this scaffold into the final graph by validating elements, correcting temporal properties, and adding natural-language node and edge semantics.}
\label{tab:source-ingestion-human-enrichment}\\

\toprule
\textbf{Element} &
\textbf{Primary source} &
\textbf{Automatic representation} &
\textbf{Human correction and enrichment} \\
\midrule
\endfirsthead

\toprule
\textbf{Element} &
\textbf{Primary source} &
\textbf{Automatic representation} &
\textbf{Human correction and enrichment} \\
\midrule
\endhead

\bottomrule
\endfoot

Image sequence and ego state &
Argoverse camera streams, ego poses, and calibration &
- &
Used as visual context for all annotation decisions. \\
\midrule

Lanes &
Argoverse HD map lane polygons and topology, with OpenLane-V2 traffic-control topology &
\texttt{lane} nodes with left/right boundary geometry, coarse \prop{lane type} and per-frame visibility. Semi-automated edge annotations are stored from left/right neighbor IDs. &
Annotators instantiate only relevant lanes, add properties such as \prop{direction} or \(\prop{variant}=\texttt{one\_way}\), remove false or irrelevant lanes, add or correct lane adjacency relations, and connect lanes leaving or approaching intersections. \\
\midrule

Lane lines &
Argoverse HD map lane boundaries &
\texttt{lane\_lines} with 3D geometry, \texttt{mark\_type} and per-frame visibility. Semi-automated edges are stored from \texttt{left\_of}/\texttt{right\_of} lane references. &
Annotators confirm which projected markings should enter the graph, correct marking properties, and express or correct marking relations such as ``is left marking of'' or ``is right marking of''. \\
\midrule

Traffic lights, road markings and signs &
OpenLane-V2 traffic elements and topology matrix &
\texttt{traffic\_lights} and \texttt{signs} with category, per-frame image bounding boxes, traffic-light \texttt{status}, or sign \texttt{meaning}; semi-automated lane associations are stored for edge creation. &
Annotators decide whether each element should become a graph node, correct traffic-light \prop{state} and \prop{type} (traffic lights for main road vs pedestrian or bike signals), refine sign meaning, and edit or delete regulatory edges. Additionally, human oversight is needed for differentiating between road markings and signs, which are not treated as separate categories in OpenLane-V2. \\
\midrule

Pedestrian crossings &
Argoverse HD map pedestrian crossing polygons &
\texttt{pedestrian\_crossings} with geometry and visibility. Semi-automatic edges are created between crossings and intersections (``is in'').  &
Annotators confirm crossings that are semantically relevant and connect them to intersections with natural-language predicates such as ``is on'', define additional properties such as \prop{style} or \prop{marking\_quality}, among others. \\
\midrule

Intersections and turn structure &
Argoverse HD map lane polygons and predecessor/successor topology &
\texttt{intersections} with geometry, incoming and outgoing lanes, linked crossings. &
Annotators select visible intersections, validate incoming/outgoing structure, and edit relations that describe approach, departure, and allowed maneuvers, as these can not reliably be extracted programmatically in natural language for complex intersections. \\
\midrule

Lane splits and merges &
Derived from lane predecessor/successor branching &
\texttt{splits} and \texttt{merges} store a 3D event point, incoming/outgoing lanes, and per-frame visibility. &
Annotators instantiate or discard these event nodes and edit the topology predicates used in the final graph. \\
\midrule

Tracked objects &
Argoverse 3D tracked object annotations &
\texttt{objects} with object type, per-camera per-frame 2D boxes, and per-frame lane/intersection relationship (``is in''). &
Annotators choose which objects matter, correct object class or visibility, add missing image positions with point/box clicks, and refine object descriptions, properties, etc. \\
\midrule

All remaining nodes &
human annotation &
-  &
Annotators add, delete, or rewrite edge labels and node properties in natural language for interesting, missing or relevant objects in the scene. This can be everything within and beyond the ingested annotations. \\

\end{longtable}

\normalsize
\paragraph{Annotation effort and quality control.}
Average annotation time is approximately 5 minutes per CRS scene graph, though it varies considerably across scenes: annotation is a selective curation process, not exhaustive labeling: annotators include only the entities, attributes, and relations they judge salient for a given scene, so time scales with what an annotator chooses to curate, not with a fixed per-scene inventory. On quality control, we deliberately do not report a classical inter-annotator agreement statistic (e.g., label overlap or Cohen's kappa), because such metrics assume a single ground-truth annotation two annotators should converge on, which does not hold here by design: CRS operates under an explicit open-world, no-completeness assumption (\Cref{sec:welldefinedness}) — a graph need not include every instance of an entity type unless completeness is required for a specific query — so two annotators can validly produce graphs with different emphasis (different nodes included, different attributes added, different uniqueness anchors chosen) without either being incorrect. 

What quality control does target, and what actually matters for validity, is not agreement on what was selected but correctness of what was included: is a given property value accurate, is a relation correctly labeled, is a chosen descriptor actually unique. We check this via (i) a second-pass review of every annotated graph by a second person, verifying well-definedness and correctness rather than agreement with the first annotator's selections, and (ii) manual verification of the generated question-answer pairs specifically for the held-out validation set (our existing expert-verified set, see \Cref{sec:experiments}). 

\subsection{CRS-enriched scene statistics}
\label{sec:stats}

This appendix reports the graph statistics for the 80 CRS-enriched scenes used in this study. Following the notation in \Cref{sec:methodology}, each annotated scene is treated
as a temporal directed graph representation \(G=\{G_0,\ldots,G_T\}\), with per-frame graphs \(G_t=(N_t,E_t)\) with a temporal
window size of \(w=4\). For each training example, the first \(w-1\) frames provide
temporal context, and statistics are accumulated only for the queried current
frame. Thus, the 80 CRS-enriched scenes for training yield
\[
M =
\sum_{G}
G_t
=
\sum_{G}
\left|\{t : t \geq w-1\}\right|
= 2{,}320
\]
evaluated frame graphs, with an average of \(29.0\) queried frame graphs per CRS graph. 

The number of CRS graphs is small at the scene level but dense at the graph-observation
level. An average queried frame graph contains \(14.79\) nodes, \(38.46\)
directed edges, and \(87.12\) node-property entries. Equivalently, nodes have
an average incoming-or-outgoing edge incidence of \(2|E_t|/|N_t|=5.20\) and
carry \(5.89\) properties. The uniqueness signal used by recursive descriptors
is also present throughout the train graphs: an average frame graph contains
\(1.63\) uniquely identifiable node anchors, \(16.76\) unique edge anchors, and
\(3.75\) unique property anchors. Completeness is more frequent for lanes than
for crossings: \(1{,}904\) frame graphs are lane-complete (\(82.1\%\)), while
\(932\) are crossing-complete (\(40.2\%\)). \Cref{tab:train-crs-graph-stats} summarizes these findings.

\begin{table*}[h!]
\centering
\small
\caption{CRS graph statistics. Totals are accumulated over the
\(M=2{,}320\) evaluated frame graphs induced by the 80 CRS graphs used in training.}
\label{tab:train-crs-graph-stats}
\begin{tabular}{@{}p{0.48\linewidth}r r p{0.18\linewidth}@{}}
\midrule
\textbf{Statistic} & \textbf{Total} & \textbf{Mean} & \textbf{Unit} \\
\toprule
Evaluated frame graphs \(|\sum G(t)|\) & 2,320 & 29.00 & per CRS graph \\
Node observations \(\sum_t |N_t|\) & 34,316 & 14.79 & per \(G_t\) \\
Directed edge observations \(\sum_t |E_t|\) & 89,230 & 38.46 & per \(G_t\) \\
Node-property entries \(\sum_{t,n} |P_t(n)|\) & 202,117 & 87.12 & per \(G_t\) \\
Incoming/outgoing edge incidence \(2|E_t|/|N_t|\) & -- & 5.20 & per node \\
Node-property density & -- & 5.89 & per node \\
\midrule
Lane-complete frame graphs, \(c_t(\mathrm{lane})=1\) & 1,904 & 82.1 & percent of \(G_t\) \\
Crossing-complete frame graphs, \(c_t(\mathrm{crossing})=1\) & 932 & 40.2 & percent of \(G_t\) \\
\midrule
Unique node anchors & 3,785 & 1.63 & per \(G_t\) \\
Unique edge anchors & 38,891 & 16.76 & per \(G_t\) \\
Unique property anchors & 8,702 & 3.75 & per \(G_t\) \\
\bottomrule
\end{tabular}
\end{table*}
\normalsize
\subsection{Question type analysis}
\label{sec:question-type-analysis}
The training set contains 23,343 samples and the validation set contains 1,014
expert-verified samples, both spanning the same 19 question types. These types
cover direct perception of road attributes, counting over lanes and crossings,
pairwise comparisons, localization/existence queries, and relational control or
containment queries. We assign each type to the same reasoning split and
evaluation bucket used in the benchmark analysis: \textit{perception-like} questions
primarily ask for visible attributes which require lower degrees of reasoning, counts, temporal status changes, or simple
pairwise lane-direction comparisons, while \textit{reasoning-heavy} questions require
resolving a localized target or traversing a graph relation before selecting the
answer.

\begin{table*}[t]
\centering
\scriptsize
\setlength{\tabcolsep}{3pt}
\renewcommand{\arraystretch}{1.12}
\caption{Question type distribution in the training and validation sets.}
\label{tab:question-type-analysis}
\begin{tabular}{@{}p{0.25\linewidth}r r p{0.30\linewidth} p{0.10\linewidth} p{0.15\linewidth}@{}}
\toprule
\textbf{Question type} & \textbf{Train} & \textbf{Val.} & \textbf{Asks about} & \textbf{Reasoning} & \textbf{Bucket} \\
\midrule
\texttt{counting at intersection per direction} & 1,388 & 67 & Lane counts by approach/exit direction at an intersection & Low & Counting \\
\texttt{counting crossing} & 225 & 43 & Number and layout of pedestrian crossings & Low & Counting \\
\texttt{counting per direction} & 1,087 & 74 & Lane counts split by ego/opposite direction off intersection & Low & Counting \\
\texttt{counting generic} & 2,174 & 80 & Generic lane count for a requested direction or road extent & Low & Counting \\
\midrule
\texttt{lane direction} & 2,185 & 75 & Whether a lane follows or opposes the ego direction & Low & Properties \\
\texttt{lane type} & 1,811 & 70 & Semantic lane class such as vehicle, bike, bus lane & Low & Properties \\
\texttt{line color} & 1,150 & 63 & Color of a lane boundary marking & Low & Properties \\
\texttt{traffic light status} & 1,546 & 56 & Current traffic-light or pedestrian-signal state & Low & Properties \\
\texttt{line marking} & 1,150 & 57 & Full lane-line marking style & Low & Properties \\
\texttt{line type} & 1,150 & 56 & Lane-line pattern or visibility type & Low & Properties \\
\texttt{crossing type} & 358 & 33 & Pedestrian-crossing marking style & Low & Properties \\
\texttt{traffic light change} & 1,673 & 60 & Whether a traffic light changed over the temporal window & Low & Properties \\
\midrule
\texttt{pairwise lane comparison by direction} & 1,339 & 80 & Whether two lanes share the same travel direction & Low & Comparison \\
\texttt{pairwise vehicle location} & 1,071 & 34 & Whether two actors share a lane or their relative lane position & High & Comparison \\
\midrule
\texttt{pointing} & 1,656 & 38 & Image location of a referenced scene element & High & Existence \\
\texttt{existence of crossings} & 436 & 43 & Presence, location, or style of an intersection crossing & High & Existence \\
\midrule
\texttt{sign controls lane} & 310 & 14 & Which lane set a road sign controls & High & Relational \\
\texttt{traffic light controls lane} & 1,345 & 28 & Which or how many lanes a traffic light controls & High & Relational \\
\texttt{vehicle position} & 1,289 & 43 & Lane containing the ego vehicle or another actor & High & Relational \\
\bottomrule
\end{tabular}
\end{table*}
\subsection{Evaluation protocol} \label{sec:evaluation_protocol}
All open- and closed-source baselines, as well as our finetuned models, were prompted with the identical, single-turn, zero-shot prompt template with explicit CoT instructions shown in full below, with no per-model prompt tuning. References to objects ("the vehicle marked with box 1") were drawn onto the respective images. 
\begin{tcolorbox}[
    colback=gray!5,
    colframe=gray!50,
    title={VLM Prompt},
    fonttitle=\bfseries,
    boxrule=0.5pt,
    arc=2pt
]
\small{
You are given images captured from an ego vehicle.\\
Images are provided for the past 4 frames starting from the oldest to the most recent, with 3 images per frame for a total of 12 images.\\
The question always refers to the last frame.\\
Images per frame are provided in this exact order:\\
1. LEFT camera \\
2. CENTER camera \\
3. RIGHT camera \\
\\
Output format rules:\\
Before providing the final answer, perform a step-by-step visual analysis of the scene, referencing specific cameras or grounding boxes where relevant to justify your conclusion.\\
\\
Question:\\
<question text>\\
\\
Answers:\\
A. <option>\\
B. <option>\\
C. <option>\\
D. <option>}
\end{tcolorbox}
We confirmed the model reasoning by inspecting the saved responses, which all consistently contain a multi-step visual analysis before choosing and returning the final answer choice. To assure that models are not unfairly penalized for not returning the exact answer format, we allow flexible returns ("The  answer is X", "X", "X: There are two lanes", "There are two lanes", etc...) and randomly inspect answers labeled as wrong manually. We did not include few-shot instructions in this work, for a practical reason worth noting: each example in this benchmark is itself 12 images (4 temporal frames × 3 cameras), so even a 2- or 3-shot prompt multiplies the visual context by a factor 3 or 4, approaching the per-request image-count/context limits. 

\input{tables/benchmarks_additional}
\subsection{Additional quantitative results}\label{sec:additional_results}
\input{tables/n_hops}
\paragraph{Overall accuracy.} This subsection details the performance of additional open-source models evaluated on our validation split in addition to those in \Cref{tab:model_accuracy_detailed}. The results are reported in \Cref{tab:model_accuracy_detailed_suppl}.

\paragraph{Performance with reasoning depth.} This subsection details the performance decrease with reasoning depth (see \Cref{sec:experiments} and \Cref{fig:accuracy_by_reasoning_depth}) for all evaluated models. The results are reported in \Cref{tab:model_accuracy_nhops}. Our finetuned models display the smallest delta between performance on queries with lower reasoning depth and
queries with higher reasoning depth.

\paragraph{Impact of decoys on no-image baseline.} This subsection further analyzes the no-image baseline, corresponding to the result of supervised finetuning ($-$ images) in \Cref{tab:model_accuracy_detailed}. In order to isolate the effects of the linguistic reasoning component from the grounded visual reasoning, we baselined against a model that is fine-tuned with a single image taken randomly from the training dataset, applied across all samples. This renders the model ``blind'' to the true scene corresponding to the visual question, and is trained to answer based on text cues alone. We report results of this model under the multiple choice setting in \Cref{sec:experiments}. To de-correlate the visual reasoning from text cues, we train a pair of \textit{free-form} question generation models, using image inputs \textit{vs.} the placeholder fake image input, respectively. 

\begin{figure}[htbp]      
    \centering
    \input{figures/supplementary/free_form_img}   
    \vspace{-0.4cm}
    \caption{Per-type accuracy for free-form generation comparing models \texttt{with image} \textit{vs.} models with \texttt{no image}.}
    \label{fig:free-form-img-supplementary}
\end{figure}
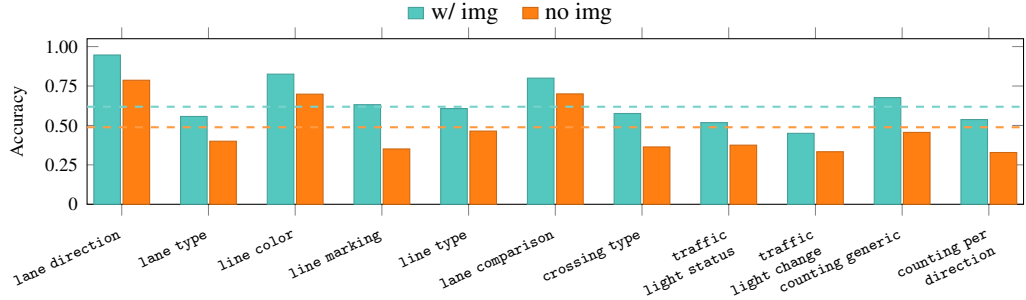

As shown in \Cref{fig:free-form-img-supplementary}, the performance difference between using visual context (\texttt{with image}) is more pronounced than without visual context (\texttt{no image}). In order to compute accuracy from free-form answer generation, we limit the question types to 11 that can be matched \textit{via} regex. 
We further analyze the distribution of the outputs with and without images under the counting task in \Cref{fig:counting-histograms-supplementary}, showing that models that are trained without using images tend to collapse to the most likely answer. This highlights the fact that models trained without visual information can retain road structure, but there is a need for visual grounding for accurate road understanding.  

\input{figures/supplementary/free_form_img_counting}
\subsection{Additional qualitative results}\label{sec:additional_results_qualitative}
We provide additional qualitative results of the output of the finetuned Qwen-4B model trained with CoT reasoning (last row in \Cref{tab:model_accuracy_detailed}) in \Cref{fig:qualitative_examples}. The reasoning trace is an excellent means of investigating model failures.
\newcommand{\AnswerPill}[3]{%
  \tcbox[
    answerpill,
    colback=#3!30,
    colframe=#3!60
  ]{%
    \parbox{2.7cm}{\tiny\textbf{#1}\quad #2} 
  }%
  \vspace{1mm}
}

\begin{figure}[t]

\centering


\begin{subfigure}[t]{\linewidth}
\tikzset{
  point/.style={
    circle,
    fill=#1,
    draw=white,
    line width=0.35pt,
    inner sep=1.4pt,
    font=\tiny\bfseries,
    text=white
  }
}

\begin{minipage}[t]{0.4\linewidth}
\vspace{0pt}
\centering

\begin{minipage}{0.29\linewidth}
\centering
\begin{tikzpicture}
\node[inner sep=0, anchor=south west] (img) at (0,0) {
  \includegraphics[width=\linewidth,height=1.5cm]{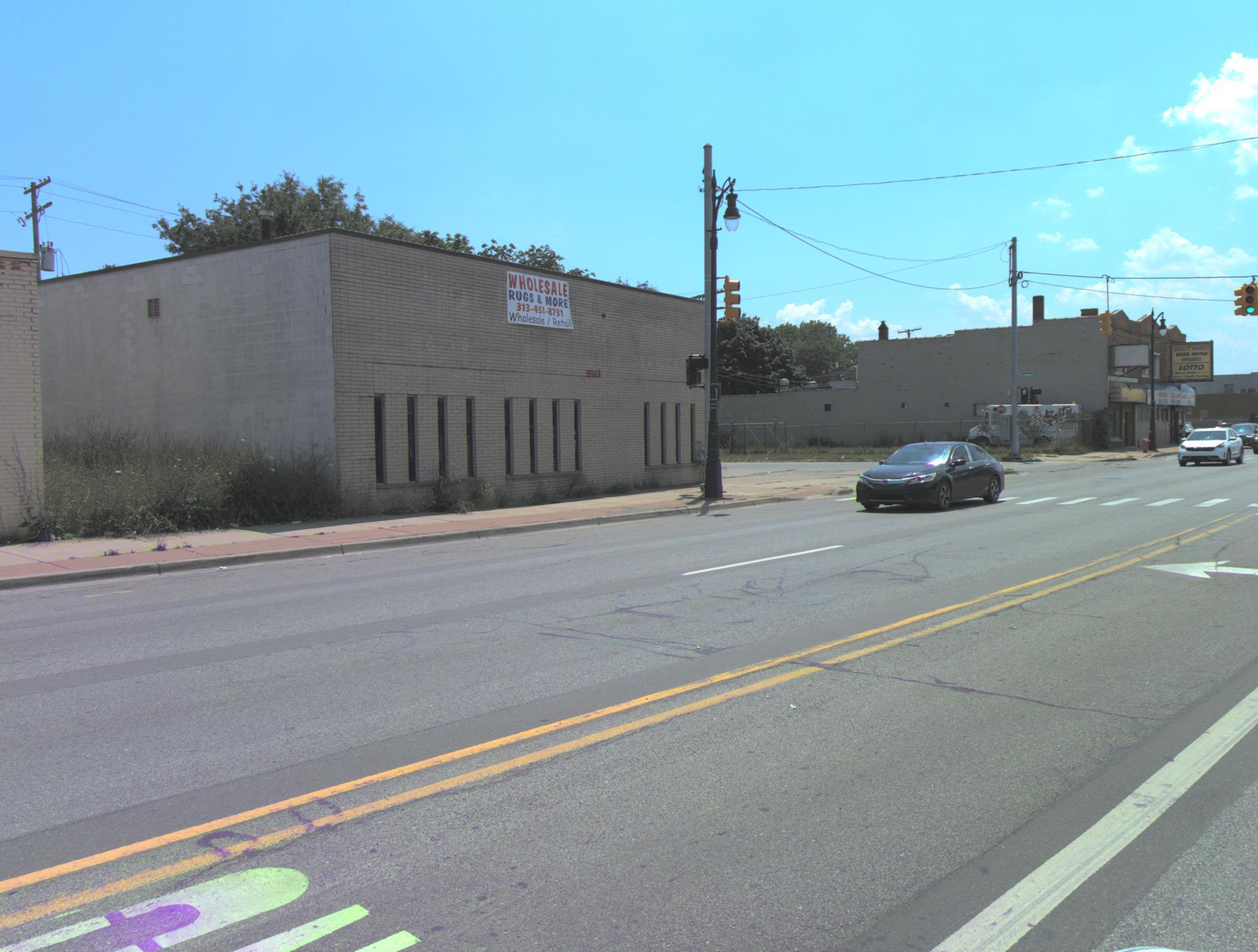}
};
\begin{scope}[x={(img.south east)}, y={(img.north west)}]
\end{scope}
\end{tikzpicture}
\end{minipage}%
\begin{minipage}{0.38\linewidth}
\centering
\begin{tikzpicture}
\node[inner sep=0, anchor=south west] (img) at (0,0) {
  \includegraphics[width=\linewidth,height=3.2cm]{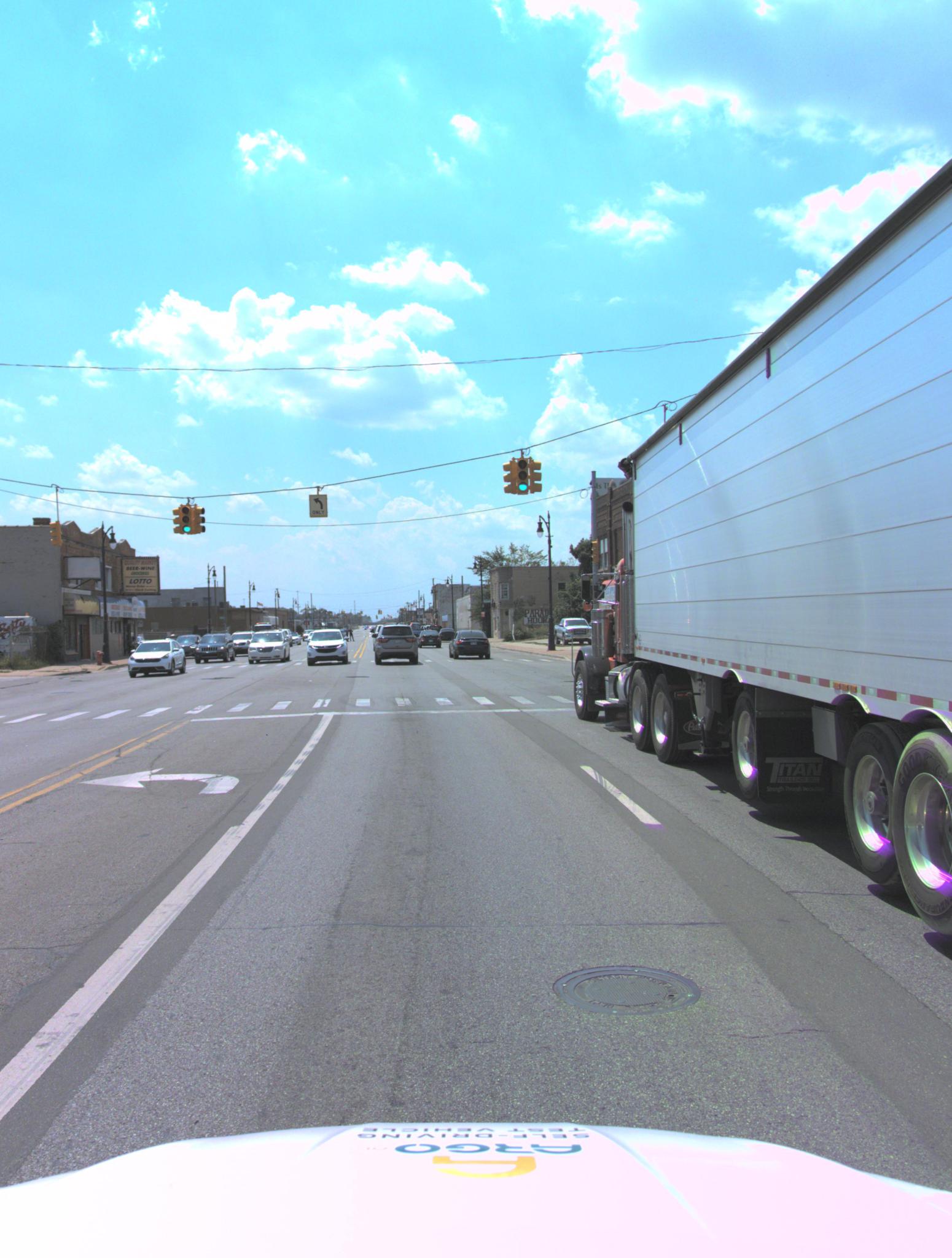}
};
\begin{scope}[x={(img.south east)}, y={(img.north west)}]
\ImgPin{0.505}{1-0.824}{P1}{_orange}
\ImgPin{0.809}{1-0.696}{P2}{_teal}
\ImgPin{0.770}{1-0.721}{P2}{_purple}
\ImgBoxPin
  {0.649}{0}   
  {0.999}{1-0.105}   
  {P3}           
  {_red}  
  {1pt}         
  {1pt}         
\end{scope}
\end{tikzpicture}
\end{minipage}%
\begin{minipage}{0.29\linewidth}
\centering
\begin{tikzpicture}
\node[inner sep=0, anchor=south west] (img) at (0,0) {
  \includegraphics[width=\linewidth,height=1.5cm]{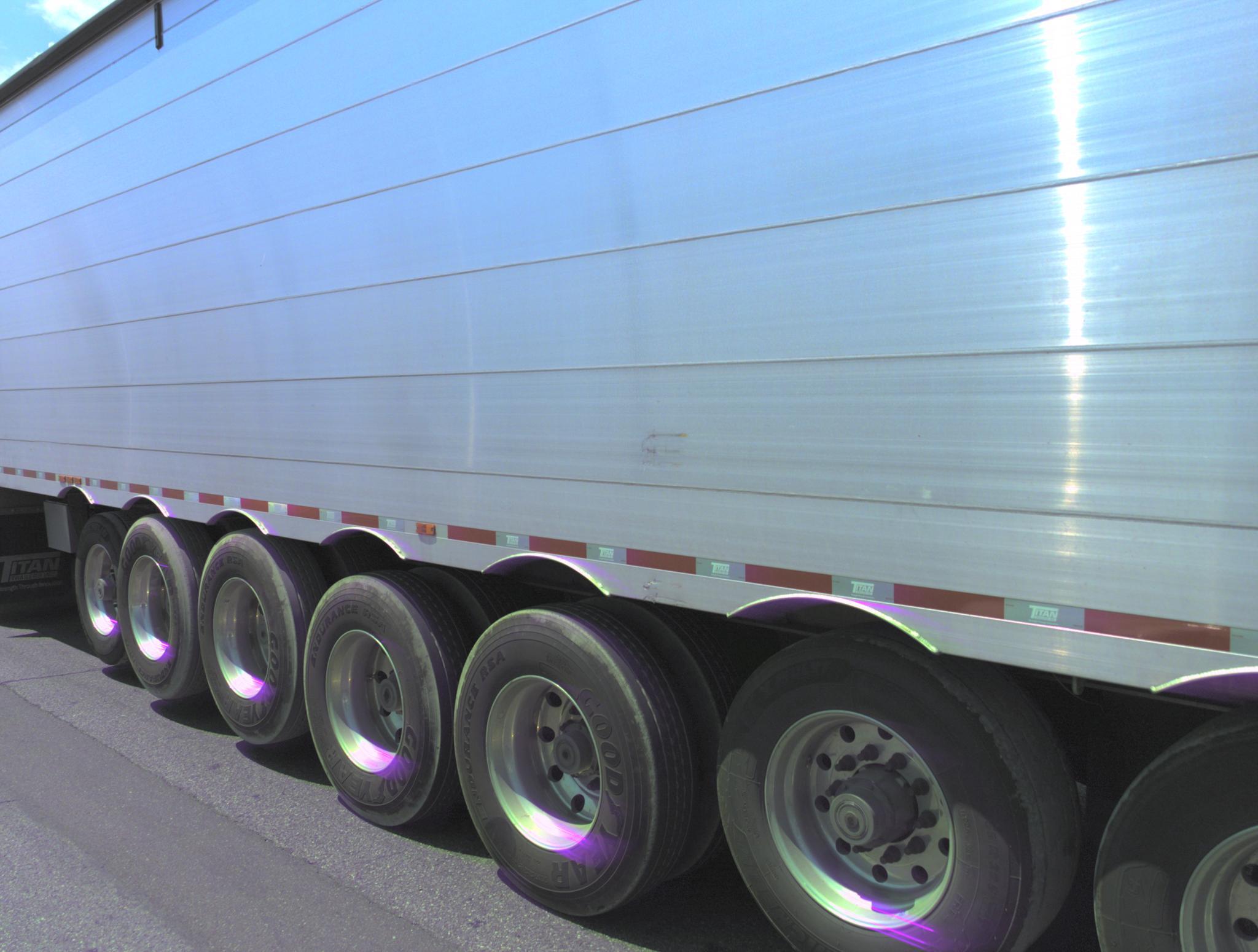}
};
\begin{scope}[x={(img.south east)}, y={(img.north west)}]
\ImgPin{0.447}{1-0.835}{P1}{_teal}
\ImgPin{0.11}{1-0.892}{P1}{_purple}

\ImgBoxPin
  {0.01}{0}   
  {0.98}{1-0.01}   
  {P3}           
  {_red}  
  {1pt}         
  {1pt} 
\end{scope}
\end{tikzpicture}
\end{minipage}

\end{minipage}%
\hspace{0.1cm}
\begin{minipage}[t]{0.18\linewidth}

{\tiny\bfseries QUESTION}\par\par

{\tiny
\parbox{\linewidth}{%
  \baselineskip=6pt
  What is the left lane line type of the lane that is right of the ego lane?
}}\par

\tcbset{
  answerpill/.style={
    on line,
    boxrule=0.25pt,
    arc=1.2mm,
    left=1mm,
    right=1mm,
    top=0.4mm,
    bottom=0.4mm,
    boxsep=0pt,
    width=3cm  
  }
}

\vspace{0.1cm}
{\tiny\bfseries ANSWERS}
\AnswerPill{A}{There is no lane line visible.}{_lightgray}
\AnswerPill{\color{green!50!black}\checkmark}{dashed.}{_green}\par
\AnswerPill{C}{double dashed.}{_lightgray}
\AnswerPill{D}{double solid.}{_lightgray}
\end{minipage}
\hfill
\begin{minipage}[t]{0.37\linewidth}
\tcbset{
  stepbox/.style={
    colback=gray!8,
    colframe=gray!20,
    boxrule=0.25pt,
    arc=1mm,
    left=0.6mm, right=0.6mm,
    top=0.4mm, bottom=0.4mm,
    boxsep=0pt
  }
}

\newcommand{\StepBox}[1]{%
  \begin{tcolorbox}[stepbox]
  {\tiny #1}
  \end{tcolorbox}
}
{\tiny

\textbf{Reasoning Output:}
\vspace{-0.1cm}
\StepBox{\textbf{Step 1:} Identify the ego lane, which is visible in the CENTER view at \TextPin{P1}{_orange}.}

\StepBox{\textbf{Step 2:} The lane in question is the lane that is right of that lane, and is visible in the CENTER and RIGHT view at \TextPin{P2}{_teal}.}

\StepBox{\textbf{Step 3:} The lane's description is rightmost lane. The lane contains a box\_truck located in CENTER and RIGHT view at \TextPin{P3}{_red}. }

\StepBox{\textbf{Step 4:} Finally, the lane marking in question is located in the CENTER and RIGHT view at \TextPin{P4}{_purple}. It is DASHED\_WHITE.}

\StepBox{\textbf{Conclusion:} the correct answer is `dashed'.}
}

\end{minipage}

    \caption{Example output of the model trained with CoT on a question inquiring about the marking type of a lane line.}
    \label{fig:qualitative_example1}
\end{subfigure}

\vspace{2mm}


\begin{subfigure}[t]{\linewidth}

\centering

\tikzset{
  point/.style={
    circle,
    fill=#1,
    draw=white,
    line width=0.35pt,
    inner sep=1.4pt,
    font=\tiny\bfseries,
    text=white
  }
}

\begin{minipage}[t]{0.4\linewidth}
\vspace{0pt}
\centering

\begin{minipage}{0.29\linewidth}
\centering
\begin{tikzpicture}
\node[inner sep=0, anchor=south west] (img) at (0,0) {
  \includegraphics[width=\linewidth,height=1.5cm]{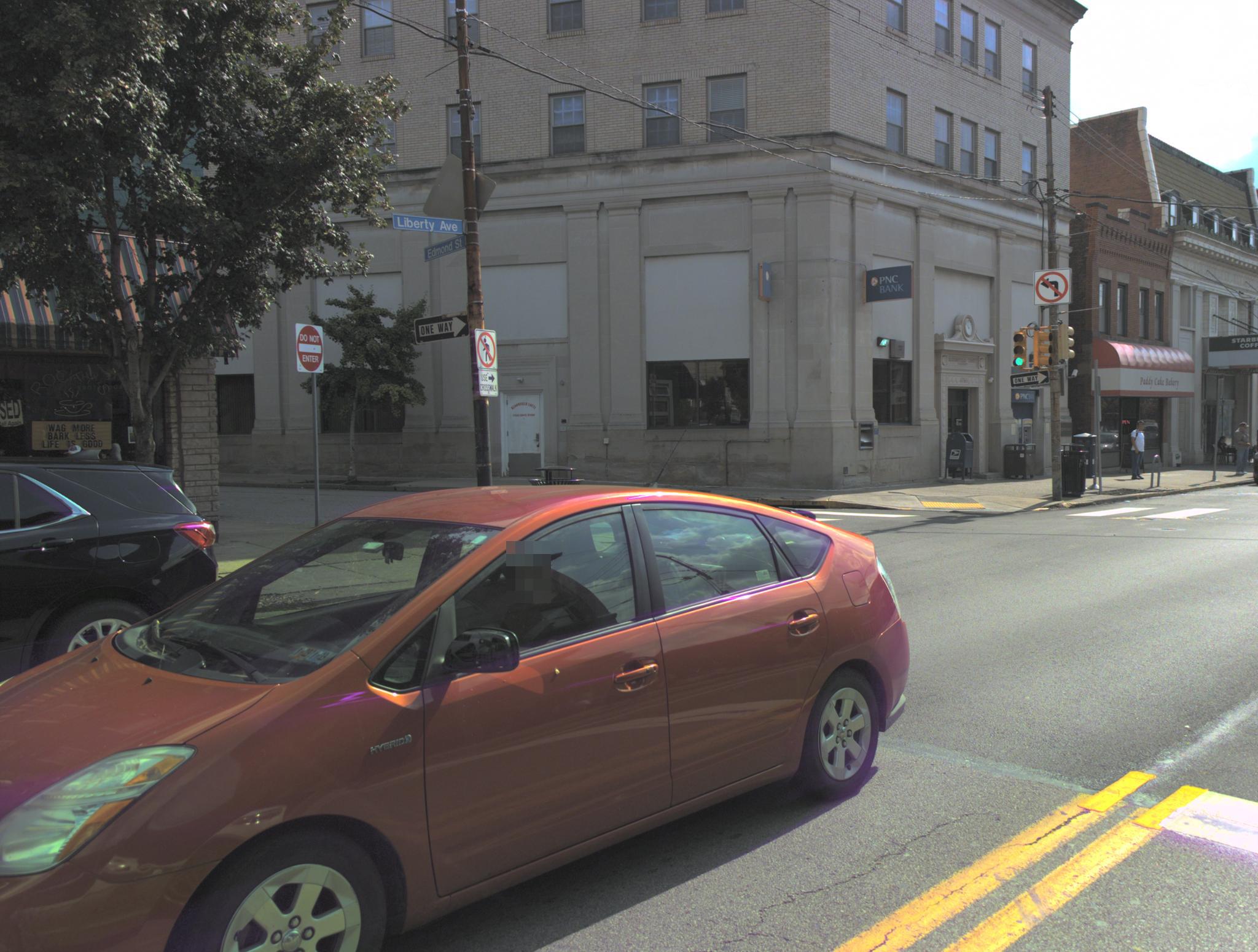}
};
\begin{scope}[x={(img.south east)}, y={(img.north west)}]
\end{scope}
\end{tikzpicture}
\end{minipage}%
\begin{minipage}{0.38\linewidth}
\centering
\begin{tikzpicture}
\node[inner sep=0, anchor=south west] (img) at (0,0) {
  \includegraphics[width=\linewidth,height=3.2cm]{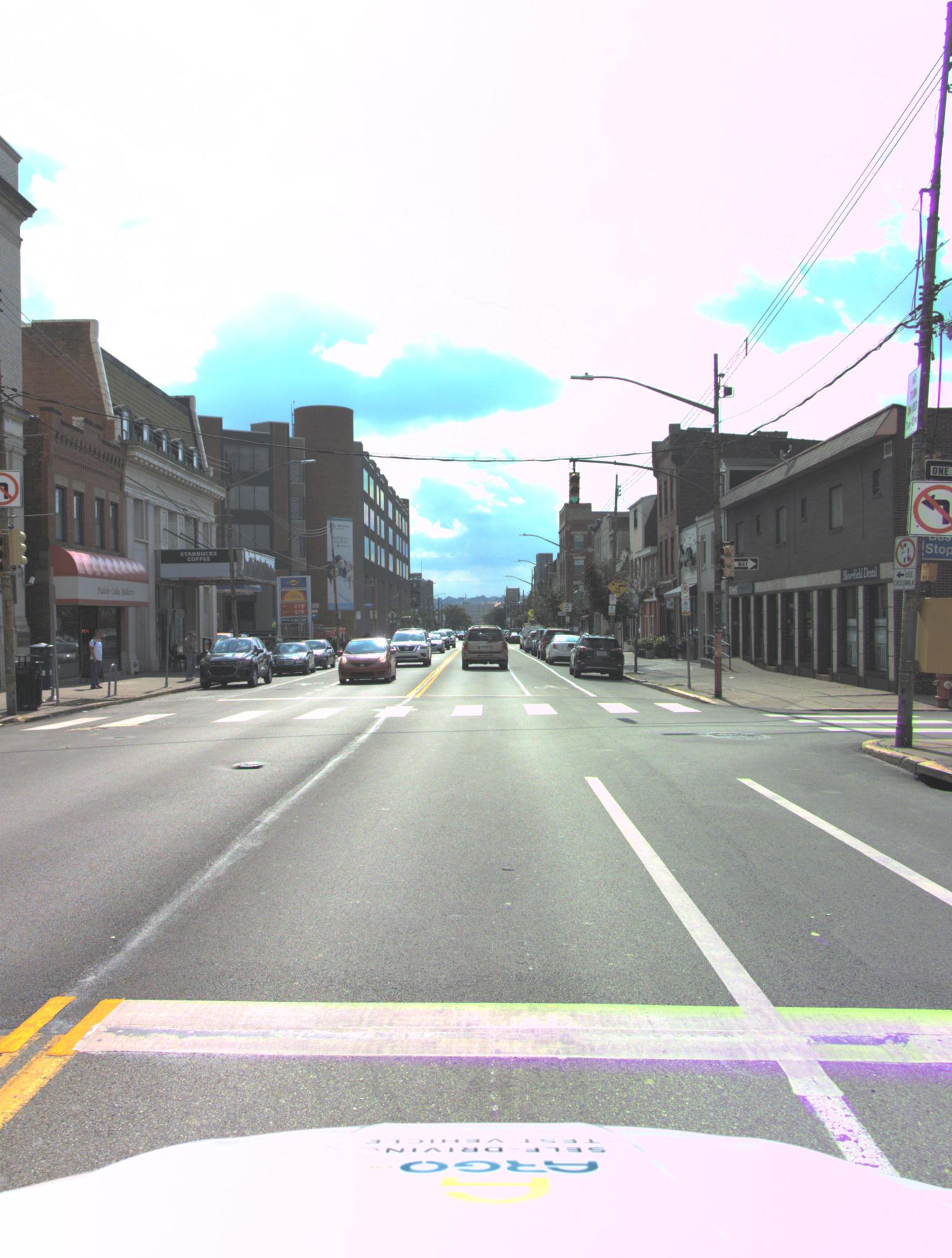}
};
\begin{scope}[x={(img.south east)}, y={(img.north west)}]
\ImgPin{0.864}{1-0.592}{P1}{_red}
\end{scope}
\end{tikzpicture}
\end{minipage}%
\begin{minipage}{0.29\linewidth}
\centering
\begin{tikzpicture}
\node[inner sep=0, anchor=south west] (img) at (0,0) {
  \includegraphics[width=\linewidth,height=1.5cm]{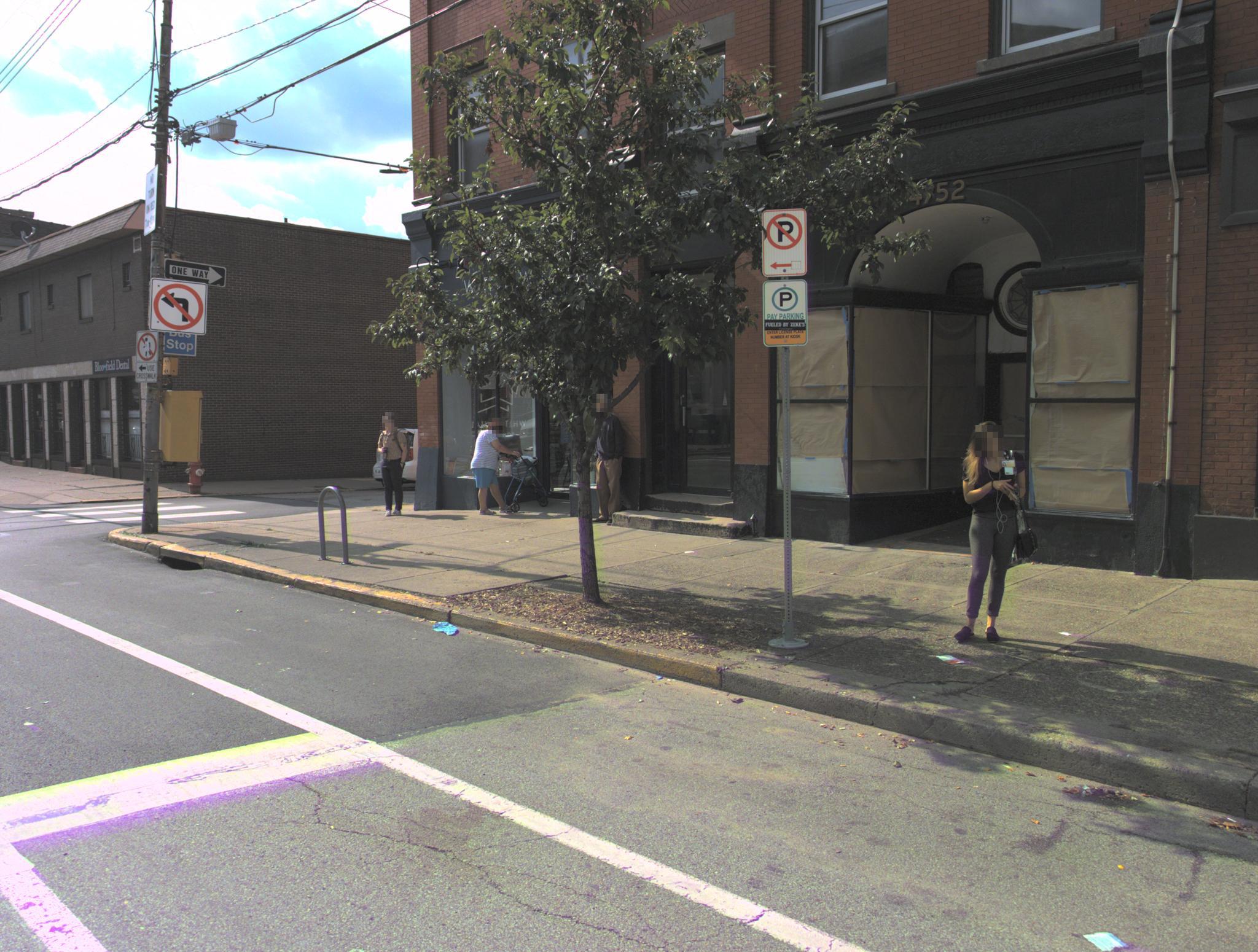}
};
\begin{scope}[x={(img.south east)}, y={(img.north west)}]
\ImgPin{0.111}{1-0.571}{P1}{_red}
\end{scope}
\end{tikzpicture}
\end{minipage}

\end{minipage}%
\hspace{0.1cm}
\begin{minipage}[t]{0.18\linewidth}

{\tiny\bfseries QUESTION}\par\par

{\tiny
\parbox{\linewidth}{%
  \baselineskip=6pt
  Is there a crossing to the right of the intersection?
}}\par

\tcbset{
  answerpill/.style={
    on line,
    boxrule=0.25pt,
    arc=1.2mm,
    left=1mm,
    right=1mm,
    top=0.4mm,
    bottom=0.4mm,
    boxsep=0pt,
    width=3cm  
  }
}

\vspace{0.1cm}
{\tiny\bfseries ANSWERS}\par
\AnswerPill{A}{Yes, but it is unmarked.}{_lightgray}\par
\AnswerPill{\color{green!50!black}\checkmark}{Yes, marked as a zebra crossing.}{_green}\par
\AnswerPill{C}{No, but there is one to the left.}{_lightgray}\par
\AnswerPill{\color{_red!50!black}x}{None of the above.}{_red}\par
\end{minipage}
\hfill
\begin{minipage}[t]{0.37\linewidth}
\tcbset{
  stepbox/.style={
    colback=gray!8,
    colframe=gray!20,
    boxrule=0.25pt,
    arc=1mm,
    left=0.6mm, right=0.6mm,
    top=0.4mm, bottom=0.4mm,
    boxsep=0pt
  }
}

\newcommand{\StepBox}[1]{%
  \begin{tcolorbox}[stepbox]
  {\tiny #1}
  \end{tcolorbox}
}
{\tiny

\textbf{Reasoning Output:}
\vspace{-0.1cm}
\StepBox{\textbf{Step 1:} I am tasked to identify if there is a crossing to the right of the intersection, and determine its potential marking style. }

\StepBox{\textbf{Step 2:} Analyzing the sequence of frames and their camera views. I detect a crossing to the right of the intersection, visible in CENTER and RIGHT view at \TextPin{P1}{_red}. }

\StepBox{\textbf{Step 3:} Determining the marking style. This crosswalk is a white outlined rectangle crossing. 
}

\StepBox{\textbf{Conclusion:} Therefore, the correct answer is option 'None of the above'.}

}

\end{minipage}

    \caption{Example output of the model trained with CoT on a question requiring existence of a crossing. The model correctly identifies the crossing but fails to extract the correct marking style.}
    \label{fig:qualitative_example2}
\end{subfigure}

\vspace{2mm}


\begin{subfigure}[t]{\linewidth}

\centering

\tikzset{
  point/.style={
    circle,
    fill=#1,
    draw=white,
    line width=0.35pt,
    inner sep=1.4pt,
    font=\tiny\bfseries,
    text=white
  }
}

\begin{minipage}[t]{0.4\linewidth}
\vspace{0pt}
\centering

\begin{minipage}{0.29\linewidth}
\centering
\begin{tikzpicture}
\node[inner sep=0, anchor=south west] (img) at (0,0) {
  \includegraphics[width=\linewidth,height=1.5cm]{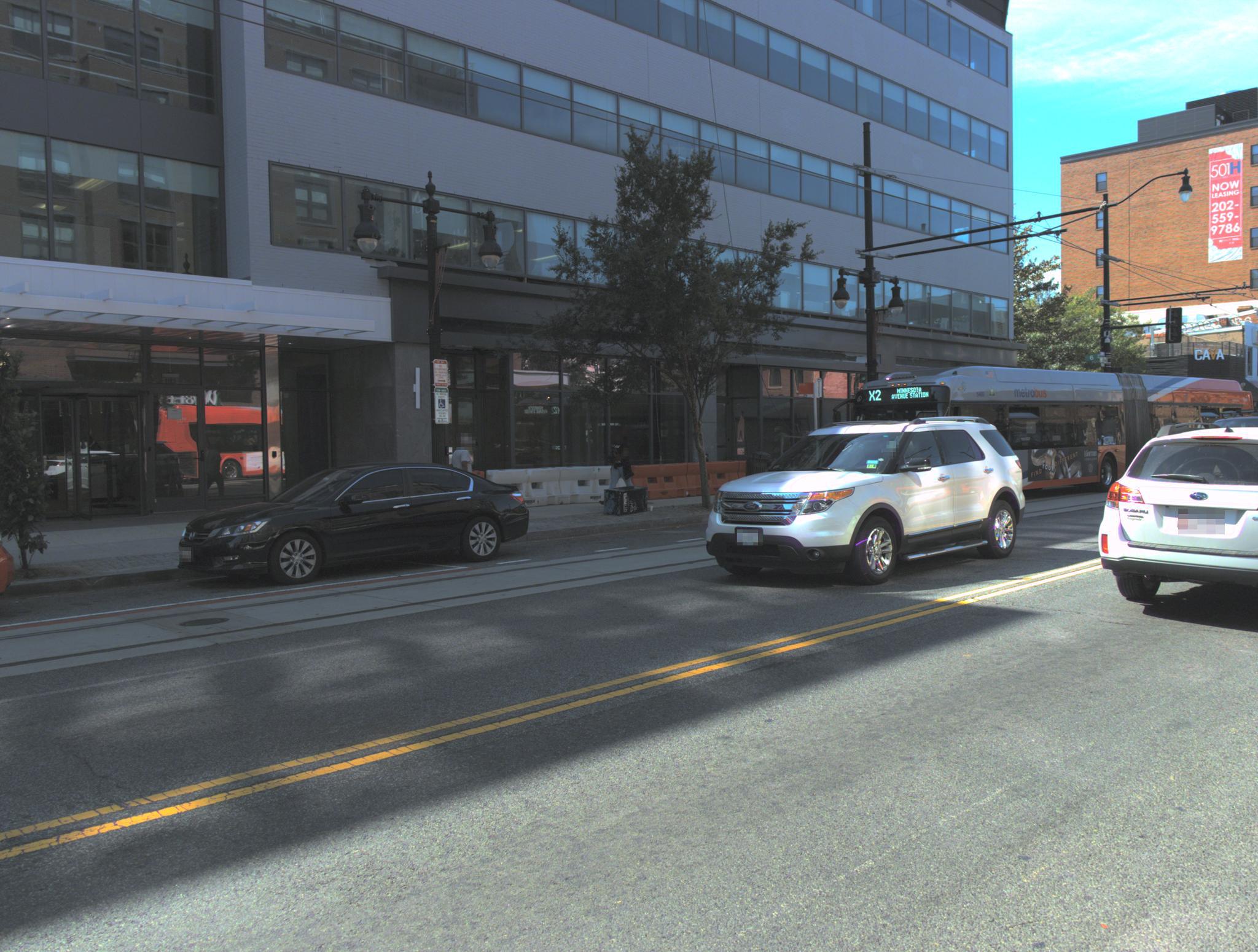}
};
\begin{scope}[x={(img.south east)}, y={(img.north west)}]
\ImgPin{0.597}{1-0.841}{P3}{_pink}
\ImgBoxPin
  {0.884}{1-0.653}   
  {0.99}{1-0.451}   
  {P4}           
  {_green}  
  {1pt}         
  {1pt}         
\ImgPin{0.402}{1-0.696}{P5}{_darkyellow}
\ImgPin{0.398}{1-0.618}{P6}{_purple}
\end{scope}
\end{tikzpicture}
\end{minipage}%
\begin{minipage}{0.38\linewidth}
\centering
\begin{tikzpicture}
\node[inner sep=0, anchor=south west] (img) at (0,0) {
  \includegraphics[width=\linewidth,height=3.2cm]{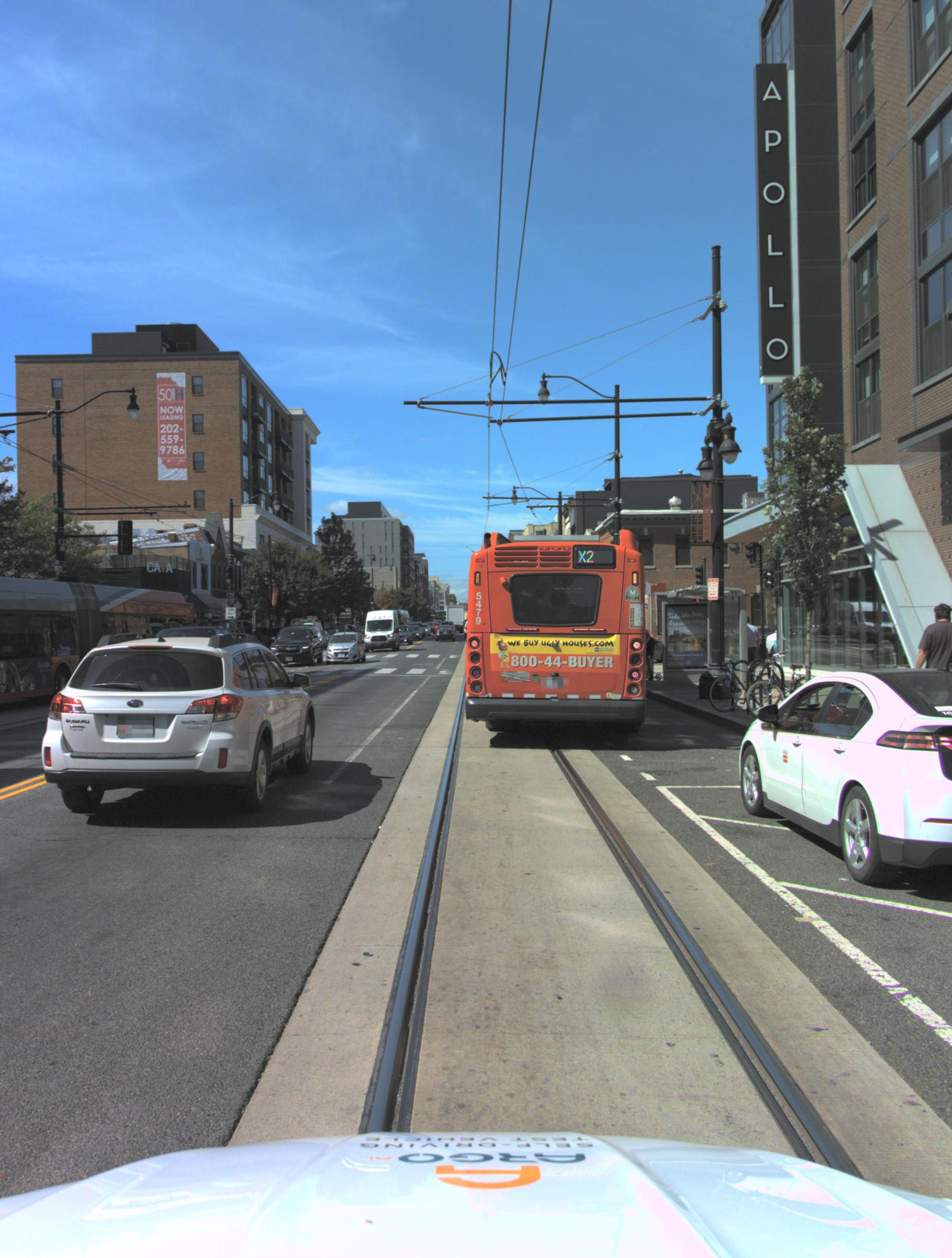}
};
\begin{scope}[x={(img.south east)}, y={(img.north west)}]
\ImgPin{0.509}{1-0.781}{P1}{_teal}
\ImgBoxPin
  {0.491}{1-0.593}   
  {0.691}{1-0.418}   
  {P2}           
  {_orange}  
  {1pt}         
  {1pt}         

\ImgPin{0.185}{1-0.695}{P3}{_pink}
\ImgBoxPin
  {0.063}{1-0.641}   
  {0.346}{1-0.508}   
  {P4}           
  {_green}  
  {1pt}         
  {1pt}         
\end{scope}
\end{tikzpicture}
\end{minipage}%
\begin{minipage}{0.29\linewidth}
\centering
\begin{tikzpicture}
\node[inner sep=0, anchor=south west] (img) at (0,0) {
  \includegraphics[width=\linewidth,height=1.5cm]{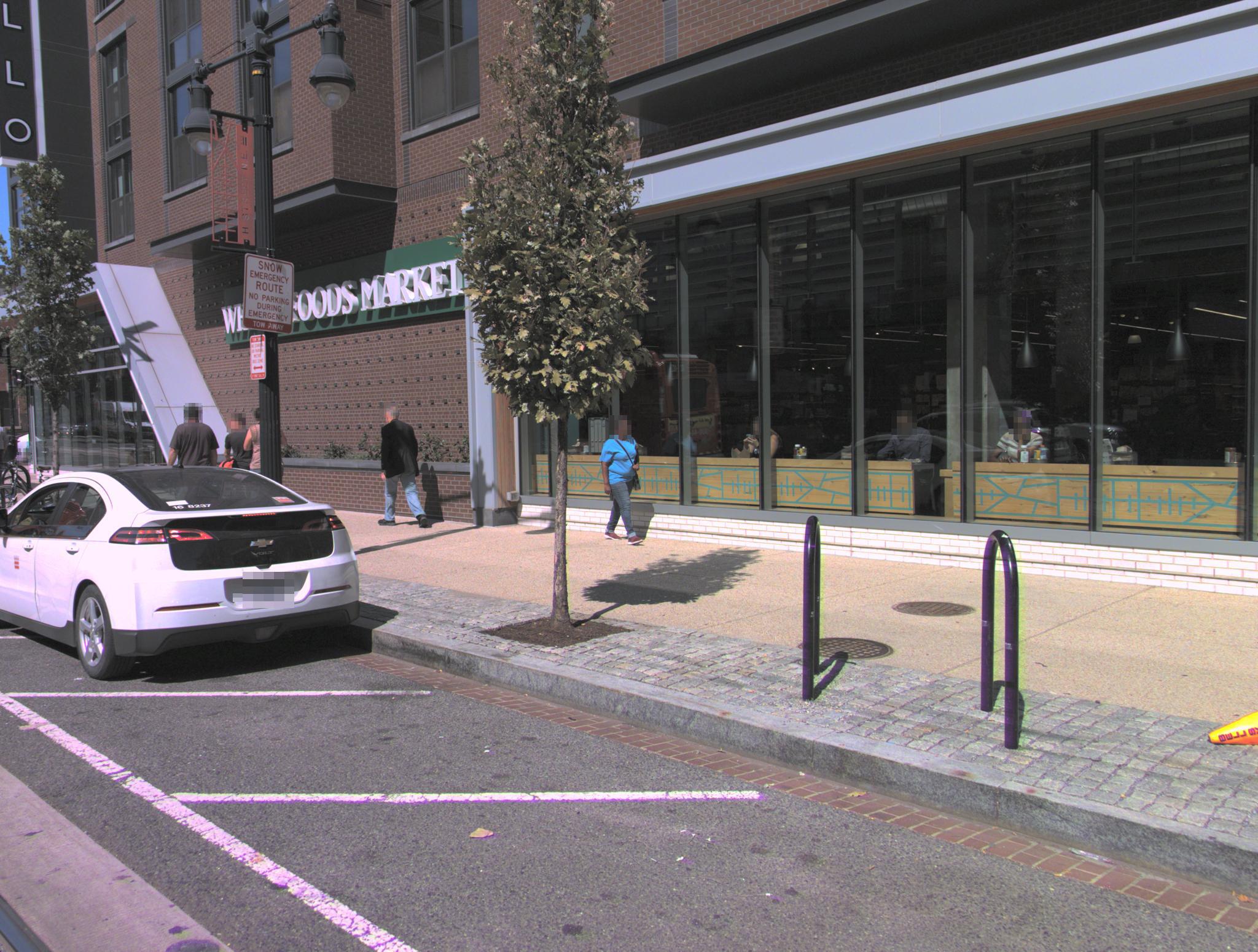}
};
\begin{scope}[x={(img.south east)}, y={(img.north west)}] 
\end{scope}
\end{tikzpicture}
\end{minipage}

\end{minipage}%
\hspace{0.1cm}
\begin{minipage}[t]{0.18\linewidth}

{\tiny\bfseries QUESTION}\par\par

{\tiny
\parbox{\linewidth}{%
  \baselineskip=6pt
 How many lanes are there?
}}\par

\tcbset{
  answerpill/.style={
    on line,
    boxrule=0.25pt,
    arc=1.2mm,
    left=1mm,
    right=1mm,
    top=0.4mm,
    bottom=0.4mm,
    boxsep=0pt,
    width=3cm  
  }
}

\vspace{0.1cm}
{\tiny\bfseries ANSWERS}\par
\AnswerPill{\color{green!50!black}\checkmark}{There are 2 lanes in direction opposite to the ego and 2 lanes leading away from the camera.}{_green}\par
\AnswerPill{B}{There are 3 lanes travelling towards the camera and 1 lane in ego direction.}{_lightgray}\par
\AnswerPill{C}{There are 2 lanes travelling towards the camera and 1 lane in ego direction.}{_lightgray}\par
\AnswerPill{D}{None of the above.}{_lightgray}\par
\end{minipage}
\hfill
\begin{minipage}[t]{0.37\linewidth}
\tcbset{
  stepbox/.style={
    colback=gray!8,
    colframe=gray!20,
    boxrule=0.25pt,
    arc=1mm,
    left=0.6mm, right=0.6mm,
    top=0.4mm, bottom=0.4mm,
    boxsep=0pt
  }
}

\newcommand{\StepBox}[1]{%
  \begin{tcolorbox}[stepbox]
  {\tiny #1}
  \end{tcolorbox}
}
{\tiny

\textbf{Reasoning Output:}
\vspace{-0.1cm}
\StepBox{\textbf{Step 1:} To determine the total number of lanes on this road, I categorize the lanes by their traffic direction.}

\StepBox{\textbf{Step 2:} I start from the ego lane, the right curb lane, which I call 'Lane 1'. Considering the images, the lane is visible in the CENTER view at \TextPin{P1}{_teal}. The lane's type is VEHICLE. The lane contains a bus with description red bus located in CENTER view at \TextPin{P2}{_orange}. Traffic in this lane moves away from the camera.}

\StepBox{\textbf{Step 3:} To the left of Lane 1, I detect another lane, the 2nd lane from the right, which I call 'Lane 2'. Considering the images, the lane is visible in the LEFT and CENTER view \TextPin{P3}{_pink}. The lane's type is VEHICLE. The lane contains a regular\_vehicle with description silver sedan and status driving located in LEFT and CENTER view at \TextPin{P4}{_green}. Traffic in this lane moves away from the camera. }

\StepBox{\textbf{Step 4:} To the left of Lane 2, I detect another lane, the 2nd lane from the left, which I call 'Lane 3'. Considering the images, the lane is visible in the LEFT view at \TextPin{P5}{_darkyellow}. The lane's type is VEHICLE. The lane leaves an intersection. Traffic in this lane moves towards the camera. 
 }
\StepBox{\textbf{Step 5:} To the left of Lane 3, I detect another lane, the left curb lane, which I call 'Lane 4'. Considering the images, the lane is visible in the LEFT view at \TextPin{P6}{_darkyellow}. The lane's type is VEHICLE. The lane leaves an intersection. Traffic in this lane moves towards the camera. 
 }
\StepBox{\textbf{Step 6:}  Based on these observations, I count exactly 4 lanes. Out of these, 2 travel away from the camera and 2 travel towards the camera. 
 }

\StepBox{\textbf{Conclusion:} Therefore, the correct answer is option A: 'There are 2 lanes in direction opposite to the ego and 2 lanes leading away from the camera.'.}
}

\end{minipage}

    \caption{Example output of the model trained with CoT on a counting question. The model steps through all lanes and identifies relevant properties and relations for grounding.}
    \label{fig:qualitative_example3}
\end{subfigure}
\caption{Qualitative examples of the model output when tasked with road reasoning.}

\label{fig:qualitative_examples}

\end{figure}

%% file: tables/benchmarks_additional.tex
\begin{table*}[t]
\centering
\caption{Accuracy is reported as decimals with 95\% scene-clustered CI. This table is an extension of \Cref{tab:model_accuracy_detailed} in the main paper.}
\label{tab:model_accuracy_detailed_suppl}
\begin{small}
\begin{tabular}{l|c|ccc}
\toprule
Model  & Total & Perception-like & Reasoning-heavy \\
\midrule
\text{gemma-4-E2B-it} \cite{google_deepmind_2026_gemma4}& 0.384 {\scriptsize (0.350, 0.414)} & 0.415 {\scriptsize (0.371, 0.455)} & 0.260 {\scriptsize (0.199, 0.332)} \\
\text{qwen3-vl-30b-a3b-instruct} \cite{Qwen3-VL}& 0.460 {\scriptsize (0.424, 0.497)} & 0.478 {\scriptsize (0.439, 0.517)} & 0.385 {\scriptsize (0.322, 0.452)} \\
\text{qwen3-vl-8b-instruct} \cite{Qwen3-VL}& 0.494 {\scriptsize (0.462, 0.526)} & 0.509 {\scriptsize (0.473, 0.545)} & 0.435 {\scriptsize (0.358, 0.513)} \\
\text{gemma-4-26b-a4b-it} \cite{google_deepmind_2026_gemma4}& 0.501 {\scriptsize (0.470, 0.533)} & 0.529 {\scriptsize (0.494, 0.564)} & 0.385 {\scriptsize (0.320, 0.448)} \\
\bottomrule
\end{tabular}
\end{small}
\end{table*}

%% file: tables/n_hops.tex
\begin{table*}[t]
\centering
\caption{Accuracy over different reasoning depths is reported as decimals with 95\% scene-clustered CI.}
\label{tab:model_accuracy_nhops}
\begin{small}
\begin{tabular}{lccccc}
\toprule
Model & 1-hop & 2-hop & 3-hop & >3-hop & -$\Delta_{1\rightarrow4}$ \\
\midrule
\text{GPT-4o-mini} \cite{openai_2024_gpt4omini} & 0.49 {\scriptsize (0.40, 0.57)} & 0.53 {\scriptsize (0.44, 0.61)} & 0.52 {\scriptsize (0.43, 0.61)} & 0.38 {\scriptsize (0.29, 0.47)} & 21.5\% \\
\text{Gemini-3.1-Flash-Lite} \cite{google2026gemini31} & 0.67 {\scriptsize (0.60, 0.74)} & 0.64 {\scriptsize (0.57, 0.71)} & 0.53 {\scriptsize (0.43, 0.63)} & 0.43 {\scriptsize (0.35, 0.51)} & 36.3\% \\
\text{Claude-Sonnet-4.5} \cite{anthropic2025claudesonnet45} & 0.61 {\scriptsize (0.54, 0.68)} & 0.57 {\scriptsize (0.49, 0.65)} & 0.57 {\scriptsize (0.46, 0.67)} & 0.51 {\scriptsize (0.43, 0.59)} & 16.3\% \\
\text{Gemini-3.1-Pro} \cite{google2026gemini31} & 0.75 {\scriptsize (0.69, 0.82)} & 0.70 {\scriptsize (0.61, 0.78)} & 0.50 {\scriptsize (0.39, 0.60)} & 0.53 {\scriptsize (0.44, 0.61)} & 30.2\% \\
\text{GPT-5.4} \cite{openai2026gpt54} & 0.77 {\scriptsize (0.70, 0.84)} & 0.76 {\scriptsize (0.68, 0.84)} & 0.65 {\scriptsize (0.56, 0.74)} & 0.50 {\scriptsize (0.42, 0.57)} & 35.8\% \\
\midrule
\text{Gemma-4-E2B-IT} \cite{google_deepmind_2026_gemma4} & 0.44 {\scriptsize (0.37, 0.53)} & 0.41 {\scriptsize (0.35, 0.48)} & 0.40 {\scriptsize (0.29, 0.50)} & 0.27 {\scriptsize (0.20, 0.35)} & 37.8\% \\
\text{Molmo2-8B} \cite{deitke2025molmo} & 0.52 {\scriptsize (0.45, 0.60)} & 0.42 {\scriptsize (0.33, 0.49)} & 0.41 {\scriptsize (0.31, 0.50)} & 0.33 {\scriptsize (0.23, 0.42)} & 37.3\% \\
\text{Kimi-k2.6} \cite{moonshotai2026kimik26} & 0.60 {\scriptsize (0.54, 0.65)} & 0.49 {\scriptsize (0.40, 0.58)} & 0.38 {\scriptsize (0.29, 0.46)} & 0.37 {\scriptsize (0.29, 0.46)} & 37.4\% \\
\text{Qwen3-VL-30b-a3b-Instruct} \cite{Qwen3-VL} & 0.54 {\scriptsize (0.47, 0.62)} & 0.57 {\scriptsize (0.48, 0.65)} & 0.50 {\scriptsize (0.42, 0.58)} & 0.39 {\scriptsize (0.30, 0.48)} & 28.1\% \\
\text{Qwen3-VL-30b-a3b-Thinking} \cite{Qwen3-VL} & 0.53 {\scriptsize (0.45, 0.61)} & 0.47 {\scriptsize (0.39, 0.56)} & 0.52 {\scriptsize (0.42, 0.63)} & 0.46 {\scriptsize (0.38, 0.53)} & 14.0\% \\
\text{Qwen3-VL-235b-a22b-Instruct} \cite{Qwen3-VL} & 0.66 {\scriptsize (0.58, 0.73)} & 0.61 {\scriptsize (0.52, 0.69)} & 0.50 {\scriptsize (0.40, 0.59)} & 0.48 {\scriptsize (0.41, 0.56)} & 26.8\% \\
\text{Gemma-4-26b-a4b-IT} \cite{google_deepmind_2026_gemma4} & 0.66 {\scriptsize (0.59, 0.73)} & 0.61 {\scriptsize (0.51, 0.70)} & 0.53 {\scriptsize (0.44, 0.63)} & 0.43 {\scriptsize (0.35, 0.51)} & 35.0\% \\
\text{Gemma-4-31b-IT} \cite{google_deepmind_2026_gemma4} & 0.68 {\scriptsize (0.60, 0.75)} & 0.64 {\scriptsize (0.54, 0.72)} & 0.61 {\scriptsize (0.52, 0.70)} & 0.53 {\scriptsize (0.44, 0.61)} & 22.6\% \\
\text{Qwen3-VL-8b-Instruct} \cite{Qwen3-VL} & 0.61 {\scriptsize (0.53, 0.68)} & 0.56 {\scriptsize (0.48, 0.64)} & 0.52 {\scriptsize (0.41, 0.63)} & 0.41 {\scriptsize (0.32, 0.50)} & 32.0\% \\
\midrule
\text{Qwen3-VL-2b-Instruct} \cite{Qwen3-VL} & 0.53 {\scriptsize (0.45, 0.61)} & 0.49 {\scriptsize (0.40, 0.58)} & 0.42 {\scriptsize (0.31, 0.51)} & 0.31 {\scriptsize (0.23, 0.39)} & 41.2\% \\
\text{~~$+$ \textit{supervised fine-tune}} & 0.82 {\scriptsize (0.76, 0.87)} & 0.75 {\scriptsize (0.68, 0.82)} & 0.72 {\scriptsize (0.63, 0.82)} & 0.68 {\scriptsize (0.60, 0.75)} & 17.1\% \\
\text{~~$+$ \textit{supervised fine-tune} $+$ \textit{CoT}} & 0.80 {\scriptsize (0.74, 0.86)} & 0.76 {\scriptsize (0.67, 0.83)} & 0.73 {\scriptsize (0.64, 0.83)} & 0.69 {\scriptsize (0.62, 0.75)} & 14.2\% \\
\text{Qwen3-VL-4b-Instruct} \cite{Qwen3-VL} & 0.55 {\scriptsize (0.46, 0.63)} & 0.53 {\scriptsize (0.45, 0.62)} & 0.47 {\scriptsize (0.38, 0.56)} & 0.36 {\scriptsize (0.28, 0.44)} & 34.3\% \\
\text{~~$+$ \textit{supervised fine-tune}} & 0.81 {\scriptsize (0.75, 0.88)} & 0.72 {\scriptsize (0.62, 0.80)} & 0.81 {\scriptsize (0.72, 0.88)} & 0.70 {\scriptsize (0.62, 0.78)} & 13.8\% \\
\text{~~$+$ \textit{supervised fine-tune} $+$ \textit{CoT}} & 0.80 {\scriptsize (0.74, 0.86)} & 0.76 {\scriptsize (0.68, 0.83)} & 0.78 {\scriptsize (0.67, 0.87)} & 0.71 {\scriptsize (0.63, 0.78)} & 11.4\% \\
\bottomrule
\end{tabular}
\end{small}
\end{table*}

%% file: figures/supplementary/free_form_img.tex
\pgfplotstableread[col sep=semicolon]{figures/supplementary/free_form_img_per_type_stats.csv}\pertype

\pgfplotstableread[col sep=comma]{figures/supplementary/free_form_img_overall_stats.csv}\overall

\begin{center}
\tikz \draw[fill=withimg, draw=withimg!80!black] (0,0) rectangle (0.2cm,0.15cm);%
\;\small w/ img\quad
\tikz \draw[fill=noimg, draw=noimg!80!black] (0,0) rectangle (0.2cm,0.15cm);%
\;\small no img%
\end{center}

\vspace{-0.5em}

\begin{tikzpicture}

\begin{axis}[
    ybar,
    bar width        = 10pt,
    width            = \textwidth,
    height           = 0.27\textwidth,
    enlarge x limits = 0.04,
    ymin = 0, ymax = 105,
    ytick = {0,25,50,75,100},
    yticklabels = {0,0.25,0.50,0.75,1.00},
    yticklabel style = {font=\scriptsize},
    ylabel       = {Accuracy},
    ylabel style = {font=\scriptsize},
    xtick          = data,
    xticklabels from table={\pertype}{type},
    xticklabel style = {
    rotate      = 25,
    anchor      = north east,
    font        = \ttfamily\tiny,
    align       = right,
    text width  = 1.8cm,
},
    axis on top    = false,
    clip = false,
]

\addplot[fill=withimg, draw=withimg!80!black, bar shift=-5.5pt]
    table[x expr=\coordindex, y=acc_with_img, col sep=semicolon]
    {figures/supplementary/free_form_img_per_type_stats.csv};

\addplot[fill=noimg, draw=noimg!80!black, bar shift=5.5pt]
    table[x expr=\coordindex, y=acc_no_img, col sep=semicolon]
    {figures/supplementary/free_form_img_per_type_stats.csv};

\draw[dashed, withimg!80, line width=0.8pt]
    ({rel axis cs:0,0}|-{axis cs:0,61.86}) --
    ({rel axis cs:1,0}|-{axis cs:0,61.86});

\draw[dashed, noimg!80, line width=0.8pt]
    ({rel axis cs:0,0}|-{axis cs:0,48.85}) --
    ({rel axis cs:1,0}|-{axis cs:0,48.85});

\end{axis}
\end{tikzpicture}

%% file: figures/supplementary/free_form_img_counting.tex
\newcommand{\countingplot}[3]{%
    \begin{tikzpicture}[baseline=(current axis.north)]
    \begin{axis}[
        ybar,
        bar width        = 4pt,
        width            = 0.40\textwidth,
        height           = 0.24\textwidth,
        enlarge x limits = 0.15,
        ymin = 0,
        xlabel       = {Count value},
        xlabel style = {font=\scriptsize},
        xtick        = data,
        xticklabel style = {font=\scriptsize},
        yticklabel style = {font=\scriptsize},
        title        = {\ttfamily\small #2},
        title style  = {font=\small, align=center},
        legend image code/.code={
            \draw[##1] (0cm,-0.1cm) rectangle (0.3cm,0.2cm);
        },
        #3,
        clip = false,
    ]
    \addplot[fill=gtcol,  draw=gtcol!80!black,  bar shift=-4.5pt]
        table[x=count_value, y=gt,            col sep=comma]
        {figures/supplementary/#1.csv};
    \addplot[fill=withimg,    draw=withimg!80!black,    bar shift=0pt]
        table[x=count_value, y=pred_with_img, col sep=comma]
        {figures/supplementary/#1.csv};
    \addplot[fill=noimg,    draw=noimg!80!black,    bar shift=4.5pt]
        table[x=count_value, y=pred_no_img,   col sep=comma]
        {figures/supplementary/#1.csv};
    \end{axis}
    \end{tikzpicture}%
}

\newcommand{\categoricalplot}[3]{%
    \pgfplotstableread[col sep=comma]{figures/supplementary/#1.csv}\pertype
    \begin{tikzpicture}[baseline=(current axis.north)]
    \begin{axis}[
        ybar,
        bar width        = 4pt,
        width            = 0.40\textwidth,
        height           = 0.24\textwidth,
        enlarge x limits = 0.15,
        ymin = 0,
        xlabel style = {font=\tiny},
        xtick        = data,
        xticklabels from table={\pertype}{count_value},
        xticklabel style = {
            rotate    = 25,
            anchor    = north east,
            font      = \ttfamily\tiny,
            align     = right,
        },
        yticklabel style = {font=\scriptsize},
        title        = {\ttfamily\small #2},
        title style  = {font=\small, align=center},
        legend image code/.code={
            \draw[##1] (0cm,-0.1cm) rectangle (0.3cm,0.2cm);
        },
        #3,
        clip = false,
    ]
    \addplot[fill=gtcol,  draw=gtcol!80!black,  bar shift=-4.5pt]
        table[x expr=\coordindex, y=gt,            col sep=comma]
        {\pertype};
    \addplot[fill=withimg,    draw=withimg!80!black,    bar shift=0pt]
        table[x expr=\coordindex, y=pred_with_img, col sep=comma]
        {\pertype};
    \addplot[fill=noimg,    draw=noimg!80!black,    bar shift=4.5pt]
        table[x expr=\coordindex, y=pred_no_img,   col sep=comma]
        {\pertype};
    \end{axis}
    \end{tikzpicture}%
}

\begin{figure}[htbp]
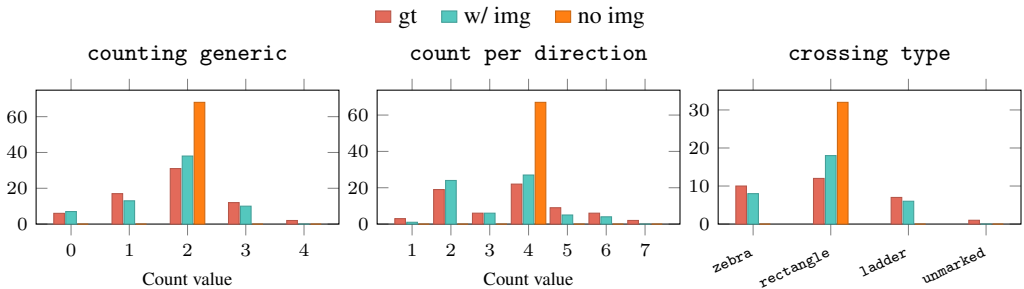

    \centering
    \centering
    \tikz \draw[fill=gtcol, draw=gtcol!80!black] (0,0) rectangle (0.2cm,0.15cm);%
    \;\small gt\quad
    \tikz \draw[fill=withimg, draw=withimg!80!black] (0,0) rectangle (0.2cm,0.15cm);%
    \;\small w/ img\quad
    \tikz \draw[fill=noimg, draw=noimg!80!black] (0,0) rectangle (0.2cm,0.15cm);%
    \;\small no img%
    \vspace{0.3em}\\
    
\begin{tabular}[t]{@{}c@{\hfill}c@{\hfill}c@{}}
    \countingplot{counting_off_intersection_generic}{counting generic}{}%
    &
    \countingplot{counting_off_intersection_direction}{count per direction}{}%
    &
    \categoricalplot{pc_type}{crossing type}{}%
\end{tabular}
    \caption{Prediction frequency distributions for 3 descriptive question types.}
    \label{fig:counting-histograms-supplementary}
\end{figure}